\documentclass{article}

\usepackage{microtype}
\usepackage{graphicx}
\usepackage{subfigure}
\usepackage{booktabs} 

\usepackage[shortcuts]{extdash}			
\usepackage{amsmath, amsfonts, amssymb }
\usepackage{ mathrsfs }
\usepackage{ commath }
\usepackage[english]{babel}
\usepackage{acronym}
\usepackage{doi}				

\usepackage[disable]{todonotes}				
\usepackage[percent]{overpic}				
\usepackage{transparent}					
\usepackage{xifthen}

\usepackage{hyperref}

\usepackage[accepted]{icml2020}

\icmltitlerunning{Inverting the Feature Visualization Process for Feedforward Neural Networks}

\DeclareMathOperator{\agg}{Agg}
\DeclareMathOperator{\net}{\widetilde{N}}

\DeclareMathOperator{\hoyer}{hoyer}
\DeclareMathOperator{\coker}{coker}

\DeclareMathOperator{\spanop}{span}
\DeclareMathOperator{\slerpop}{SLERP}
\DeclareMathOperator{\svd}{SVD}
\DeclareMathOperator*{\argmin}{arg\,min}
\DeclareMathOperator*{\argmax}{arg\,max}

\def\natural		{\mathbb{N}}
\def\opt			#1{{\mathbf{#1}^*}}
\def\real			{\mathbb{R}}
\def\fobj			{\mathbf{x}}
\def\fresp			{\mathbf{y}}
\def\id				{\mathit{Id}}
\def\bigO			{\mathcal{O}}

\def\defeq			{\mathrel{\mathop{:}}=}

\newcommand{\textcite}[2][]{%
	\ifthenelse{\isempty{#1}}%
		{\citeauthor{#2}\ (\citeyear{#2})}%
		{\citeauthor{#2}\ (\citeyear[#1]{#2})}%
	}

\acrodef{cnn}[CNN]{convolutional neural network}
\acrodef{relu}[ReLU]{rectified linear activation function}
\acrodef{slerp}[SLERP]{spherical linear interpolation}
\acrodef{bb}[BB]{Barzilai-Borwei}
\acrodef{svd}[SVD]{singular value decomposition}
\acrodef{ssim}[SSIM]{Structural Similariy}
\acrodef{FV}[FV]{Feature Visualization}
\acrodef{IFV}[IFV]{\emph{Inverse Feature Visualization}}
\acrodef{Grad-IFV}[Grad\=/IFV]{\emph{Gradient based Inverse Feature Visualization}}
\acrodef{ifft}[iFFT]{inverse discrete fourier transform}
\acrodef{lstm}[LSTM]{long short-term memory}
\acrodef{kde}[kde]{kernel density estimation}

\newcommand{\figureTeaser}[1]{%
	\begin{figure}[#1]
		\centering
		\includegraphics[width=\columnwidth]{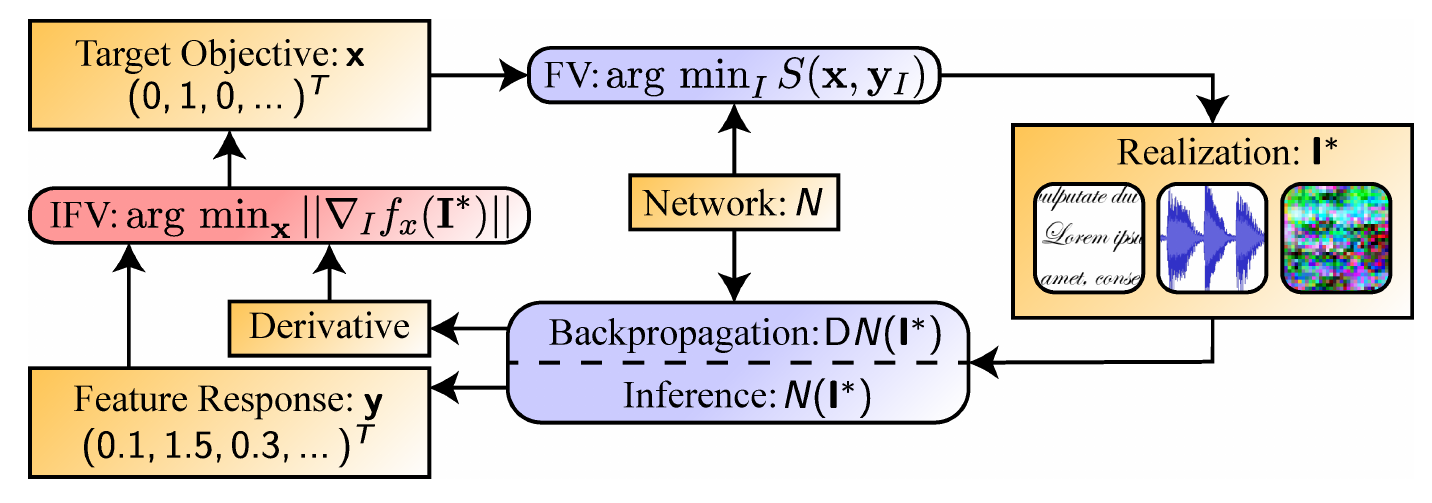}
		\caption{%
			Interplay between \acf{FV} and Inverse Feature Visualization \acs{IFV} (in red). Given an input $\opt{I}$ that is optimized via \ac{FV} for a prescribed target objective $\fobj$, \ac{IFV} recovers $\fobj$ solely from the network's response to $\opt{I}$ and gradient information.
			If features and neurons of the network are not in 1:1\=/correspondence, this is ensured via an aggregation term $\agg$---concatenated with the network function $N$---that maps activations to features. The function $f_\fobj$ is given by $I \mapsto S(\fobj,\fresp_I)$.
		}
		\label{fig:teaser}
	\end{figure}
}

\newcommand{\figureCoefJoint}[1]{%
	\begin{figure*}[#1]
		\centering
		\begin{overpic}[width=0.325\textwidth, tics = 10, trim = 0 0 0 0 , clip]
			{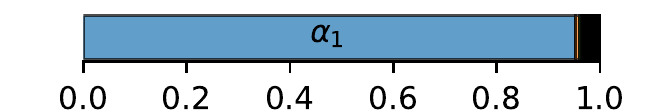}
		\end{overpic}
		\begin{overpic}[width=0.325\textwidth, tics = 10, trim = 0 0 0 0 , clip]
			{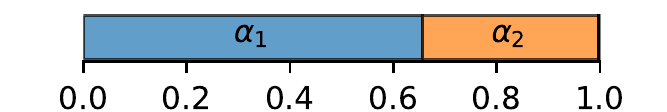}
		\end{overpic}
		\begin{overpic}[width=0.325\textwidth, tics = 10, trim = 0 0 0 0 , clip]
			{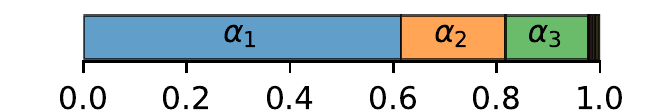}
		\end{overpic}
	
		\vspace*{2mm}
		
		\begin{overpic}[width=0.325\textwidth, tics = 10, trim = 0 0 0 0 , clip]
			{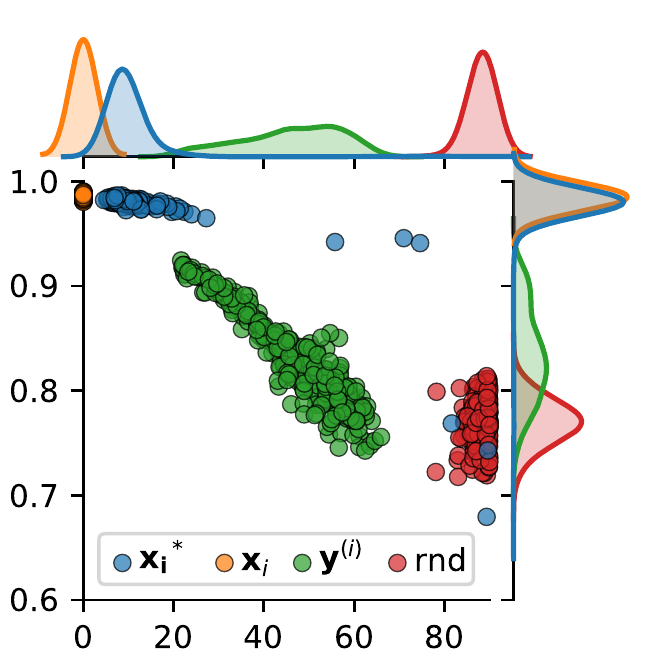}
			\put(0, 2){\textbf{\textcolor{black}{\large (a)}}}
		\end{overpic}
		\begin{overpic}[width=0.325\textwidth, tics = 10, trim = 0 0 0 0 , clip]
			{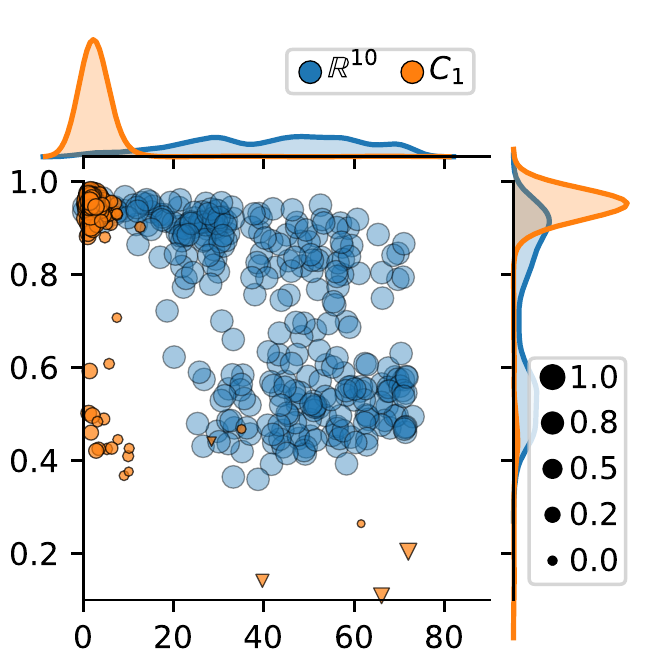}
			\put(0, 2){\textbf{\textcolor{black}{\large (b)}}}
		\end{overpic}
		\begin{overpic}[width=0.325\textwidth, tics = 10, trim = 0 0 0 0 , clip]
			{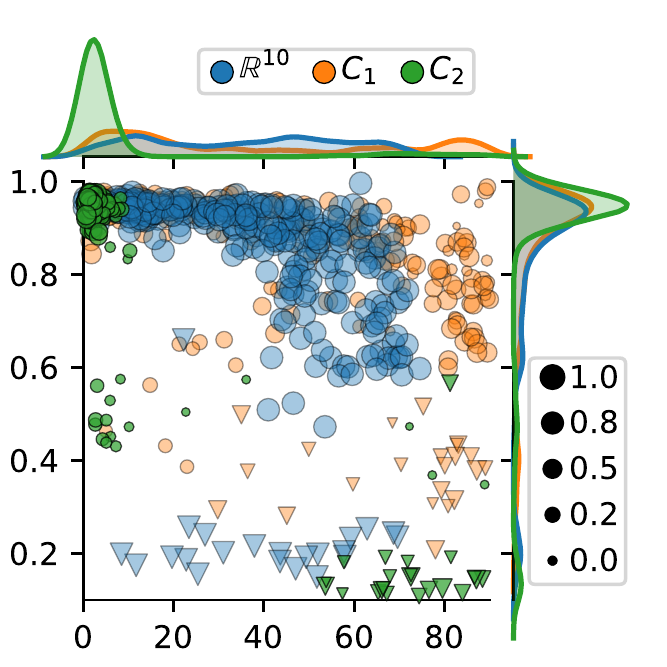}
			\put(0, 2){\textbf{\textcolor{black}{\large (c)}}}
		\end{overpic}
		\caption{%
			\textbf{(a)} GoogLeNet, %
			\textbf{(b)} DenseNet, %
			\textbf{(c)} DenseNetEx4. %
			\textbf{(top)} Values $\alpha_1,\dots,\alpha_{10}$ indicating the mean contribution of $j$\=/th left\=/singular vectors to the target objectives, stacked in a bar chart in increasing order of $j$. Bars of missing values are too small for plotting. %
			\textbf{(bot (a))} Bivariate distribution of angular distances to the target objectives (horizontal axis) and \acs{ssim} indices (vertical axis) when re\=/optimizing w.r.t.\ predicted objectives, target objectives, feature responses or random feature objectives. %
			\textbf{(bot (b,c))} Trivariate distribution when re\=/optimizing w.r.t.\ predicted objectives for different critical spaces. Point size encodes the fraction of the target objective $\fobj_i$ that lives in the respective critical space, that is $||v||/(||v||+||w||)$ with $\fobj_i = v + w$, $v\in Z_k(\fresp^{(i)})^{-1}C$, $w\in (Z_k(\fresp^{(i)})^{-1}C)^\perp$.
			Triangles represent instable samples for which re\=/optimization with the target objective yields an \ac{ssim} index lower than $0.7$.
		}
		\label{fig:joint_coef_plots}
	\end{figure*}
}

\newcommand{\figureDropCosine}[1]{%
	\begin{figure}[#1]
		\centering
		\includegraphics[width=\columnwidth]{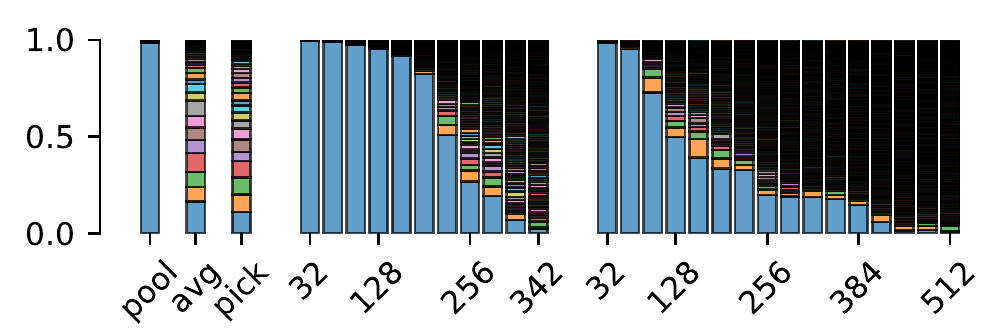}
		\caption{%
			Values of $\alpha_j$ (i.e., mean contribution of $j$-th singular vectors, ordered by singular values, to target objectives) for different experiments with dropped cosine term, stacked in a bar chart in increasing order of $j$. Black indicates high fragmentation. %
			Statistics for \textbf{(left)} $80$ filters in the \texttt{dense3} block of DenseNet for different aggregation routines,
			\textbf{(middle)} $n_f$ filters of DNde, and \textbf{(right)} GNcl, where $n_f$ increases in steps of $32$.
		}
		\label{fig:drop_cosine}
	\end{figure}
}

\newcommand{\figureRealizationsForParams}[1]{%
	\begin{figure}[#1]
		\centering
		\begin{overpic}[width=0.49\columnwidth, tics = 10, trim = 0 0 0 0 , clip]
			{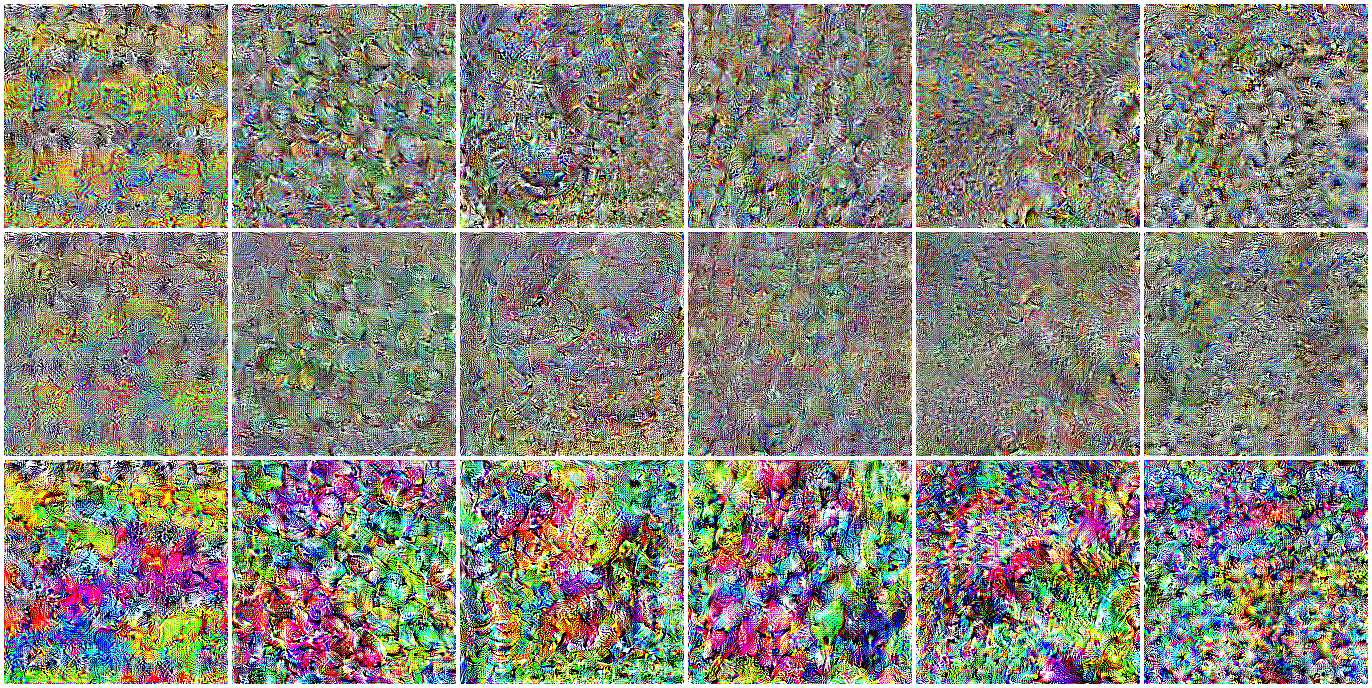}
			\put(0,-0.2){\transparent{0.7}\textcolor{white}{\rule{5.2mm}{3.9mm}}}
			\put(0, 2){\textbf{\textcolor{black}{\large (a)}}}
		\end{overpic}
		\begin{overpic}[width=0.49\columnwidth, tics = 10, trim =  0 0 0 0 , clip]
			{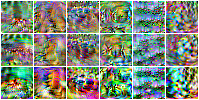}
			\put(0,-0.2){\transparent{0.7}\textcolor{white}{\rule{5.4mm}{3.9mm}}}
			\put(0, 2){\textbf{\textcolor{black}{\large (b)}}}
		\end{overpic}
		\caption{%
			Realizations returned by \acs{FV} when parameterizing images through \textbf{(top)} $P_\text{rgb}$, \textbf{(mid)} $P_\text{fft}$ and \textbf{(bot)} $P_\text{ffte}$.
			Visualized features stem from \textbf{(a)} GoogLeNet and \textbf{(b)} DenseNet.
		}
		\label{fig:realizations_for_params}
	\end{figure}
}

\newcommand{\tableTimings}[1]{%
	\begin{table}[#1]
		\caption{Timings for computing realizations via \ac{FV} ($\opt{I_i}$), back\=/propagation ($\Dif\fresp^{(i)}$), solving Opt.~(\ref{eq:reformulation}) ($\opt{x_i}$) and re\=/optimization of $\opt{I_i}$ (re-opt.), for different number of input parameters ($n_p$) and features ($n_f$). All timings are in seconds \emph{per sample}. Timings associated to \acs{IFV} are in bold. Critical space computation for DN[4]cl takes $0.5$s on average per space. Batch sizes in \ac{FV} are 8 for GN4c, 64 for DN[4]cl and DNde, and 32 for SRNet. Re\=/optimization always is performed with a batch size of 64.
		Solutions of Opt.~(\ref{eq:reformulation}) are computed for all samples at once, in a single batch.
		}
		\vspace*{0.1in}
		\label{table:timings}
		\centering
		\setlength{\tabcolsep}{5pt}
		\begin{tabular}{lrrrcrr}
			\toprule
			Model & \multicolumn{1}{c}{$n_p$} & \multicolumn{1}{c}{$n_f$} &  \multicolumn{1}{c}{$\opt{I_i}$} & \multicolumn{1}{r}{$\Dif\fresp^{(i)}$} & \multicolumn{1}{c}{$\opt{x_i}$} & \multicolumn{1}{c}{re\=/opt.} \\
			\midrule
			GN4c & $224^2\cdot3$ & $512$ & $17.3$ & $\mathbf{0.7}$ & $\mathbf{0.2}$ & $0.7$ \\
			DN[4]cl & $32^2\cdot3$ & $10$ & $2.1$ & $\mathbf{0.0}$ & $\mathbf{0.0}$ & $0.7$ \\
			DNde & $32^2\cdot3$ & $342$ & $2.0$ & $\mathbf{0.3}$ & $\mathbf{0.1}$ & $0.7$ \\
			SRNet & $128^2\cdot4$ & $64$ & $10.7$ & $\mathbf{0.4}$ & $\mathbf{0.0}$ & $3.1$ \\
			\bottomrule
		\end{tabular}
	\end{table}
}

\newcommand{\algorithmCriticalSpace}{%
	\begin{algorithm}[tb]
	   \caption{Co\=/kernel Approximation}
	   \label{alg:cokernel}
	\begin{algorithmic}
	   \STATE {\bfseries Input:} Samples $\Dif\fresp^{(1)},\dots,\Dif\fresp^{(n)}$, threshold $\rho$
	   \FOR{$i=1$ {\bfseries to} $n$}
	   \STATE Compute $U\Sigma V^T = \svd(\Dif\fresp^{(i)})$
	   \STATE $S_i \leftarrow \spanop\{U_{\cdot,j}\mid \Sigma_{j,j} < \rho \cdot ||\Dif\fresp^{(i)}||_2\}$
	   \ENDFOR
	   
	   \STATE $L \leftarrow \{(S_1, 1),\dots,(S_n, 1)\}$
	   \WHILE{$|L| > 1$}
	   \STATE Remove $(U, w_U)$, $(V, w_V)$ from $L$
	   \STATE Compute principal vectors $(p_U^1,p_V^1),\dots,(p_U^m,p_V^m)$
	   \STATE Drop all pairs with $\sphericalangle(p_U^j, p_V^j) > 45^\circ$
	   \STATE Interpolate $p_I^j \leftarrow \slerpop(p_U^j, p_V^j; w_V / (w_U + w_V))$
	   \STATE $L \leftarrow L \cup \{(\spanop\{p_I^j\mid j\ \text{not dropped}\}, w_U + w_V)\}$
	   \ENDWHILE
	   \STATE {\bfseries return} $U^\perp$ with $(U, w_U)\in L$
	\end{algorithmic}
	\end{algorithm}
}

\binoppenalty=\maxdimen
\relpenalty=\maxdimen

\begin{document}

\twocolumn[
\icmltitle{Inverting the Feature Visualization Process for Feedforward Neural Networks}



\icmlsetsymbol{equal}{*}

\begin{icmlauthorlist}
\icmlauthor{Christian Reinbold}{cgtum}
\icmlauthor{R\"udiger Westermann}{cgtum}
\end{icmlauthorlist}

\icmlaffiliation{cgtum}{Chair of Computer Graphics and Visualization, Technical University of Munich, Bavaria, Germany}


\icmlkeywords{Machine Learning, ICML, Feature Visualization}

\vskip 0.3in
]



\printAffiliationsAndNotice{}  

\begin{abstract}
	This work sheds light on the invertibility of feature visualization in neural networks. Since the input that is generated by feature visualization using activation maximization does, in general, not yield the feature objective it was optimized for, we investigate optimizing for the feature objective that yields this input.
	Given the objective function used in activation maximization that measures how closely a given input resembles the feature objective, we exploit that the gradient of this function w.r.t.\ inputs is---up to a scaling factor---linear in the objective. This observation is used to find the optimal feature objective via computing a closed form solution that minimizes the gradient. By means of Inverse Feature Visualization, we intend to provide an alternative view on a networks sensitivity to certain inputs that considers feature objectives rather than activations.
\end{abstract}

\section{Introduction}



To better understand the learning behavior of neural networks, the similarity of representations learned by differently trained networks has been assessed by statistical analysis of activation data \cite{li2015convergent, raghu2017svcca}. \textcite{wang2018towards} search for similar representations by using activation vectors and matching them over different networks.
These approaches can determine similar behavior of neurons for a finite set of inputs,
but they do not consider which patterns the neurons are sensitive for and, thus, neglect the semantic meaning of representations.
There is also no evidence that representations behave similarly on different input sets, so that the findings are sensitive to the choice of inputs.

Another class of approaches, mainly used in image understanding tasks, tackle the problem of identifying patterns a network reacts to.
For instance, pixel-wise explanations \cite{bach2015pixelwise} and saliency maps \cite{simonyan2013deep} aim to reveal areas in input samples that certain inference tasks or neurons are sensitive to.
Vice versa, activation maximization \cite{erhan2009visualizing, nguyen2015fooled, nguyen2016synthesizing, szegedy2013intriguing} or code inversion \cite{mahendran2015understanding} target the reconstruction of input samples with certain activation characteristics. The recent survey of \textcite{nguyen2019survey} gives a thorough overview of activation maximization approaches used in \ac{FV}.
Some techniques do not only analyze the activations of a single neuron, but consider groups of neurons that form a semantic unit \cite{olah2018building}.
Groupings arise from the investigated network topology, i.e., convolution filters can act as semantic units of \acp{cnn}, grouping all neurons together that share identical filter weights.
Henceforth we call these units the \emph{features} we aim to analyze, and denote by $n_f$ the finite number of available features.
To measure the features' stimulus w.r.t.\ an input sample $I$, neuron activations of a given network $N$ are aggregated into a single value per neuron group, yielding the input's \emph{feature response}---denoted by $\fresp_I$, a vector of dimension $n_f$.

In its pure form, activation maximization optimizes for an input sample $\opt{I}$ (from the set $\mathcal{I}$ of all valid inputs) so that $\fresp_\opt{I}$ resembles a prescribed vector $\fobj\in\real^{n_f}$, i.e.,
\begin{equation}
\label{eq:activation_maximization}
\opt{I} = \argmax_{I\in\mathcal{I}} S(\fobj, \fresp_I),
\end{equation}
where $S(\fobj, \fresp)$ is a measure of significance of $\fresp$ regarding $\fobj$.
We call $\opt{I}$ the \emph{realization} of $\fobj$, and $\fobj$ the \emph{target objective} of $\opt{I}$. Maximizing for a single feature can be achieved by setting $\fobj = e_i$.

Inputs stimulating a certain feature usually stimulate also many other features, yet to lesser extent.
Hence, although penalized by the optimization process, feature response $\fresp_\opt{I}$ and target objective $\fobj$ can differ substantially.
Extending the argument that $\opt{I}$ represents a facet of a neuron that hints towards patterns (and their granularity) it is sensitive for, we argue that an input does not necessarily represent the neurons it stimulates most, but instead should represent the neurons which cannot be stimulated more strongly by other inputs.
For instance, an input $I$ of stripes may stimulate a cross and a stripe detector neuron to equal amounts.
Representation matching techniques consider both neurons equal, especially when crosses are not contained in the inputs for which activations are drawn.
We argue that they represent two semantically different concepts.
Instead of using activations, as it is done in many approaches, we aim for a semantically richer representation in the form of inversely reproduced target objectives. That is, we suggest to consider the target objective $\fobj$ that yields $I$ when applying \ac{FV}.
Thus, stripes in the input will only be associated with stripe detectors, while cross detectors are stimulated more if stripes are exchanged with crosses.

In this work, we make a small step towards identifying target objectives. We propose a method for \ac{IFV} that---given a network that can be back\=/propagated and an input optimized via \ac{FV}---can reconstruct the input's target objective $\fobj$.
Since the \ac{FV} process neither has to be deterministic nor injective---i.e. different values for $\fobj$ do not necessarily infer different optima $\opt{I}$---a rigorous definition of \ac{IFV} is more complex:
When seeing \ac{FV} as a random variable $Y$ over the domain $\mathcal{I}$ with its probability density function $h_Y$ depending on $\fobj$ as additional parameter, \ac{IFV} means to compute the maximum likelihood estimator of $Y$, i.e.\ $\widehat{\fobj} = \argmax_{\fobj} \widehat{h_Y}(I;\fobj)$ for a given input configuration $I$.

Since realizations $\opt{I}$ returned by \ac{FV} are locally optimal solutions of the objective function $f_\fobj\colon I \mapsto S(\fobj,\fresp_I)$ (or are at least close to them), the necessary condition for optimality $||\nabla f_\fobj(I)|| = 0$ is supposed to hold.
In order to recover the (most likely) target objective of a realization $\opt{I}$, we solve for
\begin{equation}
\label{eq:minimize_grad}
\argmin_{\fobj\in\Omega} ||\nabla f_\fobj(\opt{I})||^2.
\end{equation}
with $\Omega$ being the space of allowed target objectives. An overview of the principle approach underlying our work is shown in Fig.~\ref{fig:teaser}.

\figureTeaser{t}

Unfortunately, solving Opt.~(\ref{eq:minimize_grad}) directly usually fails due to trivial solutions for $\fobj$.
These occur at saddle or minimal points of $f_\fobj$, or are induced by nontrivial co\=/kernels of matrices that propagate when chaining Jacobian matrices to form $\nabla f_\fobj$.
We introduce a method called \ac{Grad-IFV} to address these limitations. Our key idea is to eliminate saddle points by introducing factors in $f_\fobj$ and intersecting the search space $\Omega$ with an appropriate subspace of $\real^{n_f}$. The consequential reformulation of Opt.~(\ref{eq:minimize_grad}) is solved by computing a singular value decomposition of a matrix derived from the gradient of the objective function $f_\fobj$.

While in this work we demonstrate that \ac{Grad-IFV} can reproduce the target objective for which a given input was optimized, building upon this observation it then needs to be investigated whether \ac{Grad-IFV} can be extended to arbitrary input. In this way, it may even become possible to control certain patterns in the inputs and ask for the specific features that are sensitive for them.
This sheds light on the question whether it can be determined which patterns in the given inputs are relevant and which feature combinations of a network have learned these patterns, i.e., strive toward transfer learning.
By inverting an input representing feature $\fobj$ in one network, the feature $\widetilde{\fobj}$ that is represented by this input in another network can be obtained. This can give rise to a feature-based comparison of learned representations, including insight into the relevance of patterns for successful network training.

\section{Method}
\label{sec:method}

In this work, we consider \ac{FV} objective functions provided by the Lucid library \cite{olah2017feature}. For a thorough discussion of the use and interpretation of objective functions in the context of \ac{FV}, let us refer to the recent work by \textcite{carter2019activation}.
The objective function $f_\fobj$ is a concatenation of three functions $N$, $\agg$ and $S_\fobj$, where $N\colon\mathcal{I}\rightarrow \mathcal{A}$ represents the network and maps an input to activations that were tapped from the network, $\agg\colon\mathcal{A} \rightarrow \real^{n_f}$ aggregates neuron activations into a \emph{feature response} $\fresp$ of which component $i$ represents excitement of feature $i$ and $S_\fobj\colon\real^{n_f} \rightarrow \real$ is defined as $S_\fobj(\fresp) = \fobj^T\fresp\cdot (\fobj^T\fresp/||\fresp||)^k$ with $k\in\natural_0$.
The set of allowed target objectives $\Omega$ is set to the $n_fD$\=/sphere containing \emph{feature directions}, i.e.\ normalized linear combinations of features.
The term $\fobj^T\fresp$ measures the length of the projection of $\fresp$ onto $\fobj$, and $\fobj^T\fresp/||\fresp||$ is the cosine similarity of $\fobj$ and $\fresp$.

If $N$ has an input domain $\mathcal{I}$ that is not a vector space $\real^m$, Opt.~(\ref{eq:minimize_grad}) is a constrained optimization problem for which a) common solvers such as gradient descent or Adam cannot be applied, and b) the necessary condition for optimality can be violated at the domain boundary $\partial \mathcal{I}$.
These issues can be circumvented by means of a differentiable, surjective mapping $P\colon\real^{n_p} \rightarrow \mathcal{I}$, which parametrizes the input domain $\mathcal{I}$ by a real\=/valued vector space of dimension $n_p$.
If such a $P$ is available, the constrained optimization problem $\argmin_{I\in\mathcal{I}} f_\fobj(I)$ can be transformed to an unconstrained one over $n_p$ variables via $\argmin_{v\in\real^{n_p}} (f_\fobj \circ P)(v)$.

In the following, if $I\in\mathcal{I}$ is an input configuration, we use $v_I$ to indicate a valid parameter vector that describes $I$, i.e., $P(v_I) = I$. Furthermore, $P$, $N$, $\agg$ are combined into a single function $\net \defeq \agg \circ N \circ P$, which maps parameter vectors to feature responses.
The feature response $\net(v_I)$ of $I$ is denoted by $\fresp_I$, and the Jacobian matrix $\Dif\net(v_I)$ is written as $\Dif\fresp_I$.

\subsection{Reformulation of $\nabla f_\fobj$}
\label{sec:reformulation}

Given the described family of objective functions, the gradient $\nabla f_\fobj$ can be computed.
The chain rule allows us to write
$\nabla (f_\fobj \circ P )(v_I) = (k+1)q_I^k x^T Z_k(\fresp_I) \Dif\fresp_I$,
where $Z_k(\fresp) = Z_k(\fresp)^T \defeq \id - k/(k+1)\overline{\fresp}\overline{\fresp}^T$,
$\overline{\fresp} \defeq \fresp / ||\fresp||$, $q_I \defeq \fobj^T\overline{\fresp_I}$ (see supp. material).
The scalar factor $(k+1)q_I^k$ equals zero if $\fobj^T\fresp_I = 0$.
Depending on the parity of $k$, values of $\fresp_I$ with $\fobj^T\fresp_I = 0$ either yield minimal or saddle points of $S_\fobj$. Since one is interested in maximizing $S_\fobj$, the factor can be safely dropped.
Hence, if $\opt{I}$ is a local maximum for $f_\fobj$, this implies that 
$||\fobj^T Z_k(\fresp_\opt{I}) \Dif\fresp_\opt{I} || = 0$.

In practice, the investigated network may have linear relations that result in $\Dif\fresp_I$ and potentially $Z_k(\fresp_I) \Dif\fresp_I$ having a nontrivial co\=/kernel that is similar for all $I\in\mathcal{I}$.
For instance, dead neurons absorb gradients and introduce zero rows such that $e_i\in\bigcap_{I\in\mathcal{I}}\coker\Dif\fresp_I$ for some $i$.
As a consequence, $e_i$ will always be a trivial solution to $||\fobj^T Z_k(\fresp_\opt{I}) \Dif\fresp_\opt{I} || = 0$, independently of the realization $\opt{I}$ and its target objective.
Thus, we introduce an additional constraint filtering out trivial solutions. Instead of solving Opt.~(\ref{eq:minimize_grad}), it is then solved
\begin{equation}
\label{eq:reformulation}
\argmin_{\fobj\in\Omega} ||\fobj^T Z_k(\fresp_I) \Dif\fresp_I||^2\quad\text{s.t.}\quad Z_k(\fresp_I)\fobj\in C
\end{equation}
Here, $C\subseteq\real^{n_f}$ denotes a freely selectable subspace not dependent on $I$, which we call \emph{critical space}.
By setting
\begin{equation}
\label{eq:critical_space}
C \defeq \left(\bigcap_{I\in\mathcal{I}}\coker\Dif\fresp_I\right)^\perp,
\end{equation}
the previously described degeneracy can be avoided.
Since the constraint is linear in $\fobj$, we can find a length preserving substitution $\fobj = U\sigma$ that reduces Opt.~(\ref{eq:reformulation}) to solving $\argmin_{||\sigma|| = 1} || \sigma^T U^T Z_k(\fresp_I) \Dif\fresp_I||^2$ (see supp. material).
The solution (up to a sign) is given by the left\=/singular vector to the smallest singular value of $M \defeq U^T Z_k(\fresp_I) \Dif\fresp_I$.
Thus, first a single forward and $n_f$ backward passes are run to determine $\fresp_I$ and $\Dif\fresp_I$.
Then, $MM^T\in\real^{n_f \times n_f}$ is computed.
The eigenvalue decomposition of $MM^T$ yields the desired left\=/singular vector of $M$.

\subsection{Identifying the Critical Space}

Since we do not make any assumptions about the architecture of $N$, the network's state (i.e., weights, biases, etc.) cannot be used to deduce $C$.
Instead, we sample $n$ random target objectives $\fobj_i$ ($i\in\{1,\dots,n\}$), compute realizations $\opt{I_i}$ by applying \ac{FV}, and then try to approximate $C$ by investigating the matrices $\Dif\fresp^{(i)}$.
We denote $\Dif\fresp_\opt{I_i}$ by $\Dif\fresp^{(i)}$ and $\fresp_\opt{I_i}$ by $\fresp^{(i)}$.
In practice, the matrices $\Dif\fresp^{(i)}$ will not develop clear co\=/kernels, due to numerical inaccuracy, incomplete training runs, or stochastic optimization, just to mention a few reasons.
Hence, special care has to be taken when designing an algorithm to determine $C$.
Instead of searching for co\=/kernels, we search for a subspace $C$ of which its vector-matrix products with the sampled matrices $\Dif\fresp^{(i)}$ diminish, i.e., the spectral norm quotients $||C^T\Dif\fresp^{(i)}||_2 / ||\Dif\fresp^{(i)}||_2$ have to become small.

\algorithmCriticalSpace

Algorithm~\ref{alg:cokernel} approximates Eq.~(\ref{eq:critical_space}).
First, the co\=/kernels are roughly approximated by computing a subspace of left\=/singular vectors with reasonable small singular values $<\rho\cdot||\Dif\fresp^{(i)}||_2$ for each sample.
Afterwards, two subspaces at a time are merged until one is left over.
The merging is implemented by computing the principal angles and vectors as suggested by \textcite{Knyazev2002principal}, dropping all vectors with principal angles of $45^\circ$ or higher and then applying \ac{slerp} to each pair of corresponding left and right principal vectors.
Thereby, the interpolation parameter is given by the ratio of subspaces that already have been merged into either of the two spaces to merge.
The set of newly acquired vectors form a basis of the merged space.
It can be interpreted as a roughly approximated intersection of two spaces, with the exact intersection being obtained when dropping all principal vectors with principal angles not exactly $0^\circ$.
Note that the result of Algorithm~\ref{alg:cokernel} depends both on choices for the singular value threshold as well as the order in which subspaces are merged.

In a perfect world, solving Opt.~(\ref{eq:reformulation}) for a sample $\opt{I_i}$ would yield a projected target objective $\widetilde{\fobj}_i\in\Omega\cap Z_k(\fresp^{(i)})^{-1}C$ such that $Z_k(\fresp^{(i)})\widetilde{\fobj}_i$ points into the same direction as $Z_k(\fresp^{(i)})\fobj_i$ projected onto $C$.
It is uniquely determined by normalizing $Z_k(\fresp^{(i)})^{-1}C_MC_M^TZ_k(\fresp^{(i)})\fobj_i$.
This means that the realization's target objective $\fobj_i$ can only be recovered modulo a shift in $(Z_k(\fresp^{(i)})^{-1}C)^\perp$ direction followed by a renormalization.
One cannot expect to be any better since the objective function parametrization $\fobj \mapsto f_\fobj(\opt{I_i})$ is (by construction of $C$) constant on $(Z_k(\fresp^{(i)})^{-1}C)^\perp$\=/shifts.
In terms of maximum likelihood estimation, one still finds a solution for $\argmax_{\fobj} f_Y(\mathbf{I};\fobj)$.
However, it may be another point on the plateau of the graph of $f_Y(\mathbf{I};\cdot)$ where the realization's target objective one started with resides as well.

\subsection{The Adversary - Aggregation}
\label{sec:aggregation}

Common choices for $\agg$ ~\textcite{olah2017feature} either average along the activations of all neurons associated with a feature, or pick a single representative activation---for instance the center pixel of a feature map. Both approaches have limitations.
When aggregating over activations returned by a convolution filter such as a Sobel filter, except for border pixels, all contributions of pixels being convolved cancel out.
Any filter would thus degenerate to a detector of color patches plus some border pattern.
More general, aggregating along a dimension would eliminate the effect of a preceding linear operation in this dimension.
When picking only a single neuron's activation as representative, the realization returned by \ac{FV} becomes sensitive to its receptive field, rendering realizations of different layers or network architectures incomparable.
Although this does not pose an immediate issue for \ac{IFV} itself, it counteracts our main motivation of using \ac{IFV} techniques for network comparison later on.

To overcome these limitations, we propose to use a max\=/pooling operation followed by mean\=/aggregation.
While the latter operation is independent of the receptive field, the former breaks linearities and allows us to still distinguish between different convolution filters.
Note that a basic mean operation may also be sufficient if the activations returned by $N$ are generated by a \ac{relu} or max\=/pooling operation.
We suggest to always include operations breaking linearity in the aggregation process.

\section{Results and Evaluation}

In the following we analyze the accuracy of \ac{Grad-IFV} and how the recovered target objectives differ from feature responses.
\ac{FV} is applied to three networks with different topology, size and application domains, to generate realizations for randomly sampled target objectives $\fobj_i$.
Given these realizations, Opt.~(\ref{eq:reformulation}) is solved to yield the solutions $\opt{\fobj_i}$, which are then compared to the target objectives $\fobj_i$.
We will subsequently call the solutions $\opt{\fobj_i}$ \emph{predicted objectives}, since they are returned by \ac{Grad-IFV} and are supposed to be an accurate estimate for the target objectives.

\subsection{Network Architectures}

{\bfseries GoogLeNet:} GoogLeNet \cite{szegedy2015going} builds upon stacked Inception modules, each of which takes the input of the previous layer, applies convolution filters with kernel sizes ($1\times 1$, $3\times 3$, and $5\times 5$), and concatenates all resulting feature maps into a single output vector.
GoogLeNet has been trained for image classification, 
its input domain is given by $224\times224$ RGB\=/images.

{\bfseries DenseNet:} The DenseNet-BC architecture \cite{huang2017densely}, with growth rate $k=12$, has been trained on the CIFAR-10 subset of the 80 million Tiny Images dataset \cite{torralba2008tinyimages, krizhevsky2009learning}.
Its input domain is given by $224\times224$ RGB\=/images that are partitioned into 10 different classes.
Three dense blocks are connected via convolution layers, followed by max pooling, form the core part of the network.
Within a dense block, each layer takes the outputs of \emph{all} preceding layers as inputs and processes them by applying a $1\times1$\=/bottleneck convolution, batch normalization, \ac{relu} activation, and finally a $3\times3$\=/convolution.
The output is passed to all subsequent layers of the dense block.
In total, 0.8M weights are trained with weight decay by stochastic gradient descent with nesterov momentum.
During training, images are flipped, padded with 4 pixels on each side, and cropped back to its original size with a random center.

{\bfseries SRNet:} The fully convolutional SRNet~\cite{sajjadi2018frame} has been modified and trained to upscale low resolution geometry images (normal and depth maps) of isosurfaces in scalar volume data by \textcite{weiss2019volumetric}.
The low resolution input size is set to $128\times 128$ pixels. Residual blocks of $3\times 3$ convolutions transform the input maps into a latent space representation, which is then upsampled, folded, and added as residual to bilinearly upscaled versions of the input. The input domain is given by a 2D normal field comprised of 3D vectors, a 2D depth map with values in the range $[0,1]$, and a 2D binary mask with values in $\{-1, 1\}$ to indicate surface hits during rendering.

\subsection{Feature Visualization}

If not stated otherwise, the components of the objective function $f_\fobj$ used for \ac{FV} are as follows: The network function $N$ applies the network up to the investigated layers (that is \texttt{inception4c}, \texttt{dense3} etc.) and outputs their respective activations as a 3D-tensor with 2 spatial and 1 channel dimension.
The aggregation operation $\agg$ then applies 2D max pooling with kernel size $3\times 3$ and a stride of $1$ followed by taking the mean.
Both operations are performed along the spatial dimension so that one ends up with a 1D vector of ,,channel activations'' that act as our feature response.
Here, each channel constitutes a separate feature.
The feature response is combined with the target objective $\fobj$ as described in Sec.~\ref{sec:method}. The power $k$ of the cosine term is set to $2$.
Last but not least, the used parametrization $P_\text{rgb}$ yields batches of RGB images clamped to values in $[0, 1]$ by applying the sigmoid function to the unbounded space $\real^{H\times W\times 3}$, where $H$ and $W$ denote the respective input image dimensions for the investigated network.

The target objectives $\fobj$ with nonnegative entries are
sampled from a vector distribution $X$ with uniform Hoyer sparseness measure (i.e. $\hoyer(X) \sim \mathcal{U}\{0,1\}$) \cite{hoyer2004non}.
After sampling $\fobj$, we optimize for $\argmin_{v\in\real^{n_p}} (f_\fobj \circ P)(v)$ by first applying Adam optimization \cite{kingma2014adam} followed by some steps of fine tuning with L\=/BFGS \cite{byrd1995lbfgs}.
For DenseNet and SRNet, we run 800 steps of Adam and 300 steps of L\=/BFGS. To let GoogLeNet converge to an optimum, we run 3000 steps of Adam followed by 500 steps of L\=/BFGS.

\subsection{Time Complexity \& Performance}

The runtime complexity of solving Opt.~(\ref{eq:reformulation}) and Algorithm~\ref{alg:cokernel} is $\bigO(\max(n_p n_f^2, n_f^3))$ per sample. Runtime is dominated by singular value decompositions and multiplication of $n_f \times n_f$ and $n_f \times n_p$ matrices.
The computation of $\Dif\fresp$ requires $n_f$ backward passes through the network, making it strongly dependent on the network architecture.
Performance measurements are performed on a server architecture with 4x Intel Xeon Gold 6140 CPUs with 18 cores @ 2.30GHz each, and an NVIDIA GeForce GTX 1080 Ti graphics card with 11 GB VRAM. Timings are shown in Table~\ref{table:timings}.
Note that the generation of realizations takes significantly longer due to the additional L\=/BFGS steps.
Except for Algorithm~\ref{alg:cokernel}, all computations are executed on the GPU.

\subsection{The Simple Case}
\label{sec:gnet_intro}

In the first experiment, the 512 filters in the \texttt{inception4c} layer of GoogLeNet, which we abbreviate by GN4c, are investigated.
We sample 150 target objectives $\fobj_i$ as described above and add further 150 canonical vectors of sparsity measure 1.
Then, \ac{FV} is performed.
For each resulting realization $\opt{I_i}$, a target objective $\opt{x_i}$ is predicted by solving Opt.~(\ref{eq:reformulation}), and then compared to $\fobj_i$ by measuring their angular distance in degrees, i.e., $180/\pi\cdot \min\{\cos^{-1}(-\fobj_i^T\opt{x_i}), \cos^{-1}(\fobj_i^T\opt{x_i})\}$.
The elimination of trivial solutions in not required yet, hence we set $C = \real^{n_f}$.

\tableTimings{t}

The horizontal margin distribution in Fig.~\ref{fig:joint_coef_plots}a shows the angular distances between the target objectives and either the predicted objectives $\opt{x_i}$, feature responses $\fresp^{(i)}$, or randomly sampled vectors.
Note that whereas the angular distance between $\fobj_i$ and $\fresp^{(i)}$ is about $45^\circ$ (or a cosine similarity of $\approx 0.7$) with large variance, the predictions $\opt{x_i}$ show only a deviation of approx. $10^\circ$ ($\approx 0.98$ in terms of cosine similarity!) with very few outliers.

\figureCoefJoint{t}

Next, it is verified that the realizations are optimal w.r.t.\ the predicted objective.
To achieve this, a realization is re\=/optimized w.r.t.\ either $\fobj_i$, $\opt{x_i}$, $\fresp^{(i)}$, or a random objective with Adam for 500 steps.
Finally, the \ac{ssim} index between the realization before and after optimizing is computed.
\ac{ssim} is a perception\=/based model to measure the similarity of two images w.r.t.\ structural information \cite{wang2004ssim}.

The results are shown in Fig.~\ref{fig:joint_coef_plots}a. High \ac{ssim} indices confirm that almost all realizations remain optima under the predicted objectives, yet there is clearly more change happening when re\=/optimizing for the feature response as objective.
This observation confirms the rationale underlying the Deep\=/Dream process \cite{mordvintsev2015inceptionism} that enhancing features changes the input significantly.
In particular, in many cases re\=/optimizing for the feature response yields similar results to re\=/optimizing for a random vector.

Note that the \ac{ssim} index is strictly lower than one when re\=/optimizing for the target objective.
This observation relates back to the Adam optimizer having an internal state that has to warm up first before Adam convergences.
Hence, even when initializing Adam with an instable, local optimum, Adam may leave it and converge to a completely different one, lowering the \ac{ssim} index significantly.

\subsection{Utilizing the Critical Space}
\label{sec:cs_analysis}

Next, we investigate the ten neurons in the \texttt{classification} layer of DenseNet for two differently trained networks.
Since aggregation along spatial dimensions becomes unnecessary in this case, we set $\agg \equiv \id$.
The second network is trained in a similar way than the first network described above, yet it never sees any training data for class 4. The \texttt{classification} layers of both networks are denoted by DNcl and DN4cl respectively.
We sample 290 target objectives and add the ten canonical ones.

When predicting objectives and comparing them to the target objectives $\fobj_i$ as in the previous section, \ac{Grad-IFV} fails.
The experiment represented by blue dots in Fig.~\ref{fig:joint_coef_plots}b,c show that the predicted objectives do not resemble the target objectives.
To see how close the target objective is to be an optimal solution for Opt.~(\ref{eq:reformulation}), we decompose each $\fobj_i$ as follows: Let $v_{i,1},\dots,v_{i,n_f}$ be the left\=/singular vectors of $Z_k(\fresp^{(i)}) \Dif\fresp^{(i)}$ with singular values $\sigma_{i,1}\leq\sigma_{i,2}\leq\dots\leq\sigma_{i,n_f}$. Then we determine coefficients $c_{i,j}$ such that $\fobj_i = \sum_j c_{i,j} v_{i,j}$.
Finally, we average along all squared coefficients of the same order and obtain values $\alpha_j = \sum_{i=1}^n (c_{i,j}^2 / n)$ that indicate how much the $j$\=/th left\=/singular vectors contribute to the $\fobj_i$s.
Note that $c_{i,0} = \fobj_i^T\opt{x_i}$ and that $90(1-c_{i,0}^2)$ is the polynomial of degree two approximating the angular distance between $\fobj_i$ and $\opt{x_i}$ with coincidence in $c_{i,0}^2 \in [0, 0.5, 1]$.

Since \ac{Grad-IFV} always predicts the left\=/singular vector of $Z_k(\fresp) \Dif\fresp$ to the lowest singular value, it performs well when $\alpha_1 \approx 1$ and $\alpha_j \approx 0$ for $j>1$.
As seen in Fig.~\ref{fig:joint_coef_plots}a, $\alpha_1 \approx 1$ holds true for the scenario of Sec.~\ref{sec:gnet_intro}.
However, the horizontal bars of Fig.~\ref{fig:joint_coef_plots}b,c indicate additional contributions indicated by significant values for $\alpha_2$ and $\alpha_3$, respectively.
When further investigating left\=/singular vectors, we realized that all contributions of $\alpha_1$ (and $\alpha_2$ for DN4cl) relate back to the same span of left\=/singular vectors for all samples. We consider these vectors to be trivial solutions of Opt.~(\ref{eq:reformulation}), which we intend to sort out.

To obtain the trivial solutions and their complement, the critical space, the first 32 of the 300 Jacobian matrices $\Dif\fresp^{(i)}$ are passed to Algorithm~\ref{alg:cokernel}.
We apply a nested intervals technique to determine all values of $\rho\in(0,1)$ that yield critical spaces of different dimensions.
For DNcl, a single 9\=/dimensional critical space $C_1$ is determined, of which the complement is spanned by (approximately) $(1,\cdots,1)^T\in\real^{10}$.
The same critical space $C_1$ is retrieved for DN4cl, accompanied by a second one $C_2$ of 8 dimensions with its complement being spanned by $(1,\cdots,1)^T$ and $e_4$.
Presumably, $(1,\cdots,1)^T$ arises as trivial solution because the network learned to exploit the softmax translation invariance in order to improve on a regularization term on its weights.
Similarly, $e_4$ relates back to a dead neuron of a class the network of DN4cl never has seen.

The solutions $\opt{x_i}$ of Opt.~(\ref{eq:reformulation}) are computed by setting $C$ to either the full 10\=/dimensional ambient space, $C_1$, or $C_2$.
We evaluate how closely the predicted objectives resemble the target objectives by projecting both onto $(Z_k(\fresp^{(i)})^{-1}C)^T$ and computing their angular distance.
The distribution of angular distances, which is shown in the horizontal margin plots of Fig.~\ref{fig:joint_coef_plots}b,c, indicate that \ac{Grad-IFV} performs well for one particular critical space ($C_1$ for DNcl and $C_2$ for DN4cl).
In practice, the correct one can be identified by running the described procedure for a set of known, i.e. sampled, target objectives $\fobj$, choosing the best performing critical space, and then solving Opt.~(\ref{eq:reformulation}) for realizations $\opt{I_i}$ with unknown target objective.

When applying re\=/optimization, Fig.~\ref{fig:joint_coef_plots}b,c indicates that all samples with high angular distance to the target objective exhibit a low \ac{ssim} index.
In particular, target objectives cannot be recovered accurately for samples where \ac{FV} failed to find a stable optimum (marked by triangles in Fig.~\ref{fig:joint_coef_plots}b,c).
Further, there exists a cluster of (stable) samples for which the \ac{ssim} index is low although the prediction is accurate. All samples in the cluster have in common that a notable fraction of their corresponding target objectives is located in the subspace $(Z_k(\fresp^{(i)})^{-1}C)^\perp$, which we just factored out.
This observation gives evidence that, in practice, the objective function parametrization $\fobj\mapsto f_\fobj$ is not entirely constant for $(Z_k(\fresp^{(i)})^{-1}C)^\perp$\=/shifts, as it would be if $C$ is chosen to be the exact union of co\=/kernels as expressed in Eq.~(\ref{eq:critical_space}).
Thus, whenever we eliminate outliers by dropping all predicted objectives with a low \ac{ssim} value, we loose some relevant results as well.

\subsection{Dropping the Cosine Term}

When experimenting with different numbers of features $n_f$ and alternative aggregation functions, we observed that---as long as we keep the cosine term of the significance measure $S_\fobj$ with a power of $k=2$---\ac{Grad-IFV} is quite resilient.
When increasing the number of features or exchanging the aggregation function by averaging or picking as described in Sec.~\ref{sec:aggregation}, we notice only slight increases in angular distances with prediction quality being similar to the results of Sec.~\ref{sec:gnet_intro}.
Except for the very few occasional outliers which can be filtered out by re\=/optimizing the input w.r.t.\ the predicted objective and investigating the \ac{ssim} index, the target objective is reliably recovered.

However, when running the experiments again with the cosine term dropped ($k=0$), predictions may become unreliable.
When only considering the first $160$ convolution filters in the \texttt{dense3} block of DenseNet, which we call DNde from now on, predicted objectives still perform significantly better than just estimating $\fobj_i$ via the feature response $\fresp^{(i)}$.
The \ac{ssim} index obtained by re\=/optimization does not drop below $0.89$.
For more features, prediction quality starts to degenerate quickly.
When considering more than $288$ of the $342$ available filters in DNde, almost all angular distances between $\opt{x_i}$s and $\fobj_i$s are in the range from $70^\circ$ to $90^\circ$.
When increasing the feature count in the \texttt{inception4c} layer of GoogLeNet (GN4c), degradation starts to set in for $96$ features and continues up to $256$. Beyond, most predicted objectives are perpendicular to the target objective.

As discussed in Sec.~\ref{sec:cs_analysis}, the target objective can be expressed as a linear combination of singular vectors, to investigate by which margin \ac{Grad-IFV} fails to extract the correct prediction.
The resulting $\alpha_j$\=/values for different feature counts $n_f$ are depicted by stacked bars in Fig.~\ref{fig:drop_cosine}.
The prediction quality of \ac{Grad-IFV} can be assessed via the height of the lowest bar, which shows the value of $\alpha_1$.
High fragmentation of the bar charts suggests that target objectives cannot solely be recovered by setting up an appropriate critical space---at least not by one perpendicular to singular vectors of small singular values as returned by Algorithm~\ref{alg:cokernel}.

\figureDropCosine{t}

Next, we consider the first $80$ filters of DNde and exchange the aggregation function. The coefficient histograms of Fig.~\ref{fig:drop_cosine} show that \ac{Grad-IFV} fails to extract meaningful predictions for averaging- and picking\=/aggregations, although the results for max pooling followed by averaging are close to perfect.

Interestingly, when the experiments are conducted with $k=2$---i.e.\ with the cosine term---accurate predictions are obtained. In this case, \ac{FV} has limited options in order to produce a visualization so that objectives pointing into perpendicular direction of $\fresp$ can be ignored during \ac{IFV}.
In Sec.~\ref{sec:reformulation}, the matrix $Z_k(\fresp)$ shrinks all vectors in the input's feature response direction and thus makes them more likely to be the left\=/singular vector to the smallest singular value.

\subsection{Different Parameterizations}

We further investigate the influence of the parametrization $P$ on \ac{Grad-IFV}.
Therefore, we define the variables $v$ over which \ac{FV} optimizes in Fourier space and perform spatial de\=/correlation \cite{olah2017feature}, i.e., $P_\text{fft}$ is set to the \ac{ifft} followed by sigmoid clamping.
A parametrization favoring low frequencies can be obtained by scaling Fourier coefficients according to their frequency energies before applying \ac{ifft}, denoted by $P_\text{ffte}$.
For each parametrization $P_\text{rgb}$, $P_\text{fft}$ and $P_\text{ffte}$,
\ac{Grad-IFV} is applied to GN4c and DNde.
Although the image quality of the resulting realizations change notably (see Fig.~\ref{fig:realizations_for_params}), the quality of predicted objectives is not influenced.
In particular, results do not suffer from low image quality of \ac{FV} in case no regularizations such as transformation robustness are used.

Lastly, we analyze the SRNet for upscaling geometry images. We parameterize the normal map by coordinates $(\varphi, \theta)$, yet $\varphi$ is not bound to an interval length of $2\pi$ so that warping gradients at interval borders can be avoided.
To ensure that the $z$\=/component is positive, we restrict $\theta$ to the interval $[-\pi/2, \pi/2]$ by applying a sigmoid function followed by an appropriate affine transformation.
The depth map is clamped to $[0, 1]$ with the sigmoid function.
The binary mask is either hard\=/coded to be one or parameterized continuously in the interval $[-1, 1]$.
Note that we cannot properly represent the mask by a discrete set of values since this would turn the optimization problem of \ac{FV} (Opt.~(\ref{eq:activation_maximization})) into a mixed-integer programming problem.

When investigating the 64 filters of the 8th residual block of SRNet, \ac{Grad-IFV} always recovers the 64 sampled target objectives accurately up to an angular distance of $5^\circ$.
We even obtain reliable results with angular distances $<20^\circ$, when dropping the cosine term of the objective function.

\section{Limitations \& Future Work}

One of our major goals is to identify convolution filters with similar feature visualizations along different networks.
Under the assumption that feature visualizations carry semantic information about what a network learns, such an approach can eventually raise network comparison from mere signal analysis of activations to a semantic level.
\ac{IFV} enables selecting and visualizing a feature using one network first, and then inverting the resulting realization using a second network.
This yields two features obeying the same visualization and, thus, representing the same semantic concept.

\figureRealizationsForParams{t}

In real scenarios, however, a realization that is optimal for one network will not be so for another network. It might not even be close to an optimum as long as the sensitivity of the objective function to high frequencies and noise is not reduced.
In particular, \ac{Grad-IFV} relies on an input that is close to optimal, since a gradient's magnitude, in general, cannot reflect how distant an input is to an optimal solution in the surrounding.

Therefore, in future work we will consider to widen the scope of objective functions to include arbitrary input priors (such as in \textcite{mahendran2015understanding} or \textcite{nguyen2016synthesizing}).
Furthermore, we intend to integrate concepts that facilitate the processing of visualizations that are robust w.r.t.\ transformations \cite{olah2017feature}.
Both, input priors and transformation robustness, are key techniques to generate consistent and interpretable visualizations.
By considering such techniques in \ac{IFV}, we hope to achieve feature predictions that are less sensitive to  network variations or input noise.


Upon resolving these issues, \ac{IFV} can be used for network comparison, to analyze learned representations of networks trained on different datasets.
Here it will be interesting to investigate which features are shared between two networks, and
whether invariant operations on a network's weight space such as permuting or rescaling neurons can be recovered.

\section{Conclusion}

We introduce the problem of identifying target objectives under which \acf{FV} yields a certain input, and propose a solution for certain types of \ac{FV} objective functions.
We demonstrate that the (possibly unknown) target objective can be accurately approximated by performing a singular value decomposition of a modified version of the network's Jacobian matrix.
In cases where the Jacobian matrix is ill\=/behaving, we identify problematic subspaces and factor them out to obtain accurate results modulo the reduction.
In a number of experiments we investigate the accuracy by which the target objective is recovered, and whether the input remains stable under the predicted objective, i.e., the result truly is an inverse of \ac{FV}.
We observe that different choices for layer size, aggregation and objective functions can have a significant impact on the proposed technique.
Finally, we envision future research directions towards feature\=/based comparison of learned representations---including assessment of the relevance of patterns for successful network training---and the use of Inverse Feature Visualization for network comparison.



\bibliography{refs}
\bibliographystyle{icml2020}

\end{document}


\twocolumn[
\icmltitle{Inverting the Feature Visualization Process for Feedforward Neural Networks\\Supplementary Material}

\icmlsetsymbol{equal}{*}

\begin{icmlauthorlist}
	\icmlauthor{Christian Reinbold}{cgtum}
	\icmlauthor{R\"udiger Westermann}{cgtum}
\end{icmlauthorlist}

\icmlaffiliation{cgtum}{Chair of Computer Graphics and Visualization, Technical University of Munich, Bavaria, Germany}

\icmlkeywords{Machine Learning, ICML, Feature Visualization}

\vskip 0.3in
]

\printAffiliationsAndNotice{}

This document accompanies our paper \emph{Inverting the Feature Visualization Process for Feedforward Neural Networks}. It contains additional material that could not be included in the manuscript due to page restrictions. In particular, we provide
\begin{itemize}
	\item a mathematically rigor formulation of the transformation of $\nabla f_\fobj$ (Sec.~\ref{sec:reformulat}),
	\item additional details on how the target objective is sampled (Sec.~\ref{sec:sample}),
	\item visual differences in realizations when re\=/optimizing,
	\item additional results of all performed experiments (Sec.~\ref{sec:results}).
\end{itemize}

\section{Reformulation of $\nabla f_\fobj$ - Calculations}
\label{sec:reformulat}

In our work, we derive the gradient $\nabla f_\fobj\circ P$ for solving
$$
\argmin_{\fobj\in\Omega} ||\nabla (f_\fobj\circ P)(\opt{I})||^2.
$$
In the following, we outline the calculations to compute the gradient.
Let us recall that $f_\fobj\circ P = S_\fobj \circ \net$, where $S_\fobj$ is defined as $S_\fobj(\fresp) = \fobj^T\fresp\cdot (\fobj^T\fresp/||\fresp||)^k$ for fixed $k\in\natural_0$ and $\net$ is a function returning the feature response $\fresp_I$ for an input parametrization $v_I$.
With $\overline{\fresp} \defeq \fresp / ||\fresp||$ and $q(\fresp) \defeq \fobj^T\overline{\fresp}$, one first obtains
\begin{align*}
\frac{\partial q(\fresp)}{\partial \fresp_i} &= \frac{\frac{\partial}{\partial \fresp_i}(\fobj^T\fresp) \cdot ||\fresp|| - \frac{\partial}{\partial \fresp_i}(||\fresp||) \cdot \fobj^T\fresp}{||\fresp||^2} \\
&= \frac{\fobj_i \cdot ||\fresp|| - \overline{\fresp}_i \cdot \fobj^T\fresp}{||\fresp||^2} = (\fobj_i - q(\fresp) \cdot \overline{\fresp}_i) / ||\fresp||,
\end{align*}
that is $\nabla q(\fresp)^T = (\fobj - q(\fresp) \cdot \overline{\fresp}) / ||\fresp||$.
By applying the chain rule, for $k>0$ we obtain 
\begin{align*}
\nabla S_\fobj(\fresp)^T &= \frac{\partial}{\partial \fresp}(\fobj^T\fresp\cdot q(\fresp)^k) \\
 &= q(\fresp)^k\cdot \fobj +  \fobj^T \fresp \cdot kq(\fresp)^{k-1} \cdot \nabla q(\fresp)^T \\
 &= q(\fresp)^k\cdot \fobj + kq(\fresp)^k \cdot (\fobj - q(\fresp)\cdot\overline{\fresp}) \\
 &= (k+1)q(\fresp)^k (\fobj - \frac{k}{k+1}\fobj^T\overline{\fresp}\cdot\overline{\fresp}) \\
 &= (k+1)q(\fresp)^k (\fobj - \frac{k}{k+1}\overline{\fresp}\cdot\overline{\fresp}^T\fobj) \\
 &= (k+1)q(\fresp)^k (I - \frac{k}{k+1}\overline{\fresp}\overline{\fresp}^T)\fobj
\end{align*}
We then set
$$
Z_k(\fresp) = Z_k(\fresp)^T \defeq \id - k/(k+1)\overline{\fresp}\overline{\fresp}^T
$$
(as in the paper) to obtain
$$
\nabla S_\fobj(\fresp)^T = (k+1)q(\fresp)^k Z_k(\fresp)\fobj.
$$
Again by the chain rule, it follows that
\begin{align*}
\nabla (f_\fobj\circ P)(v_I) &= \nabla (S_\fobj \circ \net)(v_I) \\
&= \nabla S_\fobj(\net(v_I)) \Dif\net(v_I) \\
&= (k + 1)q(\net(v_I))^k x^T Z_k(\net(v_I)) \Dif\net(v_I)
\end{align*}
Denoting $\net(v_I)$ by $\fresp_I$ and the Jacobian matrix $\Dif\net(v_I)$ by $\Dif\fresp_I$, this expression simplifies to
$$
\nabla (f_\fobj\circ P)(v_I) = (k + 1)q(\fresp_I)^k x^T Z_k(\fresp_I) \Dif\fresp_I.
$$

For $k = 0$, it holds that $\nabla (f_\fobj\circ P)(v_I) = x^T \Dif\fresp_I$.
Because $Z_0$ evaluates to the identity matrix, independently of $\fresp_I$, the reformulation
{
	\addtocounter {equation}{-1}
	\renewcommand{\theequation}{3} 
	\begin{equation}
	\label{eq:reformulation}
	\argmin_{\fobj\in\Omega} ||\fobj^T Z_k(\fresp_I) \Dif\fresp_I||^2\quad\text{s.t.}\quad Z_k(\fresp_I)\fobj\in C
	\end{equation}
}
is valid for $k = 0$ as well.

For $k\rightarrow\infty$, the matrix $Z_k(\fresp_I)$ converges to a projection matrix that introduces $\fresp_I$ in the co\=/kernel of $Z_k(\fresp_I) \Dif\fresp_I$.
Hence, the solution to Opt.~(\ref{eq:reformulation}) converges to the normalized feature response $\overline{\fresp}_I$.
This behavior is consistent with the observation that the significance measure $S_\fobj$ pushes the feature response closer to the feature objective $\fobj$ when increasing $k$.

\subsection*{Incorporating the Linear Constraint}

Opt.~(\ref{eq:reformulation}) can be further simplified by integrating the linear constraint $Z_k(\fresp_I)\fobj\in C$ into the objective function via substitution. Let $C_M\in\real^{n_f \times \dim C}$ be a matrix such that its columns form an orthonormal basis of $C$.
Then, $\fobj$ satisfies the constraint in Opt.~(\ref{eq:reformulation}) iff $\fobj\in \im Z_k(\fresp)^{-1}C_M$.
Note that $Z_k(\fresp)^{-1} = \id + k \cdot \overline{\fresp}\overline{\fresp}^T$ always exists by the Sherman-Morrison formula.
Since $Z_k(\fresp)^{-1}C_M$ has full column\=/rank, the \ac{svd} $U \Sigma V^T = Z_k(\fresp)^{-1}C_M$ yields an orthogonal matrix $U$ with $\im U = \im Z_k(\fresp)^{-1}C_M$.
The substitution of $\fobj$ then is given by $\fobj = U \sigma$, for $\sigma\in\real^{\dim C}$ with $||\sigma|| = 1$.
Opt.~(\ref{eq:reformulation}) reduces to
$$
\argmin_{||\sigma|| = 1} || \sigma^T U^T Z_k(\fresp) \Dif\fresp||^2.
$$

\section{Sampling Strategy for Target Objectives}
\label{sec:sample}

When selecting the target objectives, we noticed that uniformly sampling the $nD$\=/sphere $\Omega$ does not yield representative target objectives.
Due to the curse of dimensionality, the density of sparse target objectives, especially canonical objectives $e_i$ representing single features, rapidly tends to $0$ even for medium sizes of $n$ (see Fig.~\ref{fig:sampling_sparseness}).
\figureSamplingSparseness{t}
However, sparse target objectives are usually those which are investigated in \acsu{FV}.
Hence, we implement a sampling strategy as follows:
First, a scalar value $s$ in $[0, 1]$ is sampled from a uniform distribution. It indicates how sparse the target objective $\fobj$ is supposed to be.
We use the Hoyer sparseness measure \cite{hoyer2004non} to measure sparseness of $\fobj$. It measures the ratio of $||x||_1$ and $||x||_2$, and then applies an affine transformation such that the smallest possible ratio of 1:1 (for canonical vectors) maps to $1$ and the largest possible ratio of $\sqrt{n}$:1 (for vectors with all components being equal) maps to 0.
Second, we (uniformly) sample a random vector $\fobj_0$ from $\Omega$ and run curvilinear search \cite{Wen2013feasible} with $\fobj_0$ as initial value to find a vector $\fobj\in\Omega$ that minimizes $(\hoyer(\fobj) - s)^2$.

When experimenting with an optimization based approach of finding critical spaces (which we dropped due to the lack of reliability), we observed that introducing sampling as above highly increases the chance of finding the global optimum.
If target objectives are sampled uniformly from $\Omega$, the optimization process commonly gets stuck in a slightly worse than globally optimal solution $\opt{C}$ that does not generalize well to sparse target objectives.
That is, the solution of Opt.~(\ref{eq:reformulation}) w.r.t.\ the critical space $\opt{C}$ matches the projected target objective $\widetilde{\fobj}$ when providing a realization of a sampled target objective, but does not so when providing a realization to a canonical target objective $e_i$.
Since the proposed sampling strategy helps in this regard, we utilize it in our experiments as well to avoid potentially skewed results.

\section{Realizations and Difference maps}
\label{sec:realizations}

In our work, we commonly compare angular distances to \acsu{ssim} indices. Since \ac{ssim} is just one of many potential measures to quantify similarity between images, we additionally provide difference maps in Fig.~\ref{fig:gnet_intro_diff} and argue that \ac{ssim} represents these accurately.
The shown excerpt of GoogLeNet\=/\texttt{inception4c} samples reveals that predicted objectives mostly behave similarly to target objectives in re\=/optimization, whereas feature responses introduce significant high frequency changes of the same magnitude as randomly picked feature vectors.
Note that the outliers produced by our method also show up in the difference maps. One is contained in the excerpt. It corresponds to a blue outlier point of Fig.~2a in the paper that is located next to the red cluster.
We do not provide difference maps for all experiments to keep the size of the supplementary file to reasonable sizes. We refer to the experiment suite provided in the source code submission for creating further imagery.

\bibliography{refs}
\bibliographystyle{icml2020}

\onecolumn

\section{Detailed Results of Experiments}
\label{sec:results}

In the ``Results and Evaluation'' section of our manuscript we perform a number of experiments and summarize the resulting findings.
To further back up these findings, we provide several additional statistics below. For each experiment, we show the following diagrams:
\begin{enumerate}
	\item A stacked bar chart depicting mean contributions $\alpha_j$ of $j$\=/th left\=/singular vectors---ordered by singular values---to target objectives (details see paper).
	\item A \ac{kde} plot showing the distribution of \acsu{ssim} indices when re\=/optimizing samples w.r.t.\ the target objectives. A sample is considered instable if its \ac{ssim} index drops below 0.7 (red, dashed line).
	\item A bivariate plot relating angular distances to target objectives and \acsu{ssim} indices when re\=/optimizing samples w.r.t.\ the predicted objective. Triangular markers correspond to instable samples. If present, point size encodes the fraction of the target objective that lives in the respective critical space.
\end{enumerate}

The following experiments are conducted:

\begin{center}
	\begin{tabular}{rlllcc}
		\toprule
		ID & Network & Features & Aggregation & Cos. term & Param. \\
		
		\midrule
		\multicolumn{6}{l}{\textbf{The Simple Case}} \\
		\midrule
		
		1 & GoogLeNet & 512 filters, \texttt{inception4c} & Max pooling + mean aggr. & $k = 2$ & RGB \\
		
		\midrule
		\multicolumn{6}{l}{\textbf{Utilizing the Critical Space}} \\
		\midrule
		
		2 & DenseNet & 10 neurons, \texttt{classification} & -- & $k=2$ & RGB \\
		3 & DenseNetEx4 & 10 neurons, \texttt{classification} & -- & $k=2$ & RGB \\
		4 & DenseNet & 10 neurons, \texttt{classification} & -- & $k=0$ & RGB \\
		5 & DenseNetEx4 & 10 neurons, \texttt{classification} & -- & $k=0$ & RGB \\
		
		\midrule
		\multicolumn{6}{l}{\textbf{Dropping the Cosine Term}} \\
		\midrule
		
		6 & DenseNet & 32 to 342 filters, \texttt{dense3} & Max pooling + mean aggr. & $k=0$ & RGB \\
		7 & GoogLeNet & 32 to 512 filters, \texttt{inception4c} & Max pooling + mean aggr. & $k=0$ & RGB \\
		8 & DenseNet & 80 filters, \texttt{dense3} & Max pooling + mean aggr. & $k=0$ & RGB \\
		9 & DenseNet & 80 filters, \texttt{dense3} & mean aggr. & $k=0$ & RGB \\
		10 & DenseNet & 80 filters, \texttt{dense3} & pick center neuron & $k=0$ & RGB \\
		11 & DenseNet & 32 to 342 filters, \texttt{dense3} & Max pooling + mean aggr. & $k=2$ & RGB \\
		12 & GoogLeNet & 32 to 512 filters, \texttt{inception4c} & Max pooling + mean aggr. & $k=2$ & RGB \\
		13 & DenseNet & 80 filters, \texttt{dense3} & Max pooling + mean aggr. & $k=2$ & RGB \\
		14 & DenseNet & 80 filters, \texttt{dense3} & mean aggr. & $k=2$ & RGB \\
		15 & DenseNet & 80 filters, \texttt{dense3} & pick center neuron & $k=2$ & RGB \\
		
		\midrule
		\multicolumn{6}{l}{\textbf{Different Parameterizations}} \\
		\midrule
		
		16 & DenseNet & 342 filters, \texttt{dense3} & Max pooling + mean aggr. & $k=2$ & RGB \\
		17 & DenseNet & 342 filters, \texttt{dense3} & Max pooling + mean aggr. & $k=2$ & FFT \\
		18 & DenseNet & 342 filters, \texttt{dense3} & Max pooling + mean aggr. & $k=2$ & FFT-E \\
		19 & GoogLeNet & 512 filters, \texttt{inception4c} & Max pooling + mean aggr. & $k=2$ & RGB \\
		20 & GoogLeNet & 512 filters, \texttt{inception4c} & Max pooling + mean aggr. & $k=2$ & FFT \\
		21 & GoogLeNet & 512 filters, \texttt{inception4c} & Max pooling + mean aggr. & $k=2$ & FFT-E \\
		22 & SRNet & 64 filters, 8-th block & Max pooling + mean aggr. & $k=2$ & $\text{mask} = 1$ \\
		23 & SRNet & 64 filters, 8-th block & Max pooling + mean aggr. & $k=2$ & $\text{mask}\in[-1,1]$\\
		24 & SRNet & 64 filters, 8-th block & Max pooling + mean aggr. & $k=0$ & $\text{mask} = 1$ \\
		25 & SRNet & 64 filters, 8-th block & Max pooling + mean aggr. & $k=0$ & $\text{mask}\in[-1,1]$\\
		\bottomrule
		
	\end{tabular}
\end{center}

\figureGNetIntroImages{t}

\begin{center}
	  \begin{minipage}{0.32\textwidth}\framebox{\begin{minipage}{\textwidth}
				\begin{overpic}[width=\textwidth, tics = 10, trim = 0 0 0 0 , clip]
					{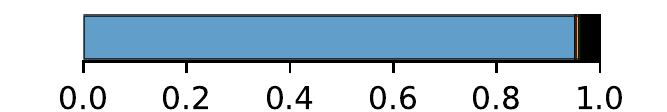}
				\end{overpic}
				\vspace*{2mm}
				\begin{overpic}[width=\textwidth, tics = 10, trim = 0 0 0 0 , clip]
					{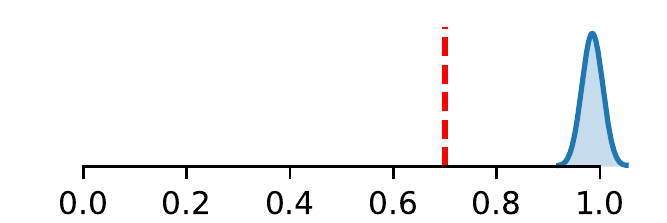}
				\end{overpic}
				\begin{overpic}[width=\textwidth, tics = 10, trim = 0 -10 0 0 , clip]
					{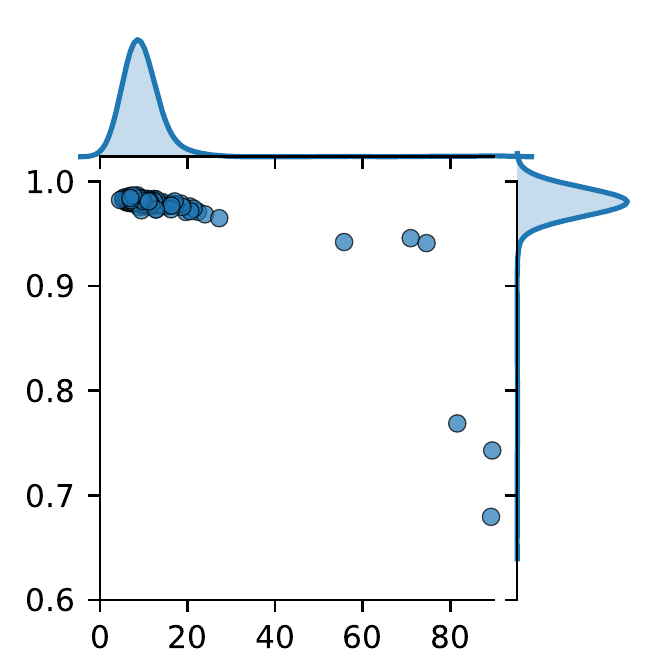}
					\put(0, 0){\textbf{\textcolor{black}{\large (1)}}}
				\end{overpic}
  \end{minipage}}\end{minipage}
  \hfill
  \begin{minipage}{0.32\textwidth}\framebox{\begin{minipage}{\textwidth}
				\begin{overpic}[width=\textwidth, tics = 10, trim = 0 0 0 0 , clip]
					{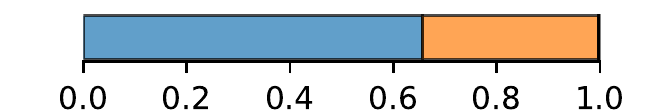}
				\end{overpic}
				\vspace*{2mm}
				\begin{overpic}[width=\textwidth, tics = 10, trim = 0 0 0 0 , clip]
					{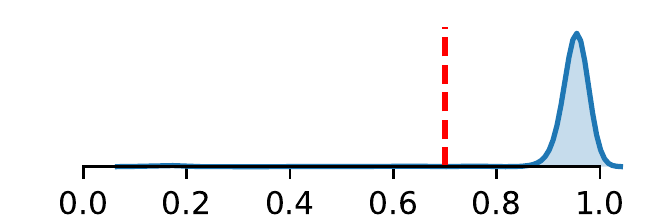}
				\end{overpic}
				\begin{overpic}[width=\textwidth, tics = 10, trim = 0 -10 0 0 , clip]
					{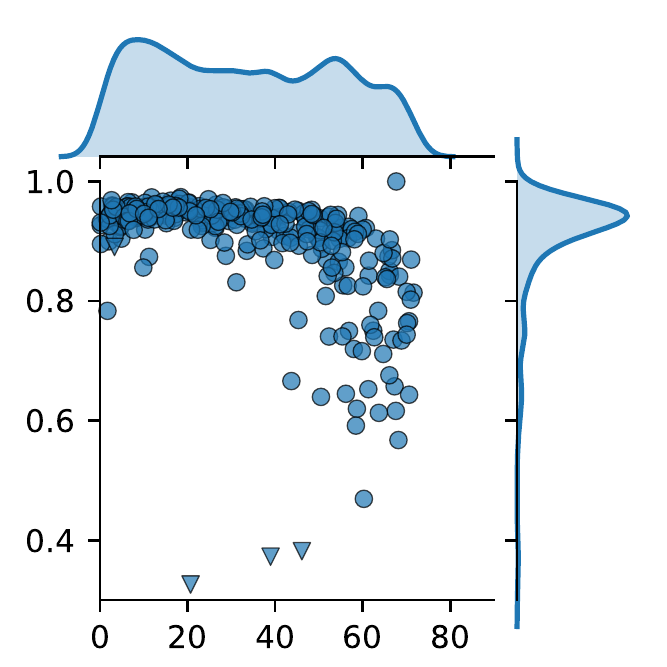}
					\put(0, 0){\textbf{\textcolor{black}{\large (2 - trivial critical space)}}}
				\end{overpic}
  \end{minipage}}\end{minipage}
  \hfill
  \begin{minipage}{0.32\textwidth}\framebox{\begin{minipage}{\textwidth}
				\begin{overpic}[width=\textwidth, tics = 10, trim = 0 0 0 0 , clip]
					{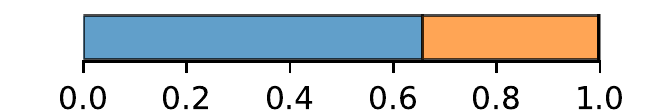}
				\end{overpic}
				\vspace*{2mm}
				\begin{overpic}[width=\textwidth, tics = 10, trim = 0 0 0 0 , clip]
					{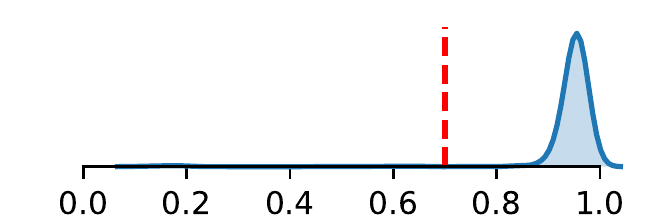}
				\end{overpic}
				\begin{overpic}[width=\textwidth, tics = 10, trim = 0 -10 0 0 , clip]
					{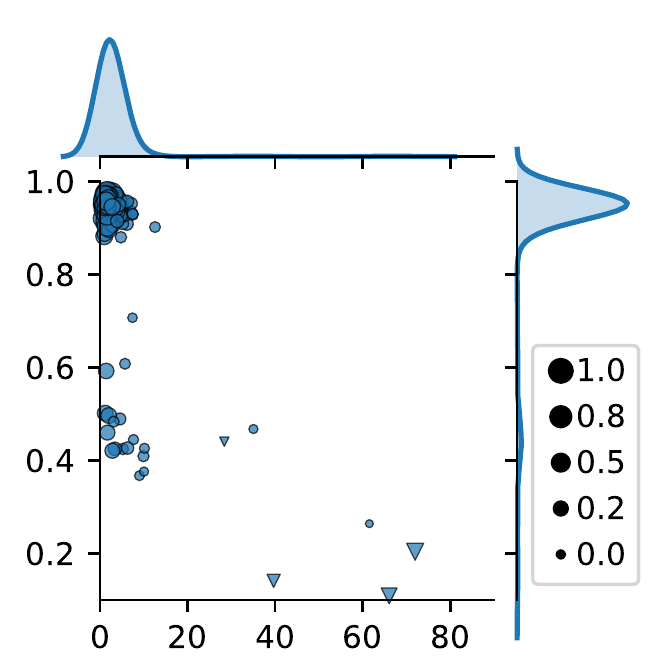}
					\put(0, 0){\textbf{\textcolor{black}{\large (2 - 9 dim. critical space)}}}
				\end{overpic}
  \end{minipage}}\end{minipage}

  \begin{minipage}{0.32\textwidth}\framebox{\begin{minipage}{\textwidth}
				\begin{overpic}[width=\textwidth, tics = 10, trim = 0 0 0 0 , clip]
					{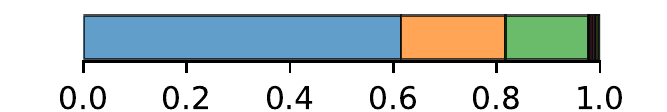}
				\end{overpic}
				\vspace*{2mm}
				\begin{overpic}[width=\textwidth, tics = 10, trim = 0 0 0 0 , clip]
					{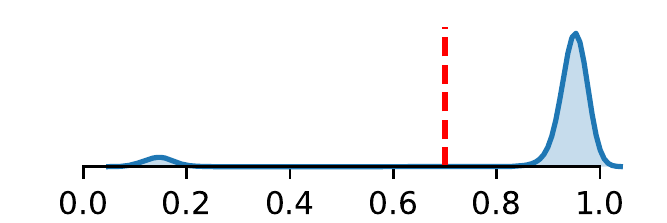}
				\end{overpic}
				\begin{overpic}[width=\textwidth, tics = 10, trim = 0 -10 0 0 , clip]
					{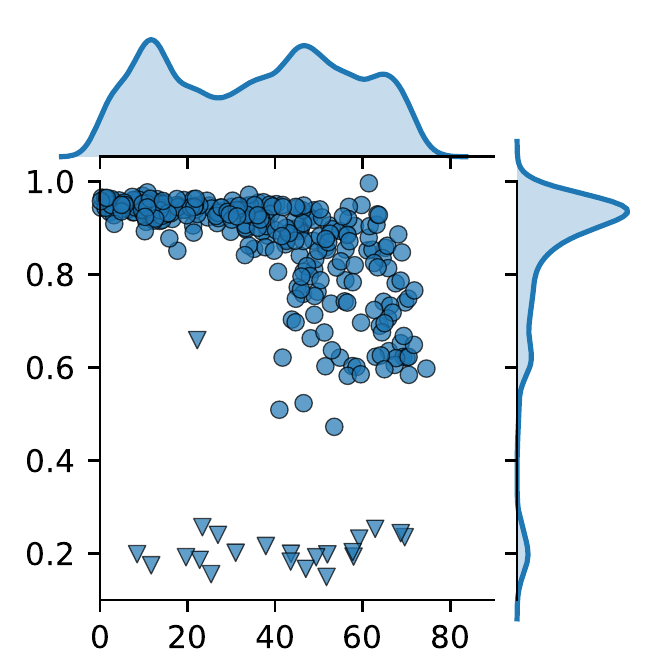}
					\put(0, 0){\textbf{\textcolor{black}{\large (3 - trivial critical space)}}}
				\end{overpic}
  \end{minipage}}\end{minipage}
  \hfill
  \begin{minipage}{0.32\textwidth}\framebox{\begin{minipage}{\textwidth}
				\begin{overpic}[width=\textwidth, tics = 10, trim = 0 0 0 0 , clip]
					{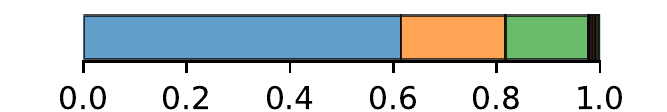}
				\end{overpic}
				\vspace*{2mm}
				\begin{overpic}[width=\textwidth, tics = 10, trim = 0 0 0 0 , clip]
					{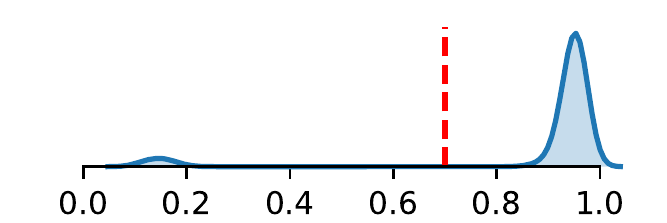}
				\end{overpic}
				\begin{overpic}[width=\textwidth, tics = 10, trim = 0 -10 0 0 , clip]
					{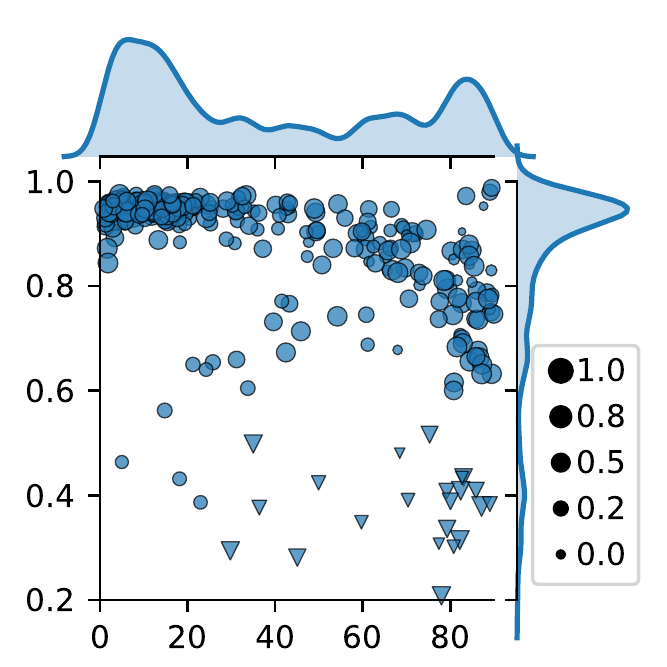}
					\put(0, 0){\textbf{\textcolor{black}{\large (3 - 9 dim. critical space)}}}
				\end{overpic}
  \end{minipage}}\end{minipage}
  \hfill
  \begin{minipage}{0.32\textwidth}\framebox{\begin{minipage}{\textwidth}
				\begin{overpic}[width=\textwidth, tics = 10, trim = 0 0 0 0 , clip]
					{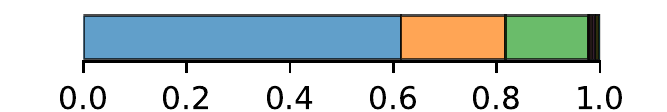}
				\end{overpic}
				\vspace*{2mm}
				\begin{overpic}[width=\textwidth, tics = 10, trim = 0 0 0 0 , clip]
					{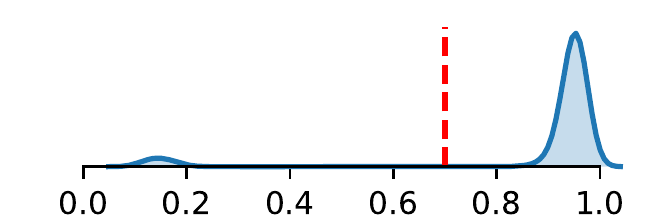}
				\end{overpic}
				\begin{overpic}[width=\textwidth, tics = 10, trim = 0 -10 0 0 , clip]
					{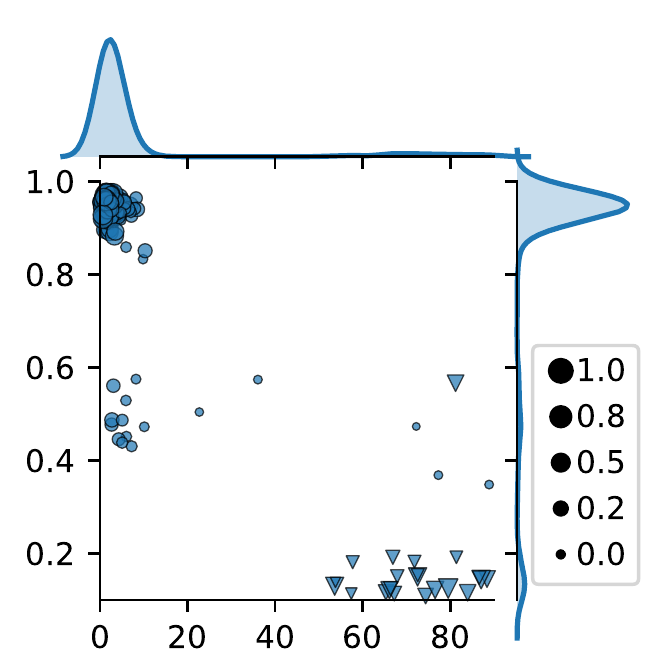}
					\put(0, 0){\textbf{\textcolor{black}{\large (3 - 8 dim. critical space)}}}
				\end{overpic}
  \end{minipage}}\end{minipage}

  \begin{minipage}{0.32\textwidth}\framebox{\begin{minipage}{\textwidth}
				\begin{overpic}[width=\textwidth, tics = 10, trim = 0 0 0 0 , clip]
					{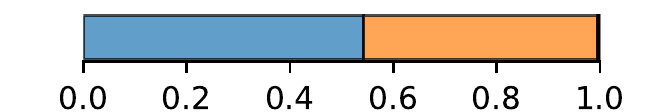}
				\end{overpic}
				\vspace*{2mm}
				\begin{overpic}[width=\textwidth, tics = 10, trim = 0 0 0 0 , clip]
					{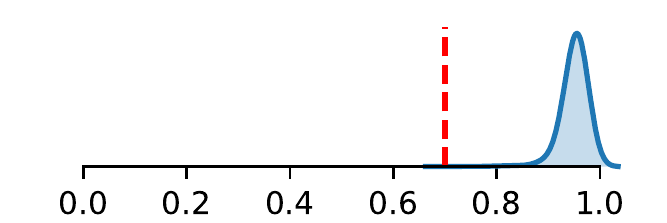}
				\end{overpic}
				\begin{overpic}[width=\textwidth, tics = 10, trim = 0 -10 0 0 , clip]
					{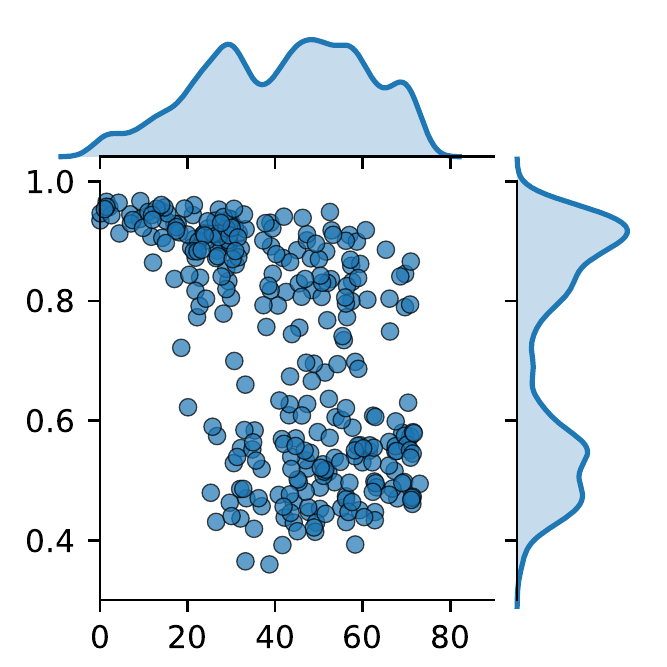}
					\put(0, 0){\textbf{\textcolor{black}{\large (4 - trivial critical space)}}}
				\end{overpic}
  \end{minipage}}\end{minipage}
  \hfill
  \begin{minipage}{0.32\textwidth}\framebox{\begin{minipage}{\textwidth}
				\begin{overpic}[width=\textwidth, tics = 10, trim = 0 0 0 0 , clip]
					{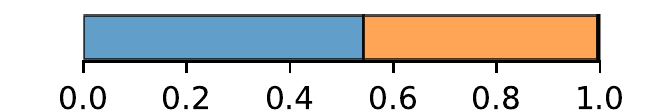}
				\end{overpic}
				\vspace*{2mm}
				\begin{overpic}[width=\textwidth, tics = 10, trim = 0 0 0 0 , clip]
					{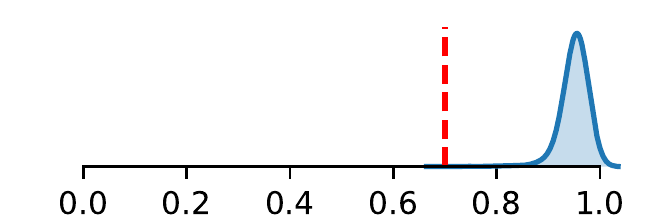}
				\end{overpic}
				\begin{overpic}[width=\textwidth, tics = 10, trim = 0 -10 0 0 , clip]
					{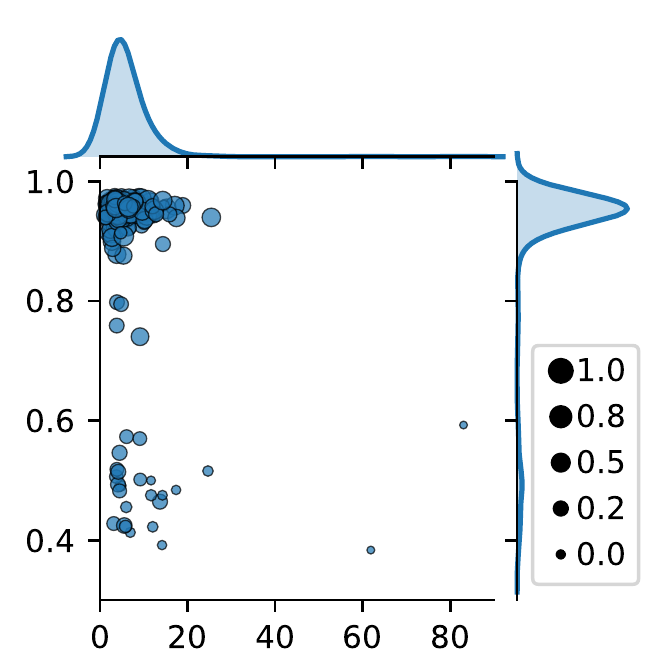}
					\put(0, 0){\textbf{\textcolor{black}{\large (4 - 9 dim. critical space)}}}
				\end{overpic}
  \end{minipage}}\end{minipage}
  \hfill
  \begin{minipage}{0.32\textwidth}\framebox{\begin{minipage}{\textwidth}
				\begin{overpic}[width=\textwidth, tics = 10, trim = 0 0 0 0 , clip]
					{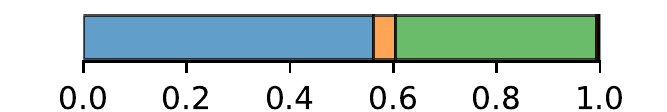}
				\end{overpic}
				\vspace*{2mm}
				\begin{overpic}[width=\textwidth, tics = 10, trim = 0 0 0 0 , clip]
					{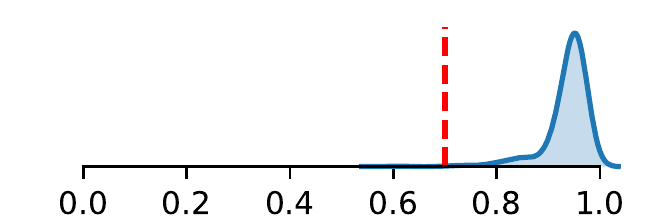}
				\end{overpic}
				\begin{overpic}[width=\textwidth, tics = 10, trim = 0 -10 0 0 , clip]
					{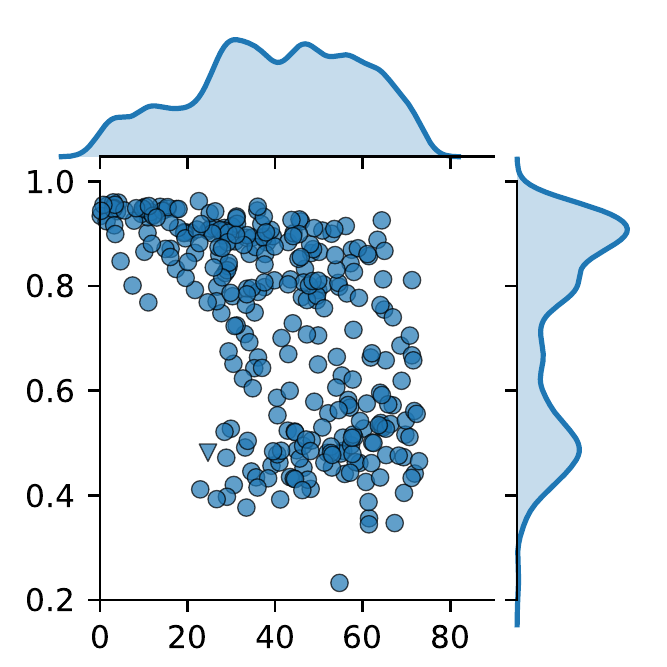}
					\put(0, 0){\textbf{\textcolor{black}{\large (5 - trivial critical space)}}}
				\end{overpic}
  \end{minipage}}\end{minipage}

  \begin{minipage}{0.32\textwidth}\framebox{\begin{minipage}{\textwidth}
				\begin{overpic}[width=\textwidth, tics = 10, trim = 0 0 0 0 , clip]
					{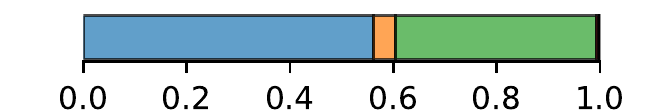}
				\end{overpic}
				\vspace*{2mm}
				\begin{overpic}[width=\textwidth, tics = 10, trim = 0 0 0 0 , clip]
					{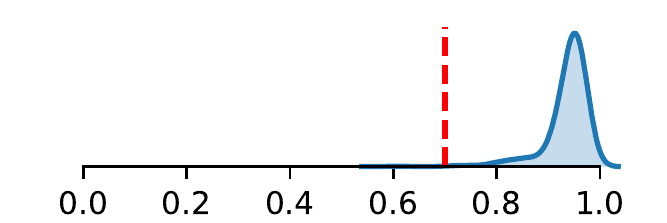}
				\end{overpic}
				\begin{overpic}[width=\textwidth, tics = 10, trim = 0 -10 0 0 , clip]
					{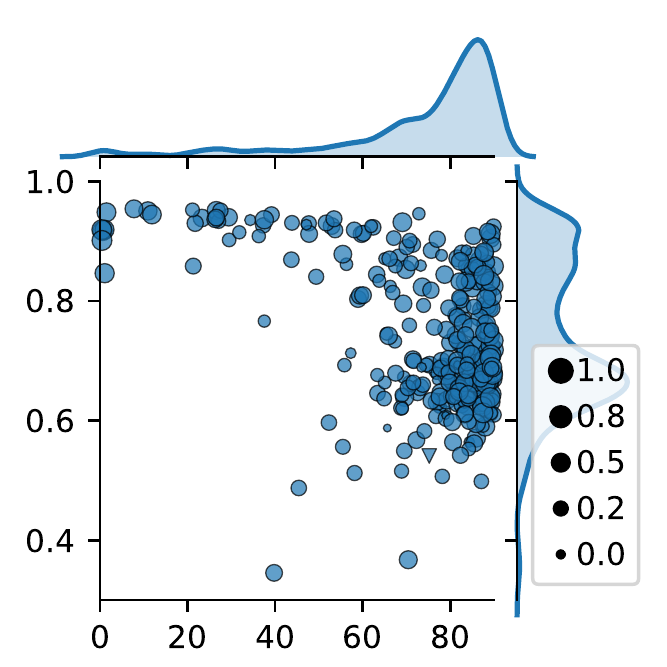}
					\put(0, 0){\textbf{\textcolor{black}{\large (5 - 9 dim. critical space)}}}
				\end{overpic}
  \end{minipage}}\end{minipage}
  \hfill
  \begin{minipage}{0.32\textwidth}\framebox{\begin{minipage}{\textwidth}
				\begin{overpic}[width=\textwidth, tics = 10, trim = 0 0 0 0 , clip]
					{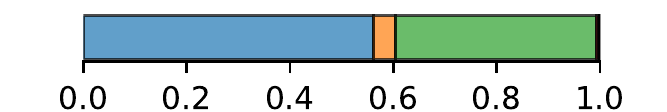}
				\end{overpic}
				\vspace*{2mm}
				\begin{overpic}[width=\textwidth, tics = 10, trim = 0 0 0 0 , clip]
					{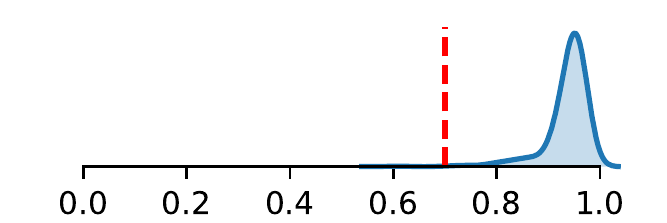}
				\end{overpic}
				\begin{overpic}[width=\textwidth, tics = 10, trim = 0 -10 0 0 , clip]
					{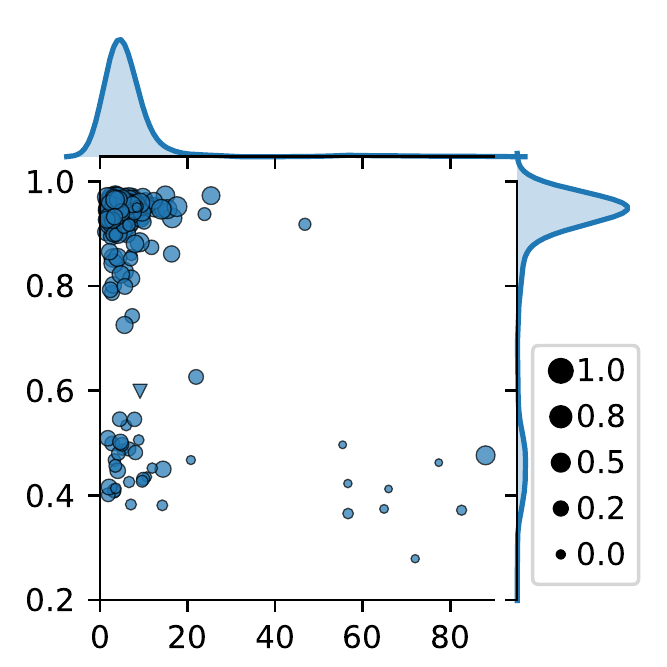}
					\put(0, 0){\textbf{\textcolor{black}{\large (5 - 8 dim. critical space)}}}
				\end{overpic}
  \end{minipage}}\end{minipage}
  \hfill
  \begin{minipage}{0.32\textwidth}\framebox{\begin{minipage}{\textwidth}
				\begin{overpic}[width=\textwidth, tics = 10, trim = 0 0 0 0 , clip]
					{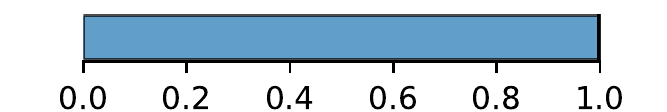}
				\end{overpic}
				\vspace*{2mm}
				\begin{overpic}[width=\textwidth, tics = 10, trim = 0 0 0 0 , clip]
					{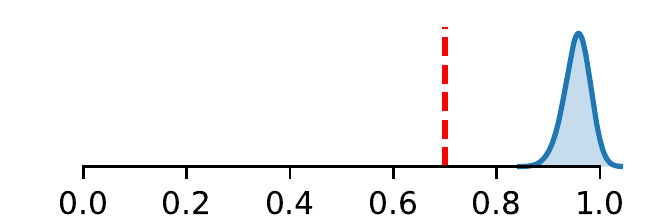}
				\end{overpic}
				\begin{overpic}[width=\textwidth, tics = 10, trim = 0 -10 0 0 , clip]
					{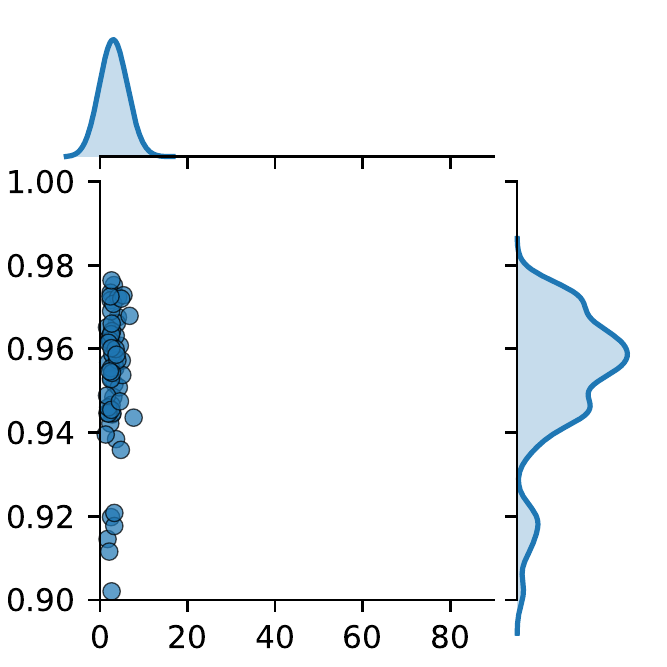}
					\put(0, 0){\textbf{\textcolor{black}{\large (6 - 32 filters)}}}
				\end{overpic}
  \end{minipage}}\end{minipage}

  \begin{minipage}{0.32\textwidth}\framebox{\begin{minipage}{\textwidth}
				\begin{overpic}[width=\textwidth, tics = 10, trim = 0 0 0 0 , clip]
					{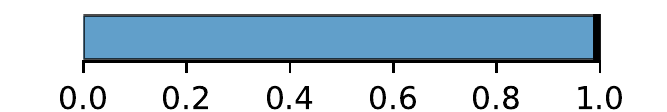}
				\end{overpic}
				\vspace*{2mm}
				\begin{overpic}[width=\textwidth, tics = 10, trim = 0 0 0 0 , clip]
					{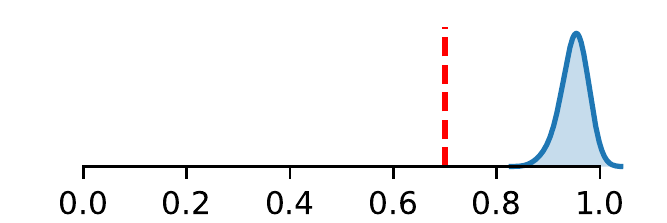}
				\end{overpic}
				\begin{overpic}[width=\textwidth, tics = 10, trim = 0 -10 0 0 , clip]
					{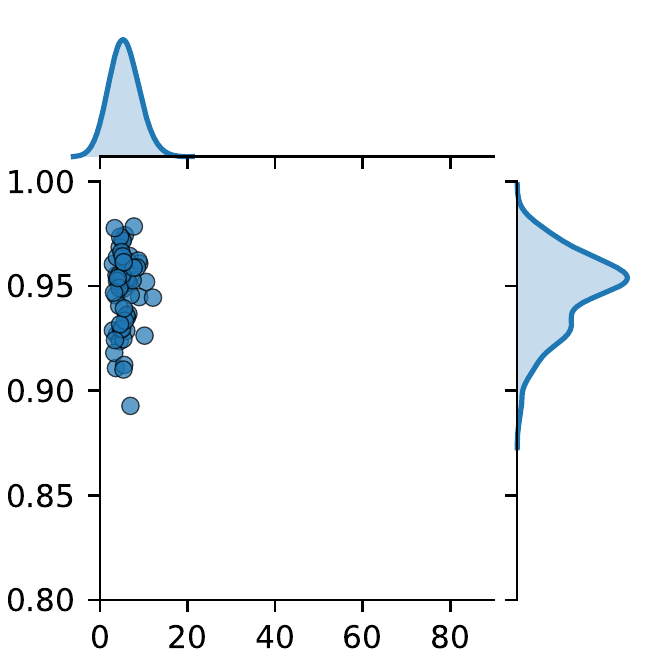}
					\put(0, 0){\textbf{\textcolor{black}{\large (6 - 64 filters)}}}
				\end{overpic}
  \end{minipage}}\end{minipage}
  \hfill
  \begin{minipage}{0.32\textwidth}\framebox{\begin{minipage}{\textwidth}
				\begin{overpic}[width=\textwidth, tics = 10, trim = 0 0 0 0 , clip]
					{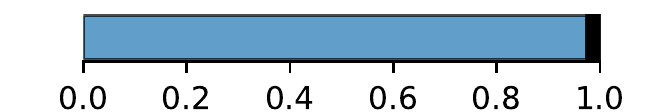}
				\end{overpic}
				\vspace*{2mm}
				\begin{overpic}[width=\textwidth, tics = 10, trim = 0 0 0 0 , clip]
					{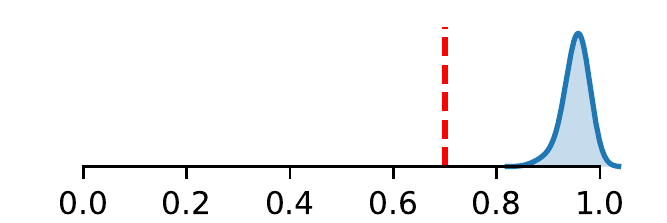}
				\end{overpic}
				\begin{overpic}[width=\textwidth, tics = 10, trim = 0 -10 0 0 , clip]
					{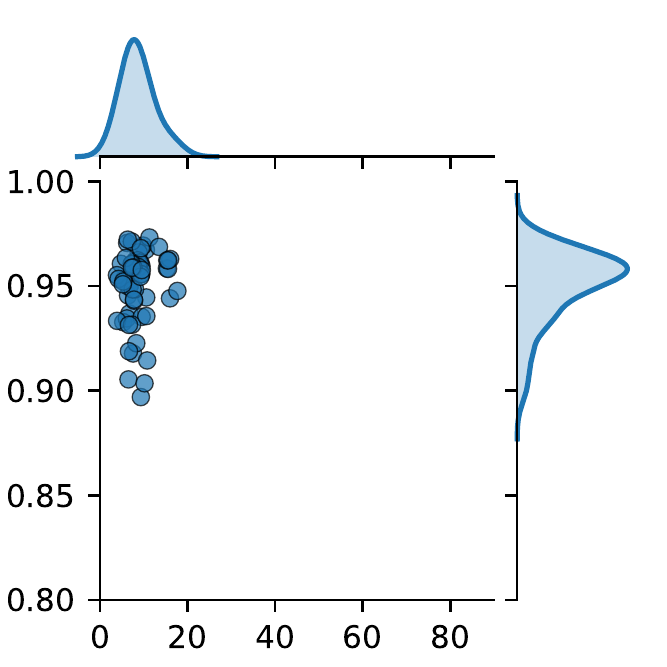}
					\put(0, 0){\textbf{\textcolor{black}{\large (6 - 96 filters)}}}
				\end{overpic}
  \end{minipage}}\end{minipage}
  \hfill
  \begin{minipage}{0.32\textwidth}\framebox{\begin{minipage}{\textwidth}
				\begin{overpic}[width=\textwidth, tics = 10, trim = 0 0 0 0 , clip]
					{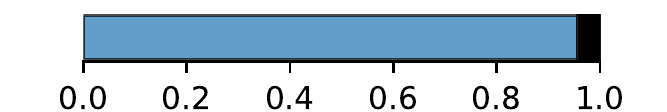}
				\end{overpic}
				\vspace*{2mm}
				\begin{overpic}[width=\textwidth, tics = 10, trim = 0 0 0 0 , clip]
					{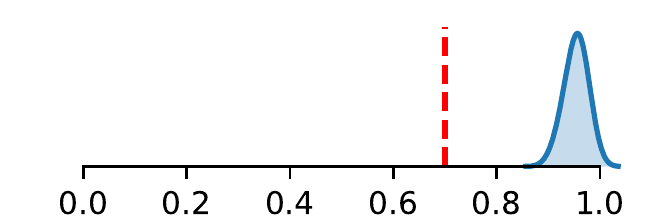}
				\end{overpic}
				\begin{overpic}[width=\textwidth, tics = 10, trim = 0 -10 0 0 , clip]
					{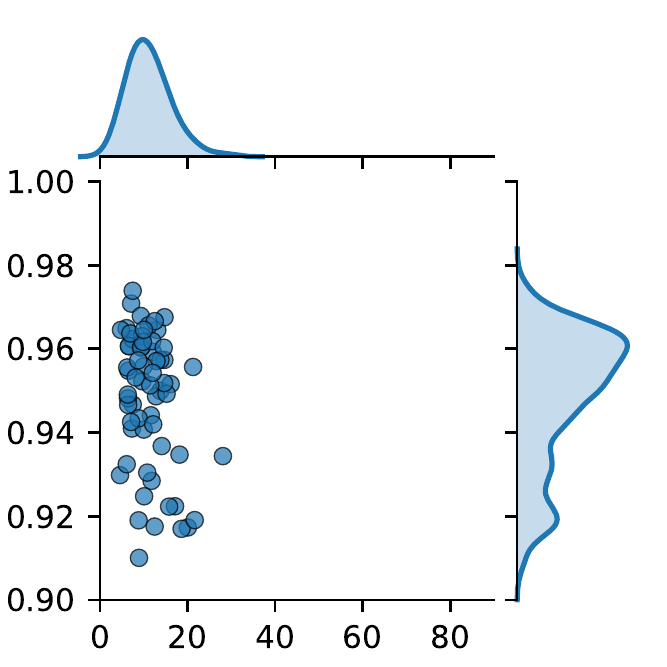}
					\put(0, 0){\textbf{\textcolor{black}{\large (6 - 128 filters)}}}
				\end{overpic}
  \end{minipage}}\end{minipage}

  \begin{minipage}{0.32\textwidth}\framebox{\begin{minipage}{\textwidth}
				\begin{overpic}[width=\textwidth, tics = 10, trim = 0 0 0 0 , clip]
					{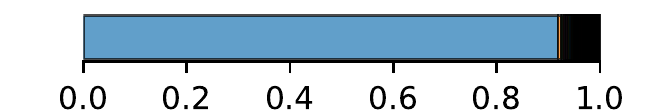}
				\end{overpic}
				\vspace*{2mm}
				\begin{overpic}[width=\textwidth, tics = 10, trim = 0 0 0 0 , clip]
					{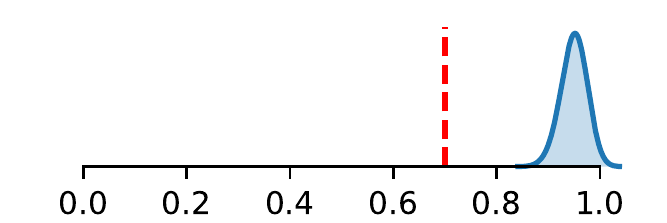}
				\end{overpic}
				\begin{overpic}[width=\textwidth, tics = 10, trim = 0 -10 0 0 , clip]
					{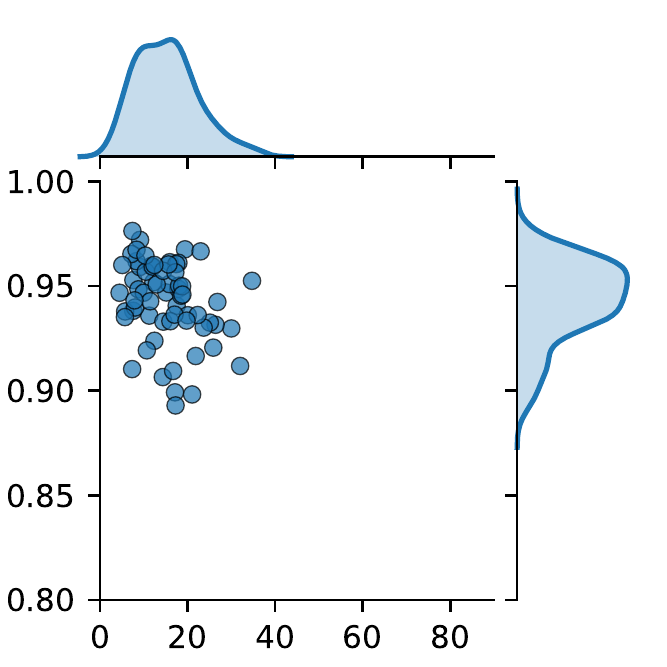}
					\put(0, 0){\textbf{\textcolor{black}{\large (6 - 160 filters)}}}
				\end{overpic}
  \end{minipage}}\end{minipage}
  \hfill
  \begin{minipage}{0.32\textwidth}\framebox{\begin{minipage}{\textwidth}
				\begin{overpic}[width=\textwidth, tics = 10, trim = 0 0 0 0 , clip]
					{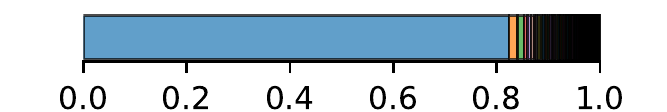}
				\end{overpic}
				\vspace*{2mm}
				\begin{overpic}[width=\textwidth, tics = 10, trim = 0 0 0 0 , clip]
					{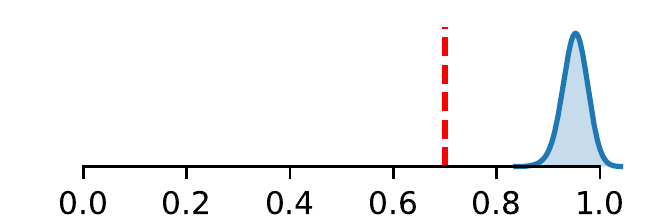}
				\end{overpic}
				\begin{overpic}[width=\textwidth, tics = 10, trim = 0 -10 0 0 , clip]
					{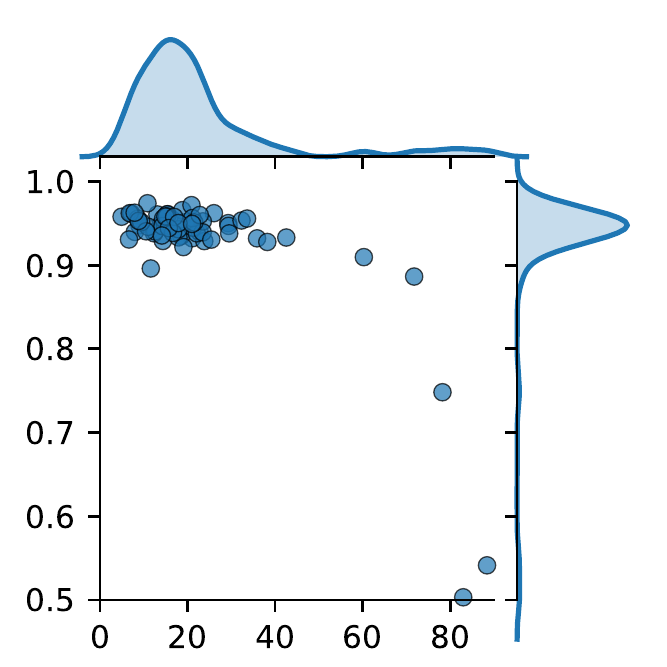}
					\put(0, 0){\textbf{\textcolor{black}{\large (6 - 192 filters)}}}
				\end{overpic}
  \end{minipage}}\end{minipage}
  \hfill
  \begin{minipage}{0.32\textwidth}\framebox{\begin{minipage}{\textwidth}
				\begin{overpic}[width=\textwidth, tics = 10, trim = 0 0 0 0 , clip]
					{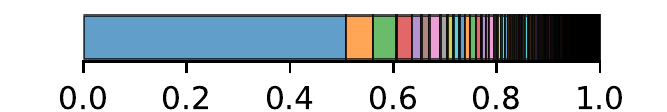}
				\end{overpic}
				\vspace*{2mm}
				\begin{overpic}[width=\textwidth, tics = 10, trim = 0 0 0 0 , clip]
					{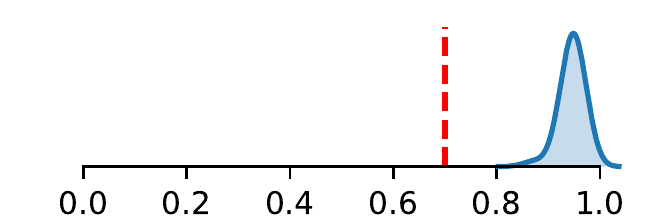}
				\end{overpic}
				\begin{overpic}[width=\textwidth, tics = 10, trim = 0 -10 0 0 , clip]
					{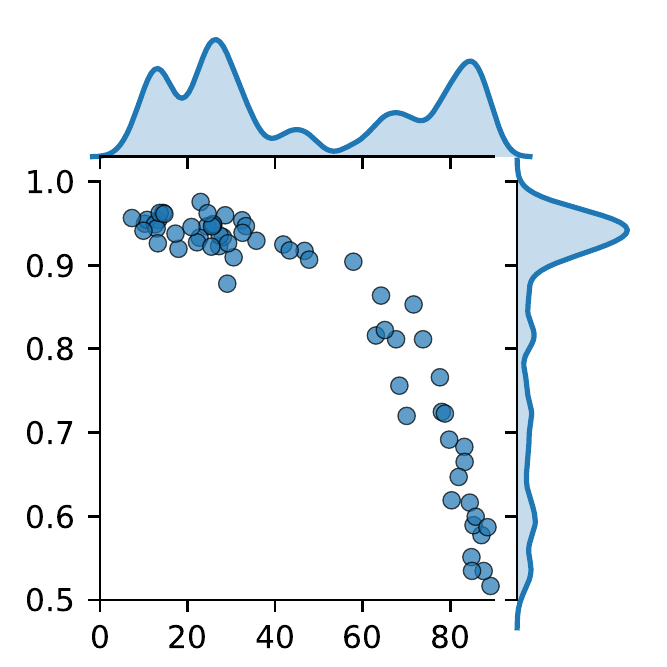}
					\put(0, 0){\textbf{\textcolor{black}{\large (6 - 224 filters)}}}
				\end{overpic}
  \end{minipage}}\end{minipage}

  \begin{minipage}{0.32\textwidth}\framebox{\begin{minipage}{\textwidth}
				\begin{overpic}[width=\textwidth, tics = 10, trim = 0 0 0 0 , clip]
					{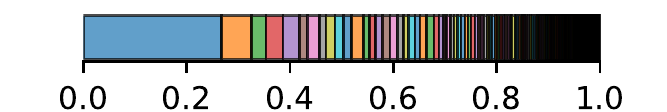}
				\end{overpic}
				\vspace*{2mm}
				\begin{overpic}[width=\textwidth, tics = 10, trim = 0 0 0 0 , clip]
					{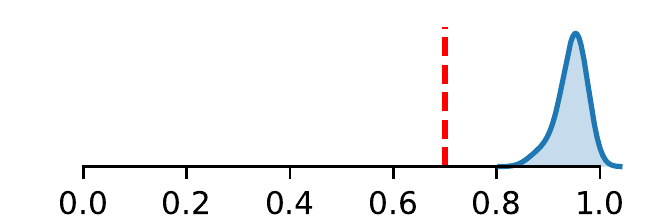}
				\end{overpic}
				\begin{overpic}[width=\textwidth, tics = 10, trim = 0 -10 0 0 , clip]
					{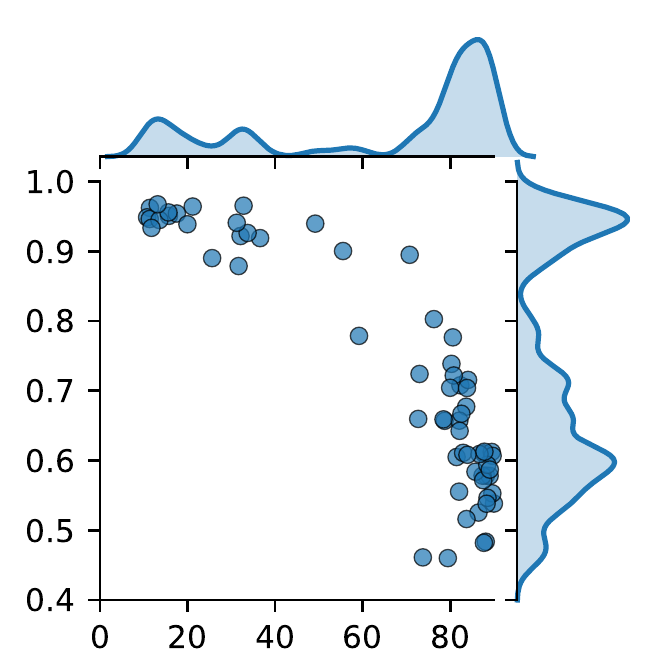}
					\put(0, 0){\textbf{\textcolor{black}{\large (6 - 256 filters)}}}
				\end{overpic}
  \end{minipage}}\end{minipage}
  \hfill
  \begin{minipage}{0.32\textwidth}\framebox{\begin{minipage}{\textwidth}
				\begin{overpic}[width=\textwidth, tics = 10, trim = 0 0 0 0 , clip]
					{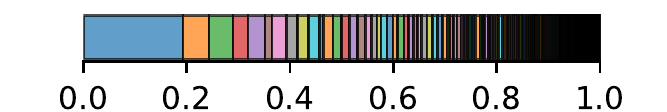}
				\end{overpic}
				\vspace*{2mm}
				\begin{overpic}[width=\textwidth, tics = 10, trim = 0 0 0 0 , clip]
					{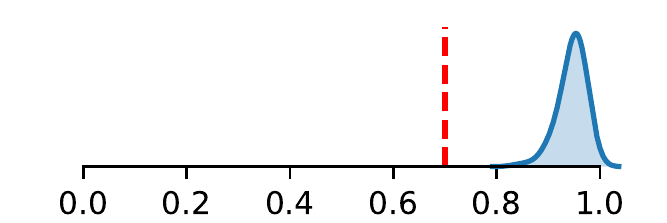}
				\end{overpic}
				\begin{overpic}[width=\textwidth, tics = 10, trim = 0 -10 0 0 , clip]
					{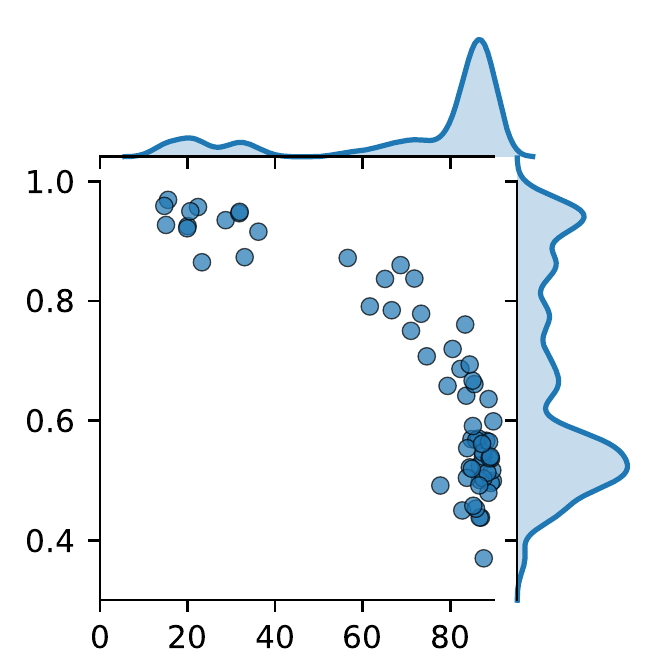}
					\put(0, 0){\textbf{\textcolor{black}{\large (6 - 288 filters)}}}
				\end{overpic}
  \end{minipage}}\end{minipage}
  \hfill
  \begin{minipage}{0.32\textwidth}\framebox{\begin{minipage}{\textwidth}
				\begin{overpic}[width=\textwidth, tics = 10, trim = 0 0 0 0 , clip]
					{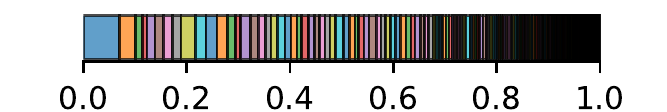}
				\end{overpic}
				\vspace*{2mm}
				\begin{overpic}[width=\textwidth, tics = 10, trim = 0 0 0 0 , clip]
					{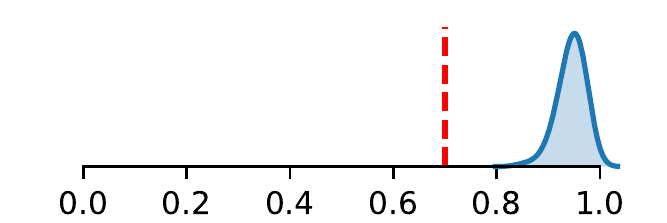}
				\end{overpic}
				\begin{overpic}[width=\textwidth, tics = 10, trim = 0 -10 0 0 , clip]
					{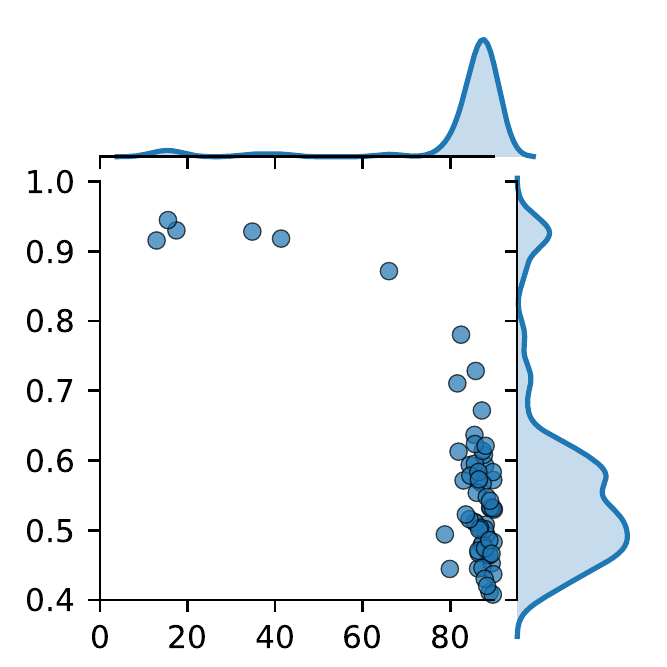}
					\put(0, 0){\textbf{\textcolor{black}{\large (6 - 320 filters)}}}
				\end{overpic}
  \end{minipage}}\end{minipage}

  \begin{minipage}{0.32\textwidth}\framebox{\begin{minipage}{\textwidth}
				\begin{overpic}[width=\textwidth, tics = 10, trim = 0 0 0 0 , clip]
					{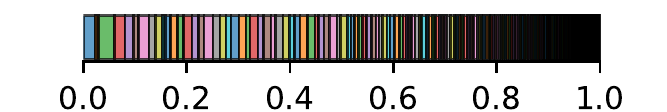}
				\end{overpic}
				\vspace*{2mm}
				\begin{overpic}[width=\textwidth, tics = 10, trim = 0 0 0 0 , clip]
					{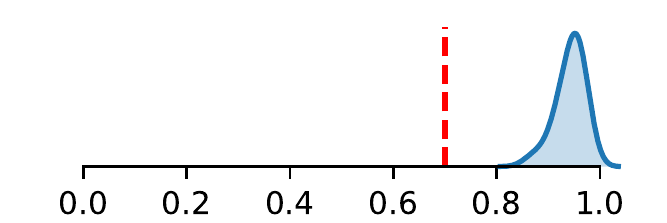}
				\end{overpic}
				\begin{overpic}[width=\textwidth, tics = 10, trim = 0 -10 0 0 , clip]
					{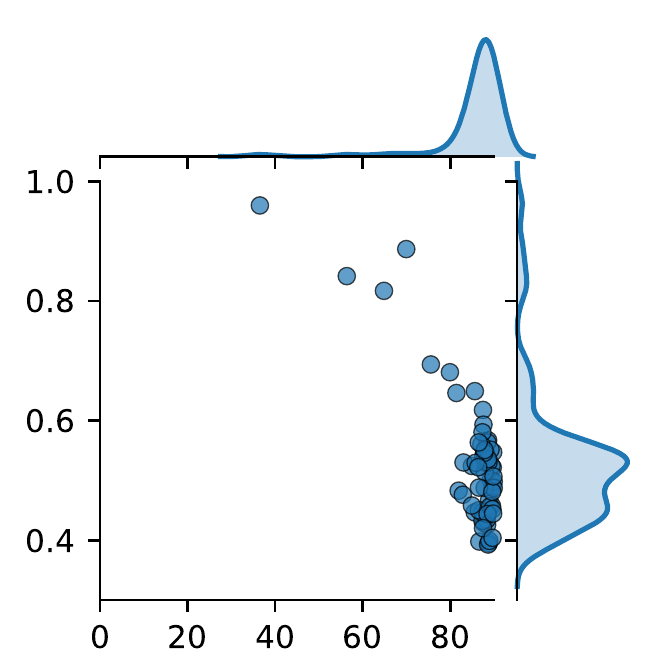}
					\put(0, 0){\textbf{\textcolor{black}{\large (6 - 342 filters)}}}
				\end{overpic}
  \end{minipage}}\end{minipage}
  \hfill
  \begin{minipage}{0.32\textwidth}\framebox{\begin{minipage}{\textwidth}
				\begin{overpic}[width=\textwidth, tics = 10, trim = 0 0 0 0 , clip]
					{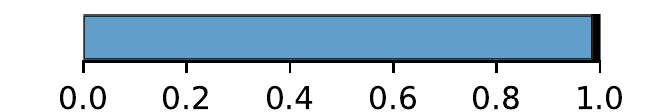}
				\end{overpic}
				\vspace*{2mm}
				\begin{overpic}[width=\textwidth, tics = 10, trim = 0 0 0 0 , clip]
					{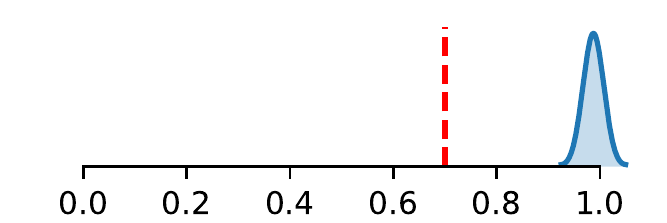}
				\end{overpic}
				\begin{overpic}[width=\textwidth, tics = 10, trim = 0 -10 0 0 , clip]
					{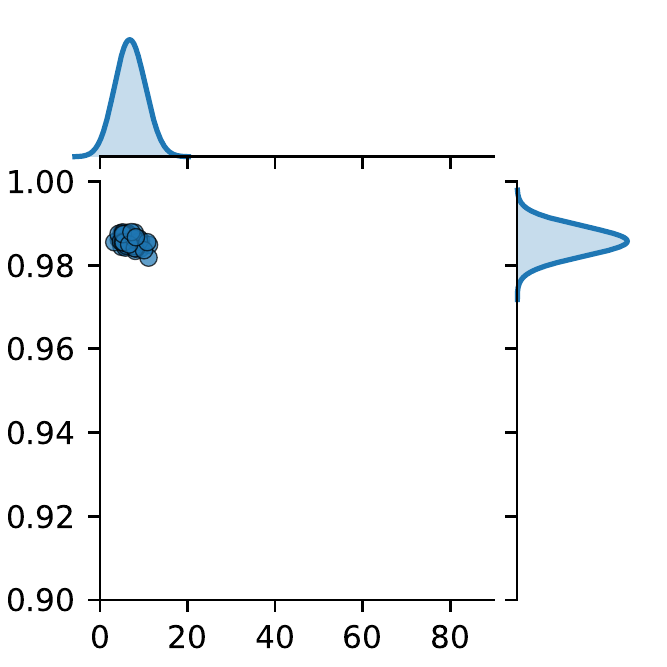}
					\put(0, 0){\textbf{\textcolor{black}{\large (7 - 32 filters)}}}
				\end{overpic}
  \end{minipage}}\end{minipage}
  \hfill
  \begin{minipage}{0.32\textwidth}\framebox{\begin{minipage}{\textwidth}
				\begin{overpic}[width=\textwidth, tics = 10, trim = 0 0 0 0 , clip]
					{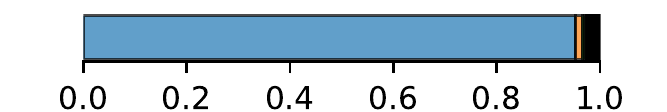}
				\end{overpic}
				\vspace*{2mm}
				\begin{overpic}[width=\textwidth, tics = 10, trim = 0 0 0 0 , clip]
					{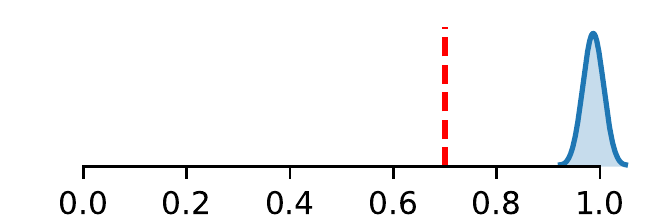}
				\end{overpic}
				\begin{overpic}[width=\textwidth, tics = 10, trim = 0 -10 0 0 , clip]
					{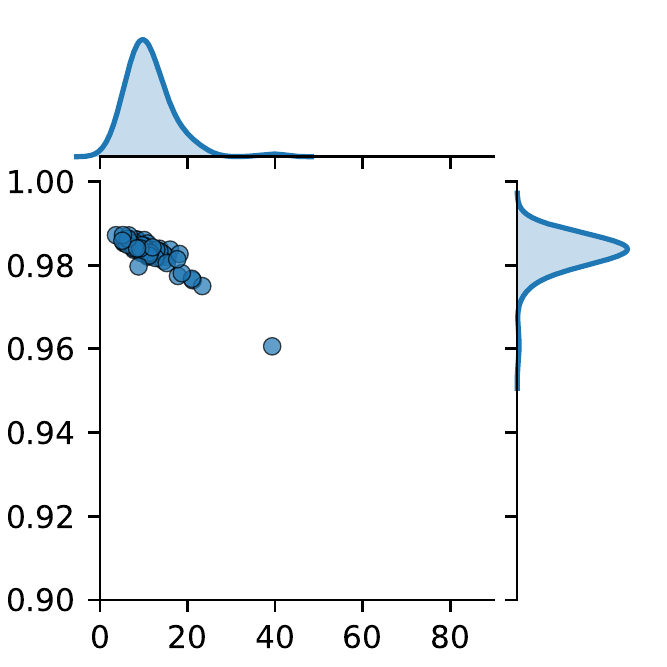}
					\put(0, 0){\textbf{\textcolor{black}{\large (7 - 64 filters)}}}
				\end{overpic}
  \end{minipage}}\end{minipage}

  \begin{minipage}{0.32\textwidth}\framebox{\begin{minipage}{\textwidth}
				\begin{overpic}[width=\textwidth, tics = 10, trim = 0 0 0 0 , clip]
					{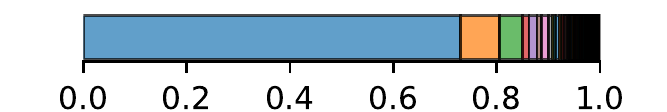}
				\end{overpic}
				\vspace*{2mm}
				\begin{overpic}[width=\textwidth, tics = 10, trim = 0 0 0 0 , clip]
					{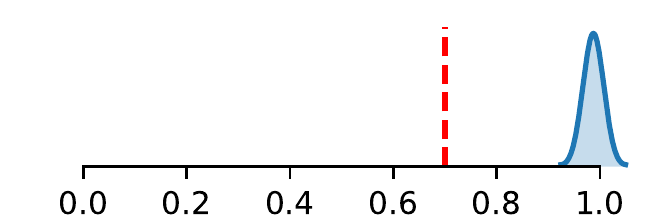}
				\end{overpic}
				\begin{overpic}[width=\textwidth, tics = 10, trim = 0 -10 0 0 , clip]
					{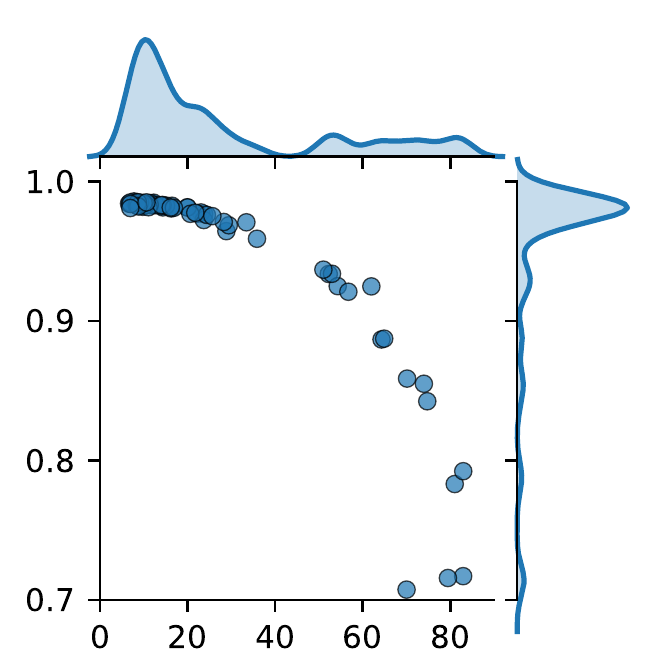}
					\put(0, 0){\textbf{\textcolor{black}{\large (7 - 96 filters)}}}
				\end{overpic}
  \end{minipage}}\end{minipage}
  \hfill
  \begin{minipage}{0.32\textwidth}\framebox{\begin{minipage}{\textwidth}
				\begin{overpic}[width=\textwidth, tics = 10, trim = 0 0 0 0 , clip]
					{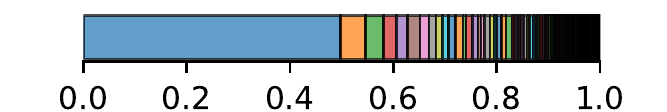}
				\end{overpic}
				\vspace*{2mm}
				\begin{overpic}[width=\textwidth, tics = 10, trim = 0 0 0 0 , clip]
					{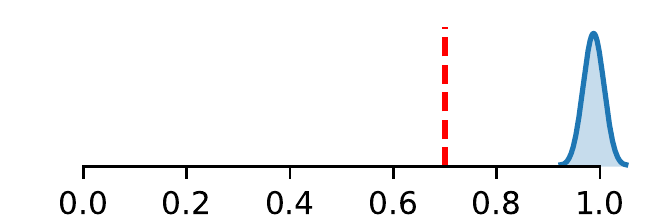}
				\end{overpic}
				\begin{overpic}[width=\textwidth, tics = 10, trim = 0 -10 0 0 , clip]
					{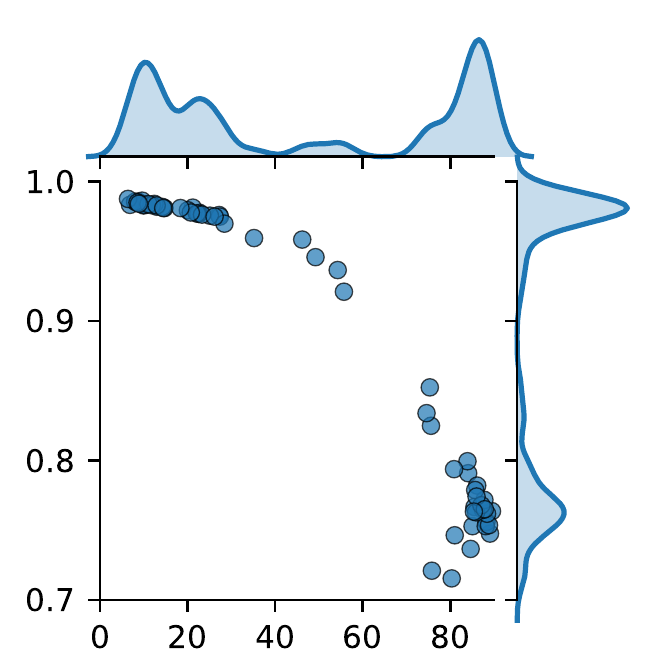}
					\put(0, 0){\textbf{\textcolor{black}{\large (7 - 128 filters)}}}
				\end{overpic}
  \end{minipage}}\end{minipage}
  \hfill
  \begin{minipage}{0.32\textwidth}\framebox{\begin{minipage}{\textwidth}
				\begin{overpic}[width=\textwidth, tics = 10, trim = 0 0 0 0 , clip]
					{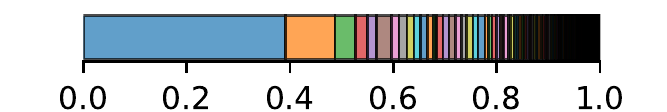}
				\end{overpic}
				\vspace*{2mm}
				\begin{overpic}[width=\textwidth, tics = 10, trim = 0 0 0 0 , clip]
					{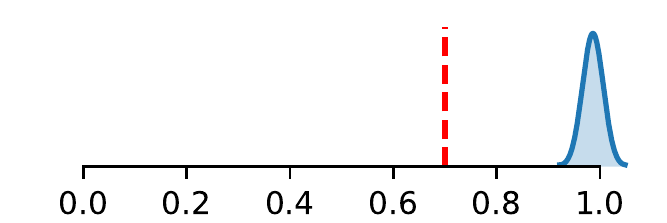}
				\end{overpic}
				\begin{overpic}[width=\textwidth, tics = 10, trim = 0 -10 0 0 , clip]
					{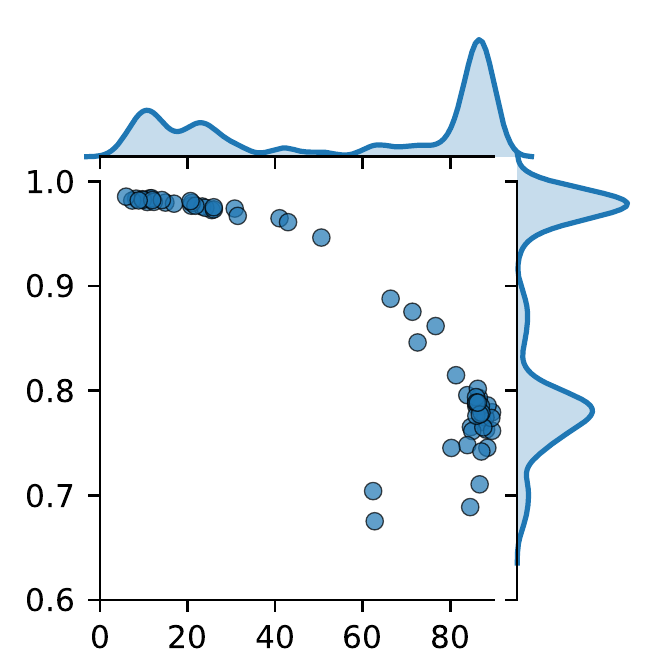}
					\put(0, 0){\textbf{\textcolor{black}{\large (7 - 160 filters)}}}
				\end{overpic}
  \end{minipage}}\end{minipage}

  \begin{minipage}{0.32\textwidth}\framebox{\begin{minipage}{\textwidth}
				\begin{overpic}[width=\textwidth, tics = 10, trim = 0 0 0 0 , clip]
					{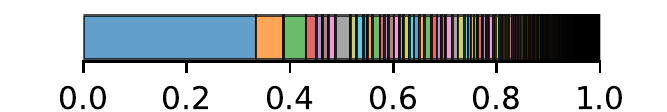}
				\end{overpic}
				\vspace*{2mm}
				\begin{overpic}[width=\textwidth, tics = 10, trim = 0 0 0 0 , clip]
					{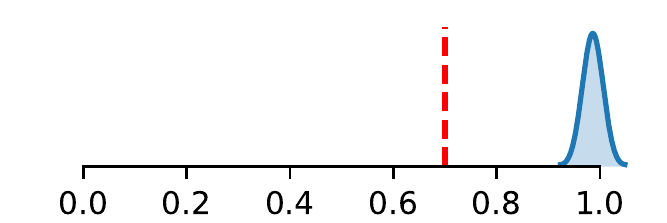}
				\end{overpic}
				\begin{overpic}[width=\textwidth, tics = 10, trim = 0 -10 0 0 , clip]
					{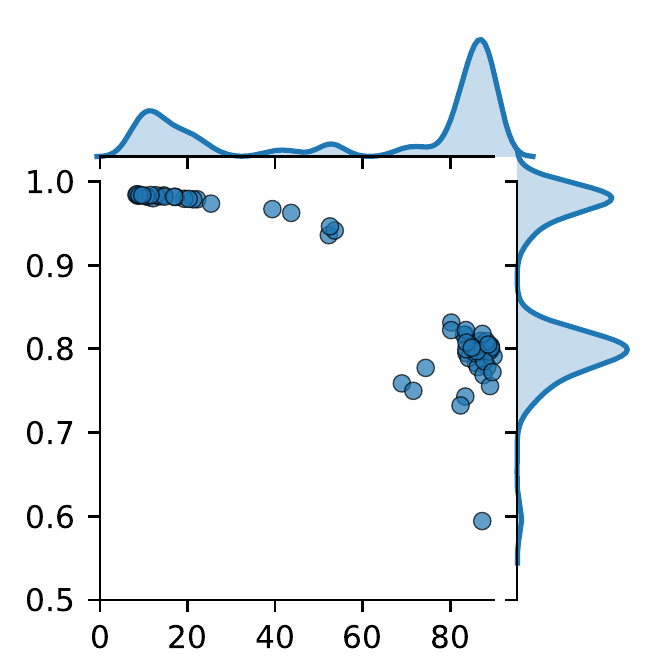}
					\put(0, 0){\textbf{\textcolor{black}{\large (7 - 192 filters)}}}
				\end{overpic}
  \end{minipage}}\end{minipage}
  \hfill
  \begin{minipage}{0.32\textwidth}\framebox{\begin{minipage}{\textwidth}
				\begin{overpic}[width=\textwidth, tics = 10, trim = 0 0 0 0 , clip]
					{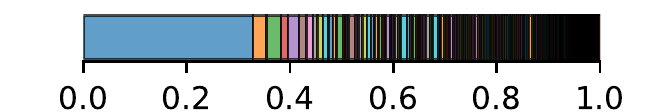}
				\end{overpic}
				\vspace*{2mm}
				\begin{overpic}[width=\textwidth, tics = 10, trim = 0 0 0 0 , clip]
					{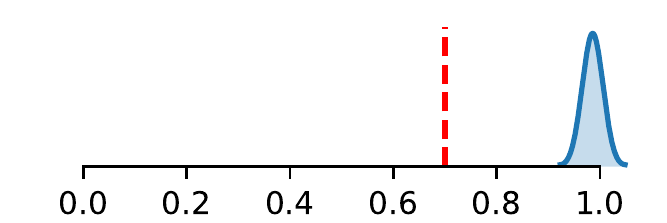}
				\end{overpic}
				\begin{overpic}[width=\textwidth, tics = 10, trim = 0 -10 0 0 , clip]
					{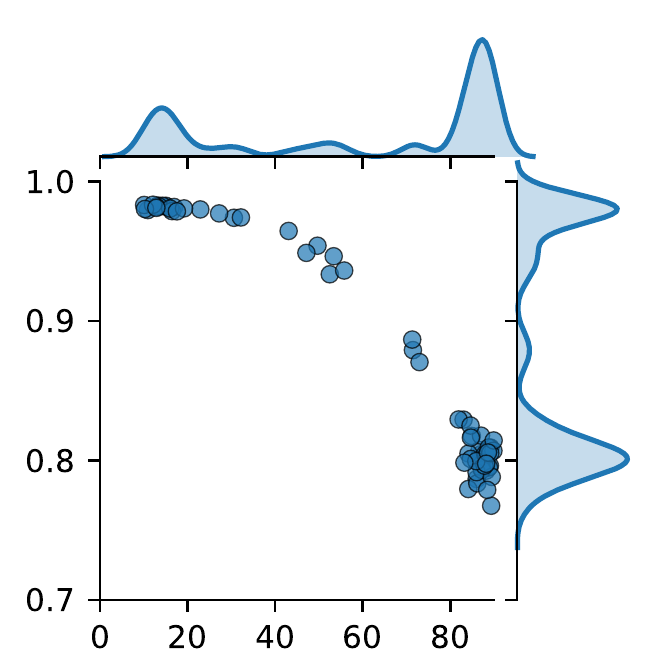}
					\put(0, 0){\textbf{\textcolor{black}{\large (7 - 224 filters)}}}
				\end{overpic}
  \end{minipage}}\end{minipage}
  \hfill
  \begin{minipage}{0.32\textwidth}\framebox{\begin{minipage}{\textwidth}
				\begin{overpic}[width=\textwidth, tics = 10, trim = 0 0 0 0 , clip]
					{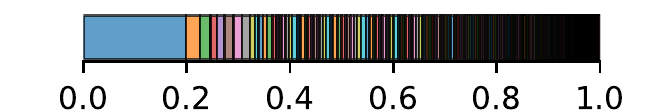}
				\end{overpic}
				\vspace*{2mm}
				\begin{overpic}[width=\textwidth, tics = 10, trim = 0 0 0 0 , clip]
					{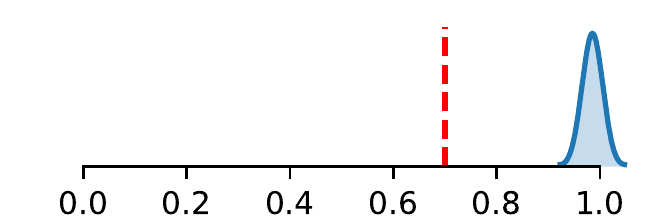}
				\end{overpic}
				\begin{overpic}[width=\textwidth, tics = 10, trim = 0 -10 0 0 , clip]
					{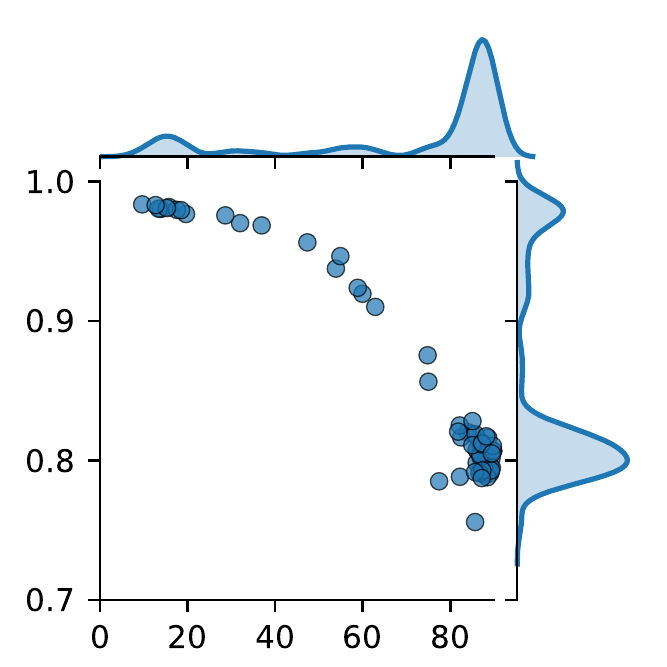}
					\put(0, 0){\textbf{\textcolor{black}{\large (7 - 256 filters)}}}
				\end{overpic}
  \end{minipage}}\end{minipage}

  \begin{minipage}{0.32\textwidth}\framebox{\begin{minipage}{\textwidth}
				\begin{overpic}[width=\textwidth, tics = 10, trim = 0 0 0 0 , clip]
					{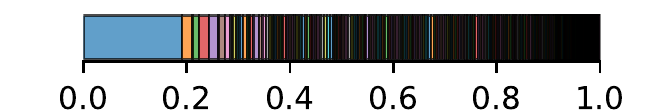}
				\end{overpic}
				\vspace*{2mm}
				\begin{overpic}[width=\textwidth, tics = 10, trim = 0 0 0 0 , clip]
					{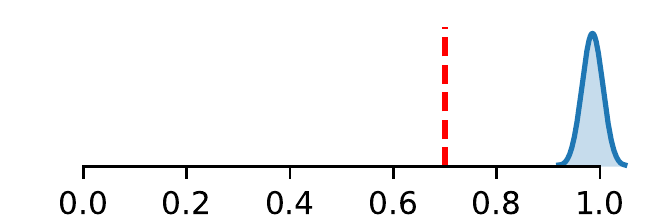}
				\end{overpic}
				\begin{overpic}[width=\textwidth, tics = 10, trim = 0 -10 0 0 , clip]
					{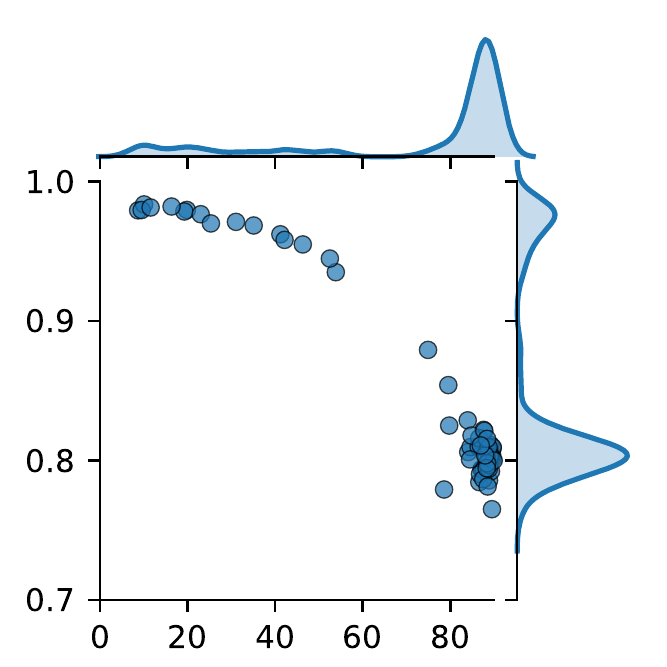}
					\put(0, 0){\textbf{\textcolor{black}{\large (7 - 288 filters)}}}
				\end{overpic}
  \end{minipage}}\end{minipage}
  \hfill
  \begin{minipage}{0.32\textwidth}\framebox{\begin{minipage}{\textwidth}
				\begin{overpic}[width=\textwidth, tics = 10, trim = 0 0 0 0 , clip]
					{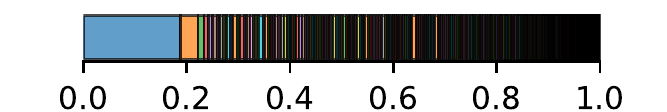}
				\end{overpic}
				\vspace*{2mm}
				\begin{overpic}[width=\textwidth, tics = 10, trim = 0 0 0 0 , clip]
					{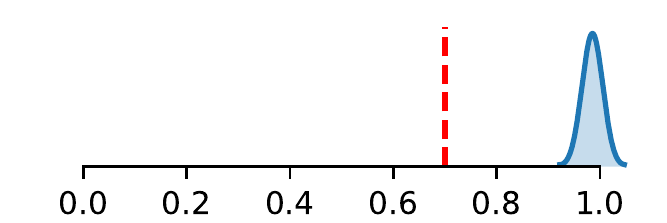}
				\end{overpic}
				\begin{overpic}[width=\textwidth, tics = 10, trim = 0 -10 0 0 , clip]
					{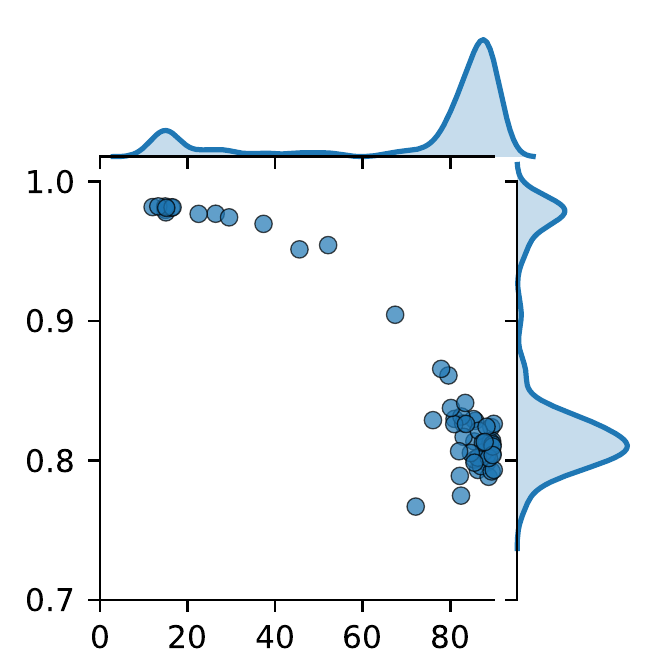}
					\put(0, 0){\textbf{\textcolor{black}{\large (7 - 320 filters)}}}
				\end{overpic}
  \end{minipage}}\end{minipage}
  \hfill
  \begin{minipage}{0.32\textwidth}\framebox{\begin{minipage}{\textwidth}
				\begin{overpic}[width=\textwidth, tics = 10, trim = 0 0 0 0 , clip]
					{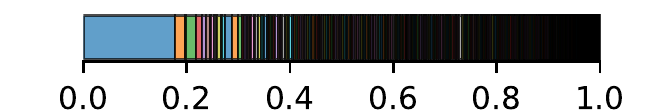}
				\end{overpic}
				\vspace*{2mm}
				\begin{overpic}[width=\textwidth, tics = 10, trim = 0 0 0 0 , clip]
					{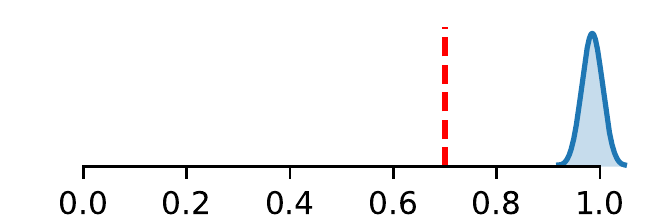}
				\end{overpic}
				\begin{overpic}[width=\textwidth, tics = 10, trim = 0 -10 0 0 , clip]
					{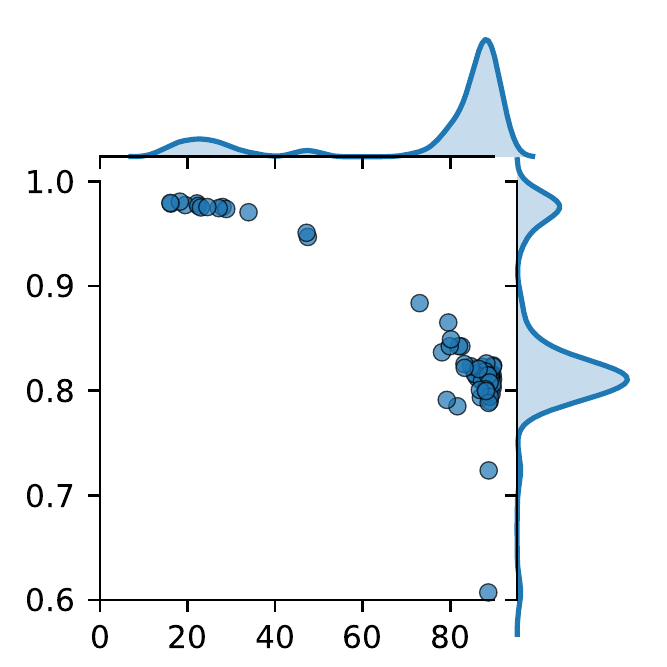}
					\put(0, 0){\textbf{\textcolor{black}{\large (7 - 352 filters)}}}
				\end{overpic}
  \end{minipage}}\end{minipage}

  \begin{minipage}{0.32\textwidth}\framebox{\begin{minipage}{\textwidth}
				\begin{overpic}[width=\textwidth, tics = 10, trim = 0 0 0 0 , clip]
					{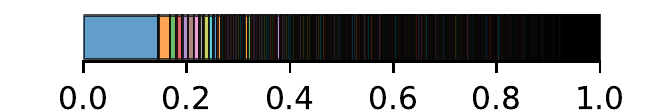}
				\end{overpic}
				\vspace*{2mm}
				\begin{overpic}[width=\textwidth, tics = 10, trim = 0 0 0 0 , clip]
					{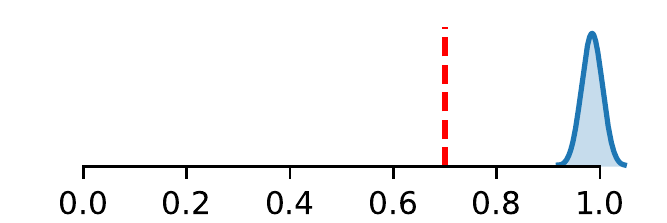}
				\end{overpic}
				\begin{overpic}[width=\textwidth, tics = 10, trim = 0 -10 0 0 , clip]
					{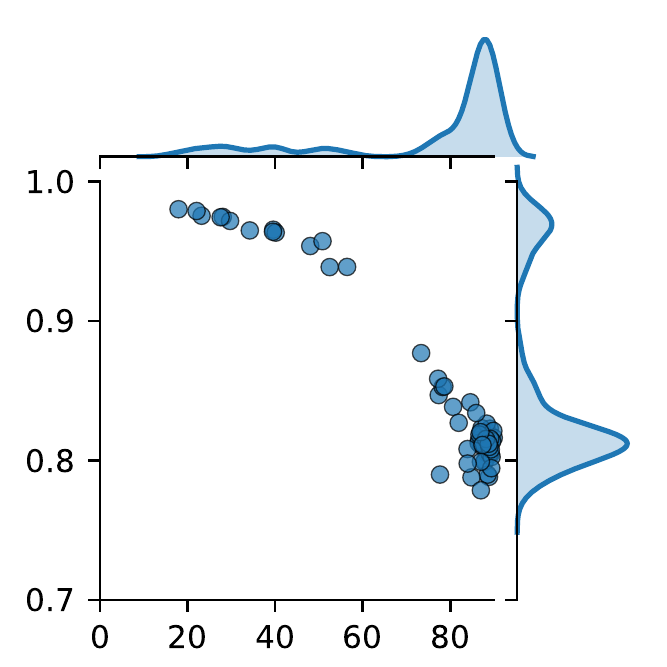}
					\put(0, 0){\textbf{\textcolor{black}{\large (7 - 384 filters)}}}
				\end{overpic}
  \end{minipage}}\end{minipage}
  \hfill
  \begin{minipage}{0.32\textwidth}\framebox{\begin{minipage}{\textwidth}
				\begin{overpic}[width=\textwidth, tics = 10, trim = 0 0 0 0 , clip]
					{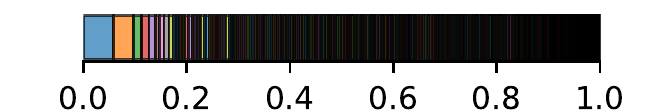}
				\end{overpic}
				\vspace*{2mm}
				\begin{overpic}[width=\textwidth, tics = 10, trim = 0 0 0 0 , clip]
					{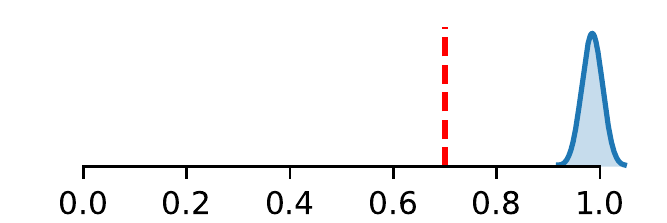}
				\end{overpic}
				\begin{overpic}[width=\textwidth, tics = 10, trim = 0 -10 0 0 , clip]
					{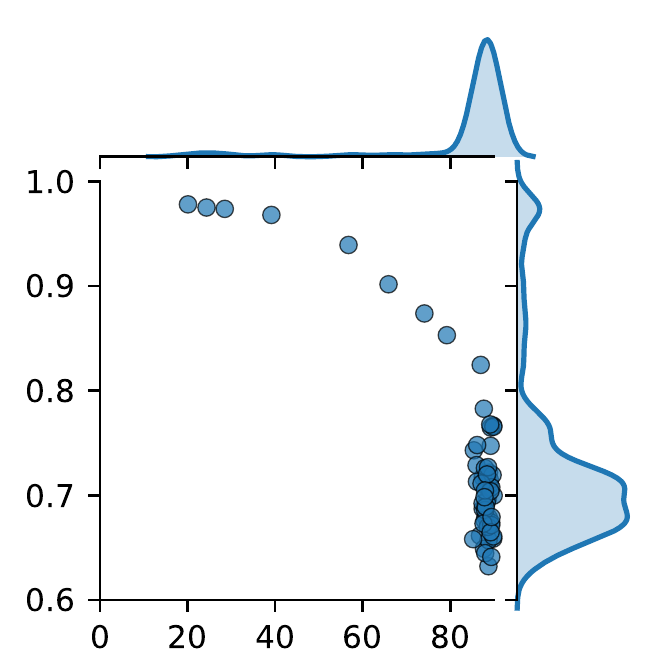}
					\put(0, 0){\textbf{\textcolor{black}{\large (7 - 416 filters)}}}
				\end{overpic}
  \end{minipage}}\end{minipage}
  \hfill
  \begin{minipage}{0.32\textwidth}\framebox{\begin{minipage}{\textwidth}
				\begin{overpic}[width=\textwidth, tics = 10, trim = 0 0 0 0 , clip]
					{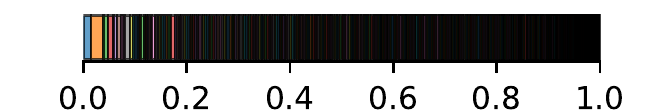}
				\end{overpic}
				\vspace*{2mm}
				\begin{overpic}[width=\textwidth, tics = 10, trim = 0 0 0 0 , clip]
					{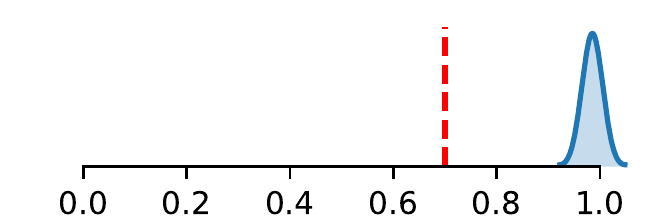}
				\end{overpic}
				\begin{overpic}[width=\textwidth, tics = 10, trim = 0 -10 0 0 , clip]
					{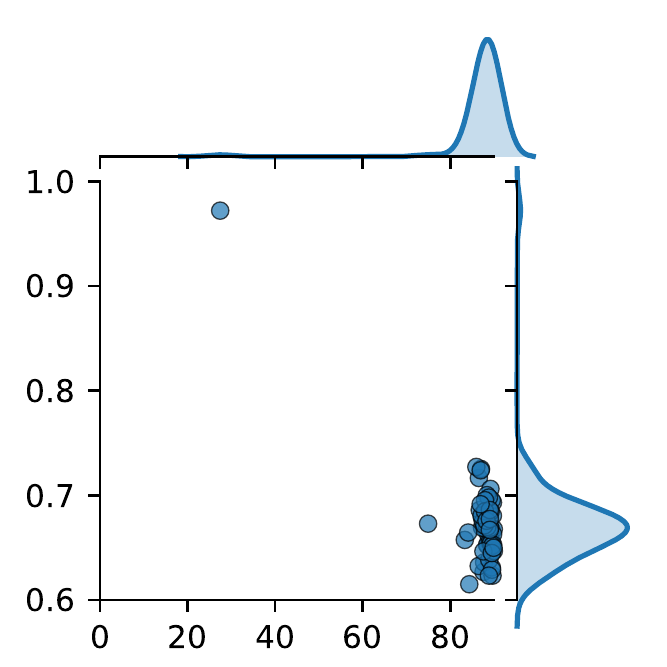}
					\put(0, 0){\textbf{\textcolor{black}{\large (7 - 448 filters)}}}
				\end{overpic}
  \end{minipage}}\end{minipage}

  \begin{minipage}{0.32\textwidth}\framebox{\begin{minipage}{\textwidth}
				\begin{overpic}[width=\textwidth, tics = 10, trim = 0 0 0 0 , clip]
					{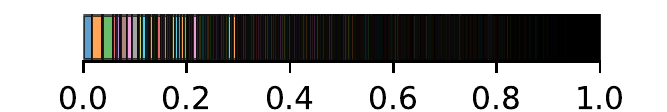}
				\end{overpic}
				\vspace*{2mm}
				\begin{overpic}[width=\textwidth, tics = 10, trim = 0 0 0 0 , clip]
					{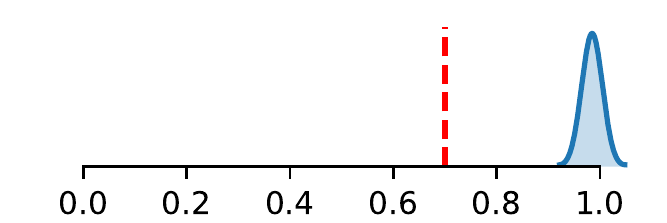}
				\end{overpic}
				\begin{overpic}[width=\textwidth, tics = 10, trim = 0 -10 0 0 , clip]
					{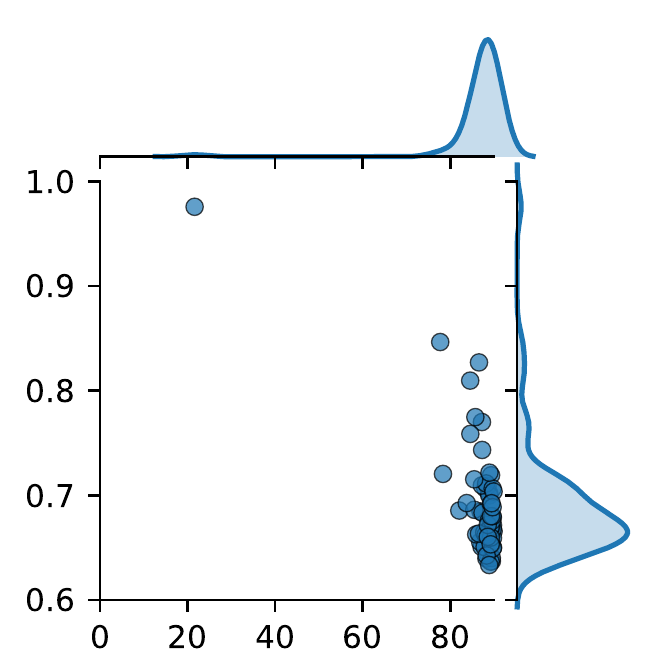}
					\put(0, 0){\textbf{\textcolor{black}{\large (7 - 480 filters)}}}
				\end{overpic}
  \end{minipage}}\end{minipage}
  \hfill
  \begin{minipage}{0.32\textwidth}\framebox{\begin{minipage}{\textwidth}
				\begin{overpic}[width=\textwidth, tics = 10, trim = 0 0 0 0 , clip]
					{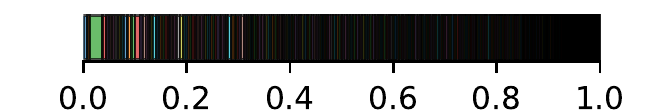}
				\end{overpic}
				\vspace*{2mm}
				\begin{overpic}[width=\textwidth, tics = 10, trim = 0 0 0 0 , clip]
					{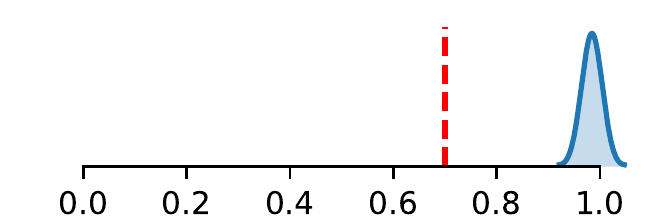}
				\end{overpic}
				\begin{overpic}[width=\textwidth, tics = 10, trim = 0 -10 0 0 , clip]
					{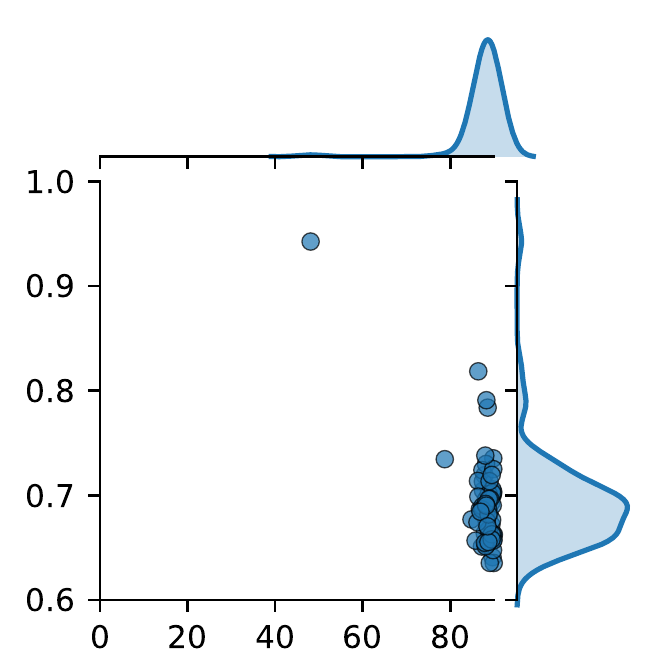}
					\put(0, 0){\textbf{\textcolor{black}{\large (7 - 512 filters)}}}
				\end{overpic}
  \end{minipage}}\end{minipage}
  \hfill
  \begin{minipage}{0.32\textwidth}\framebox{\begin{minipage}{\textwidth}
				\begin{overpic}[width=\textwidth, tics = 10, trim = 0 0 0 0 , clip]
					{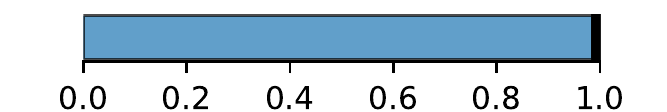}
				\end{overpic}
				\vspace*{2mm}
				\begin{overpic}[width=\textwidth, tics = 10, trim = 0 0 0 0 , clip]
					{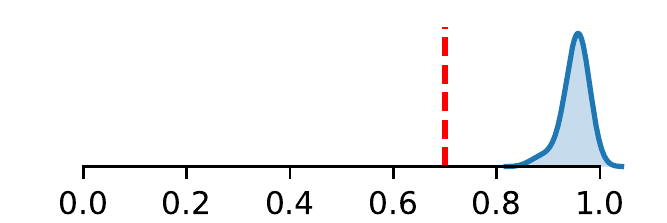}
				\end{overpic}
				\begin{overpic}[width=\textwidth, tics = 10, trim = 0 -10 0 0 , clip]
					{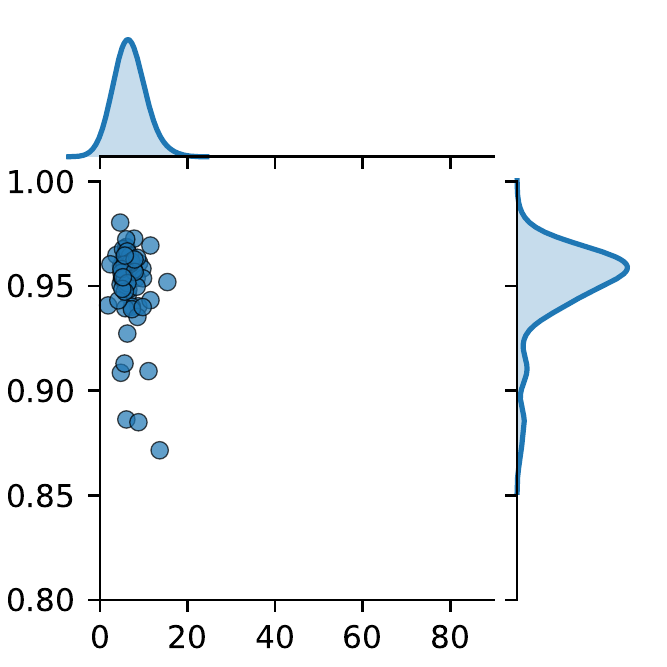}
					\put(0, 0){\textbf{\textcolor{black}{\large (8)}}}
				\end{overpic}
  \end{minipage}}\end{minipage}

  \begin{minipage}{0.32\textwidth}\framebox{\begin{minipage}{\textwidth}
				\begin{overpic}[width=\textwidth, tics = 10, trim = 0 0 0 0 , clip]
					{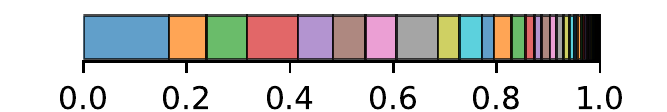}
				\end{overpic}
				\vspace*{2mm}
				\begin{overpic}[width=\textwidth, tics = 10, trim = 0 0 0 0 , clip]
					{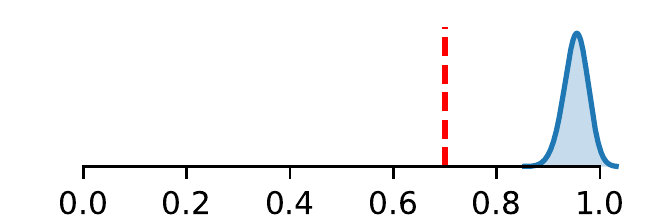}
				\end{overpic}
				\begin{overpic}[width=\textwidth, tics = 10, trim = 0 -10 0 0 , clip]
					{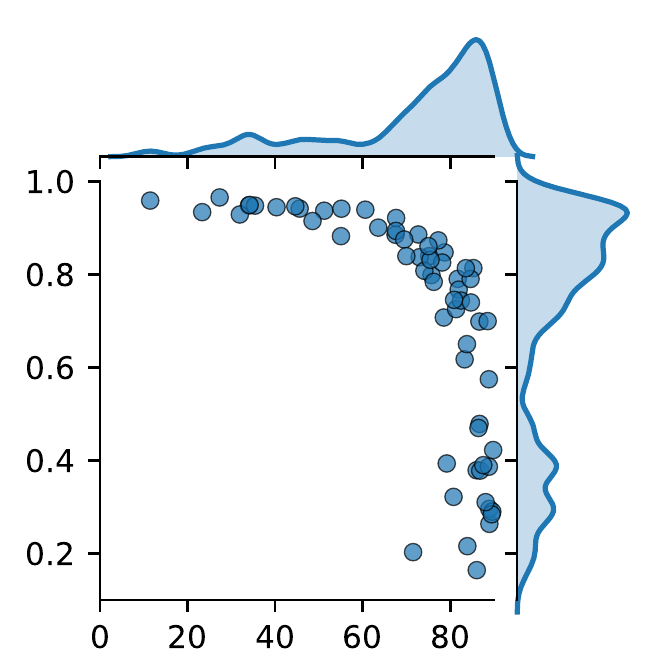}
					\put(0, 0){\textbf{\textcolor{black}{\large (9)}}}
				\end{overpic}
  \end{minipage}}\end{minipage}
  \hfill
  \begin{minipage}{0.32\textwidth}\framebox{\begin{minipage}{\textwidth}
				\begin{overpic}[width=\textwidth, tics = 10, trim = 0 0 0 0 , clip]
					{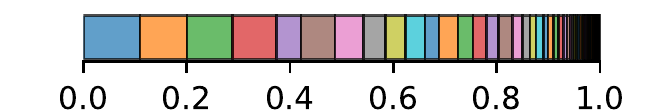}
				\end{overpic}
				\vspace*{2mm}
				\begin{overpic}[width=\textwidth, tics = 10, trim = 0 0 0 0 , clip]
					{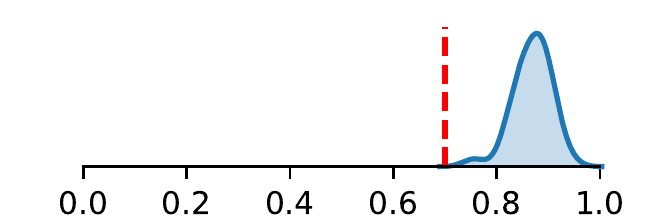}
				\end{overpic}
				\begin{overpic}[width=\textwidth, tics = 10, trim = 0 -10 0 0 , clip]
					{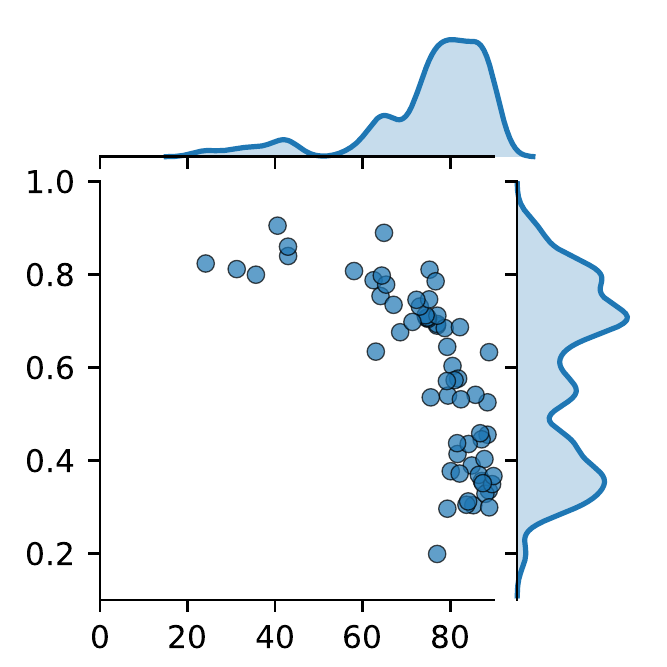}
					\put(0, 0){\textbf{\textcolor{black}{\large (10)}}}
				\end{overpic}
  \end{minipage}}\end{minipage}
  \hfill
  \begin{minipage}{0.32\textwidth}\framebox{\begin{minipage}{\textwidth}
				\begin{overpic}[width=\textwidth, tics = 10, trim = 0 0 0 0 , clip]
					{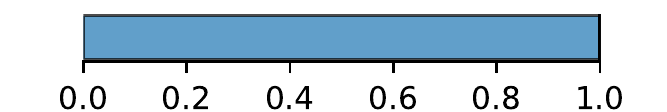}
				\end{overpic}
				\vspace*{2mm}
				\begin{overpic}[width=\textwidth, tics = 10, trim = 0 0 0 0 , clip]
					{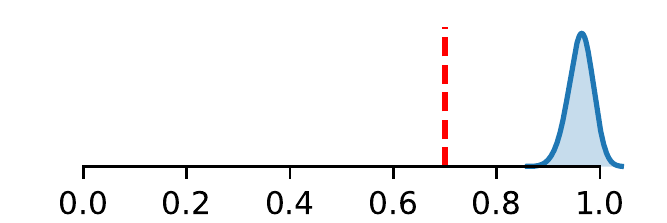}
				\end{overpic}
				\begin{overpic}[width=\textwidth, tics = 10, trim = 0 -10 0 0 , clip]
					{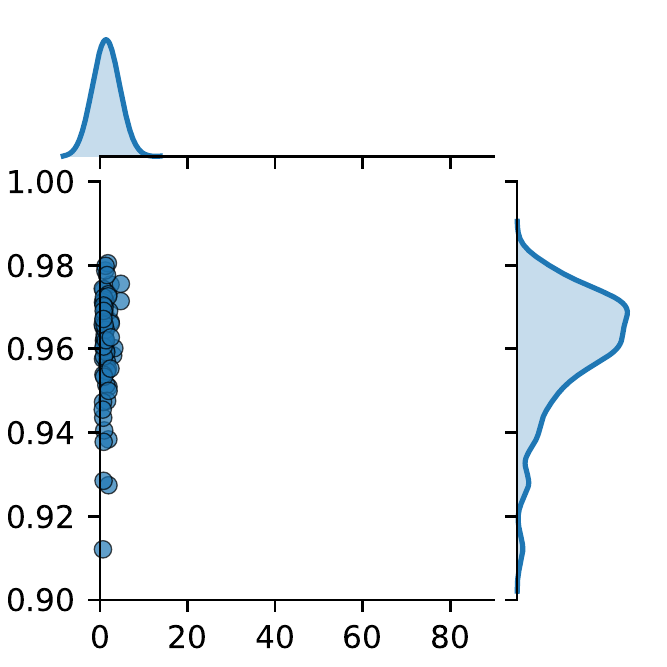}
					\put(0, 0){\textbf{\textcolor{black}{\large (11 - 32 filters)}}}
				\end{overpic}
  \end{minipage}}\end{minipage}

  \begin{minipage}{0.32\textwidth}\framebox{\begin{minipage}{\textwidth}
				\begin{overpic}[width=\textwidth, tics = 10, trim = 0 0 0 0 , clip]
					{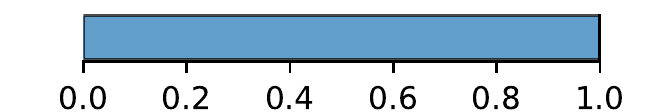}
				\end{overpic}
				\vspace*{2mm}
				\begin{overpic}[width=\textwidth, tics = 10, trim = 0 0 0 0 , clip]
					{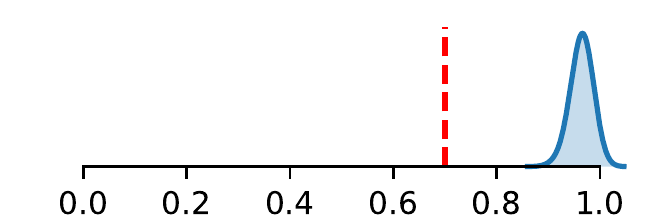}
				\end{overpic}
				\begin{overpic}[width=\textwidth, tics = 10, trim = 0 -10 0 0 , clip]
					{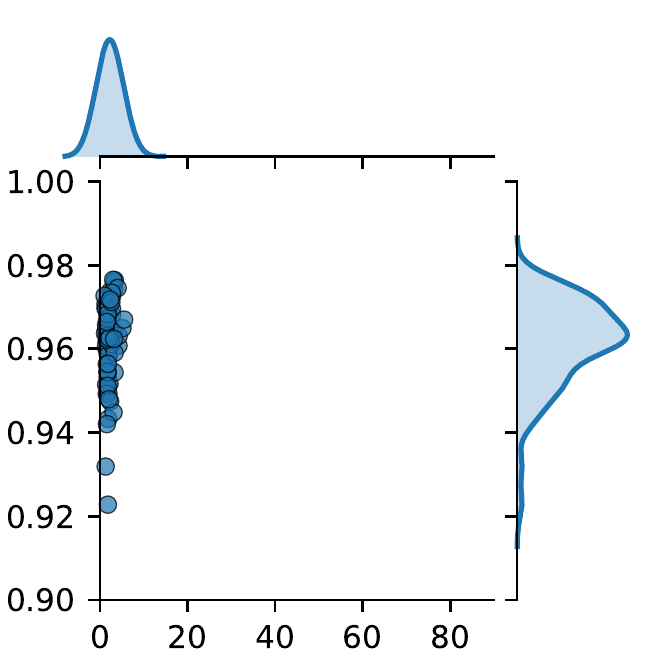}
					\put(0, 0){\textbf{\textcolor{black}{\large (11 - 64 filters)}}}
				\end{overpic}
  \end{minipage}}\end{minipage}
  \hfill
  \begin{minipage}{0.32\textwidth}\framebox{\begin{minipage}{\textwidth}
				\begin{overpic}[width=\textwidth, tics = 10, trim = 0 0 0 0 , clip]
					{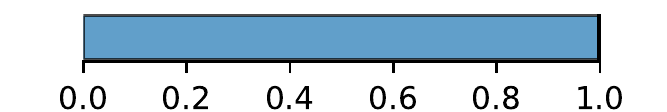}
				\end{overpic}
				\vspace*{2mm}
				\begin{overpic}[width=\textwidth, tics = 10, trim = 0 0 0 0 , clip]
					{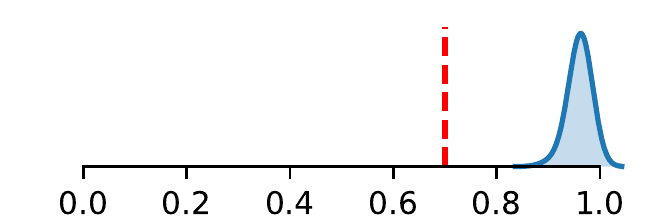}
				\end{overpic}
				\begin{overpic}[width=\textwidth, tics = 10, trim = 0 -10 0 0 , clip]
					{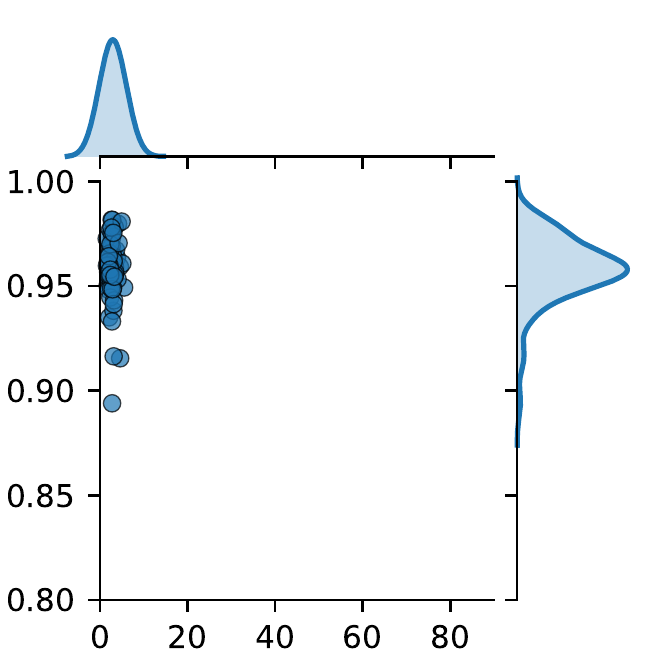}
					\put(0, 0){\textbf{\textcolor{black}{\large (11 - 96 filters)}}}
				\end{overpic}
  \end{minipage}}\end{minipage}
  \hfill
  \begin{minipage}{0.32\textwidth}\framebox{\begin{minipage}{\textwidth}
				\begin{overpic}[width=\textwidth, tics = 10, trim = 0 0 0 0 , clip]
					{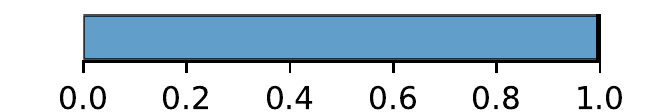}
				\end{overpic}
				\vspace*{2mm}
				\begin{overpic}[width=\textwidth, tics = 10, trim = 0 0 0 0 , clip]
					{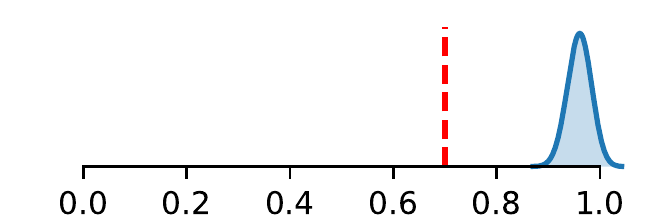}
				\end{overpic}
				\begin{overpic}[width=\textwidth, tics = 10, trim = 0 -10 0 0 , clip]
					{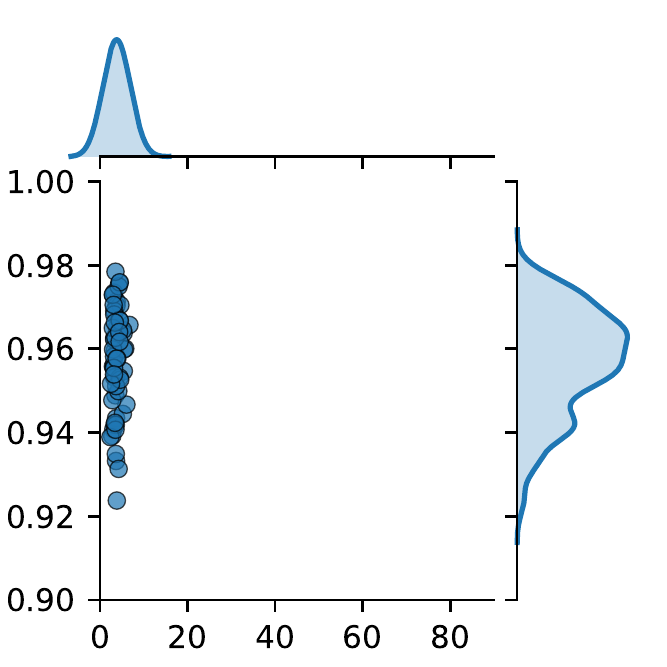}
					\put(0, 0){\textbf{\textcolor{black}{\large (11 - 128 filters)}}}
				\end{overpic}
  \end{minipage}}\end{minipage}

  \begin{minipage}{0.32\textwidth}\framebox{\begin{minipage}{\textwidth}
				\begin{overpic}[width=\textwidth, tics = 10, trim = 0 0 0 0 , clip]
					{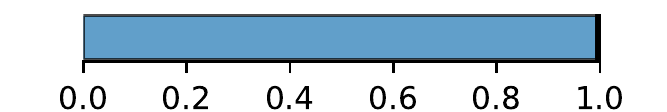}
				\end{overpic}
				\vspace*{2mm}
				\begin{overpic}[width=\textwidth, tics = 10, trim = 0 0 0 0 , clip]
					{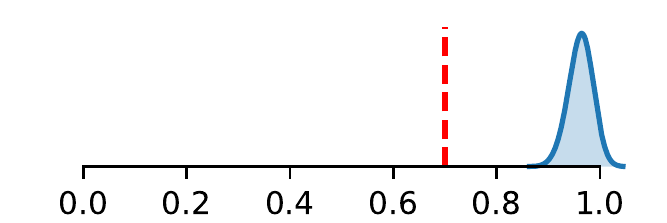}
				\end{overpic}
				\begin{overpic}[width=\textwidth, tics = 10, trim = 0 -10 0 0 , clip]
					{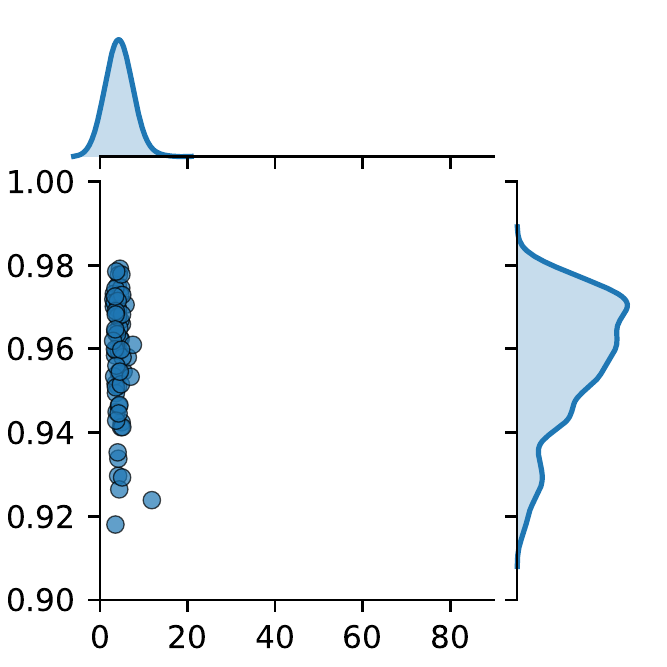}
					\put(0, 0){\textbf{\textcolor{black}{\large (11 - 160 filters)}}}
				\end{overpic}
  \end{minipage}}\end{minipage}
  \hfill
  \begin{minipage}{0.32\textwidth}\framebox{\begin{minipage}{\textwidth}
				\begin{overpic}[width=\textwidth, tics = 10, trim = 0 0 0 0 , clip]
					{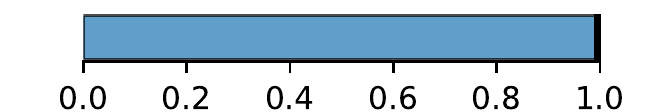}
				\end{overpic}
				\vspace*{2mm}
				\begin{overpic}[width=\textwidth, tics = 10, trim = 0 0 0 0 , clip]
					{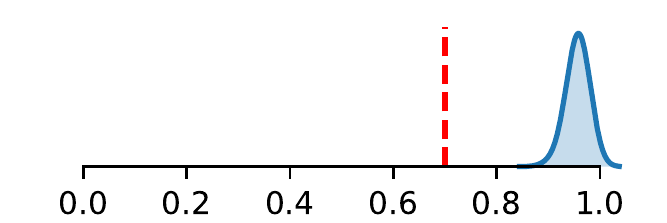}
				\end{overpic}
				\begin{overpic}[width=\textwidth, tics = 10, trim = 0 -10 0 0 , clip]
					{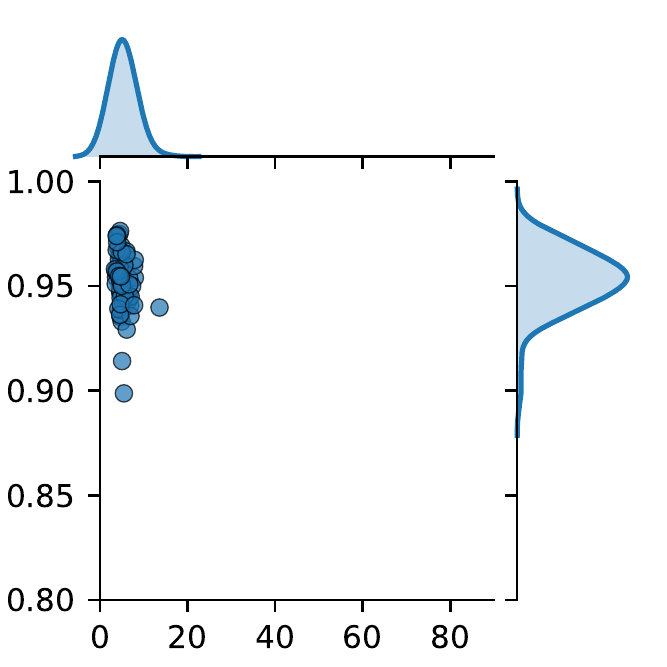}
					\put(0, 0){\textbf{\textcolor{black}{\large (11 - 192 filters)}}}
				\end{overpic}
  \end{minipage}}\end{minipage}
  \hfill
  \begin{minipage}{0.32\textwidth}\framebox{\begin{minipage}{\textwidth}
				\begin{overpic}[width=\textwidth, tics = 10, trim = 0 0 0 0 , clip]
					{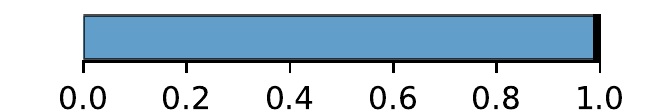}
				\end{overpic}
				\vspace*{2mm}
				\begin{overpic}[width=\textwidth, tics = 10, trim = 0 0 0 0 , clip]
					{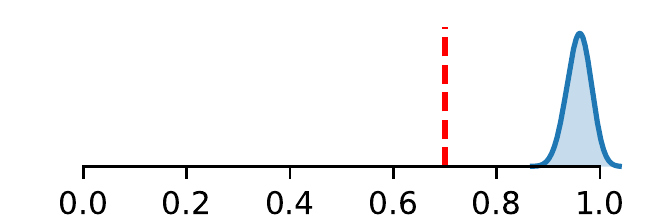}
				\end{overpic}
				\begin{overpic}[width=\textwidth, tics = 10, trim = 0 -10 0 0 , clip]
					{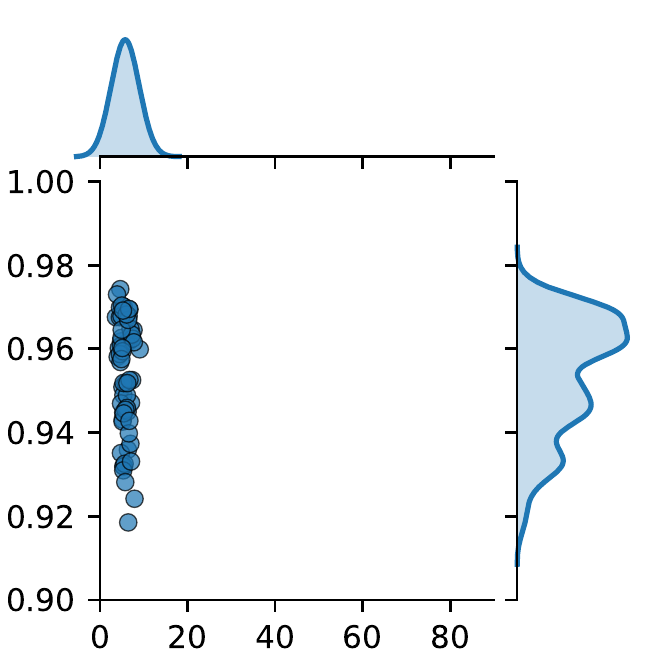}
					\put(0, 0){\textbf{\textcolor{black}{\large (11 - 224 filters)}}}
				\end{overpic}
  \end{minipage}}\end{minipage}

  \begin{minipage}{0.32\textwidth}\framebox{\begin{minipage}{\textwidth}
				\begin{overpic}[width=\textwidth, tics = 10, trim = 0 0 0 0 , clip]
					{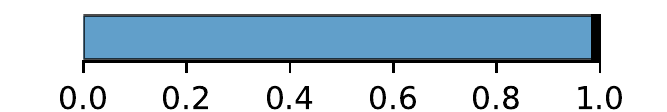}
				\end{overpic}
				\vspace*{2mm}
				\begin{overpic}[width=\textwidth, tics = 10, trim = 0 0 0 0 , clip]
					{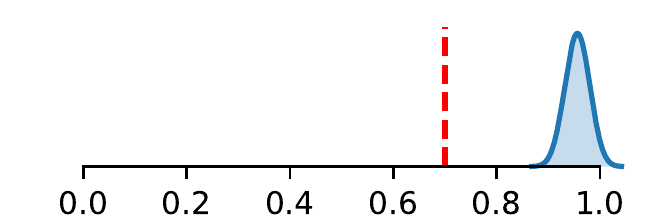}
				\end{overpic}
				\begin{overpic}[width=\textwidth, tics = 10, trim = 0 -10 0 0 , clip]
					{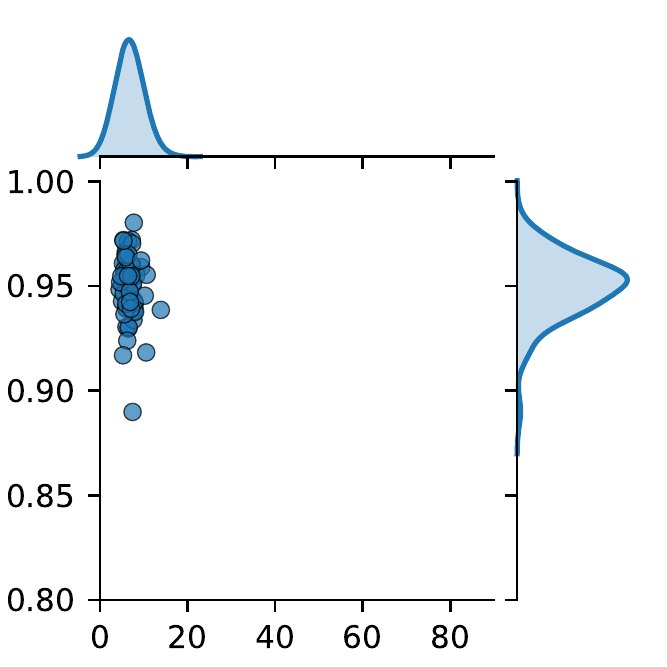}
					\put(0, 0){\textbf{\textcolor{black}{\large (11 - 256 filters)}}}
				\end{overpic}
  \end{minipage}}\end{minipage}
  \hfill
  \begin{minipage}{0.32\textwidth}\framebox{\begin{minipage}{\textwidth}
				\begin{overpic}[width=\textwidth, tics = 10, trim = 0 0 0 0 , clip]
					{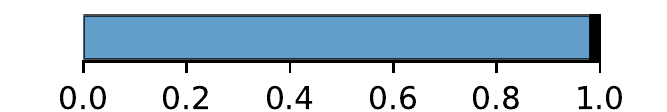}
				\end{overpic}
				\vspace*{2mm}
				\begin{overpic}[width=\textwidth, tics = 10, trim = 0 0 0 0 , clip]
					{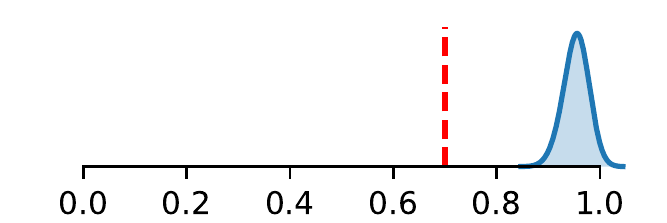}
				\end{overpic}
				\begin{overpic}[width=\textwidth, tics = 10, trim = 0 -10 0 0 , clip]
					{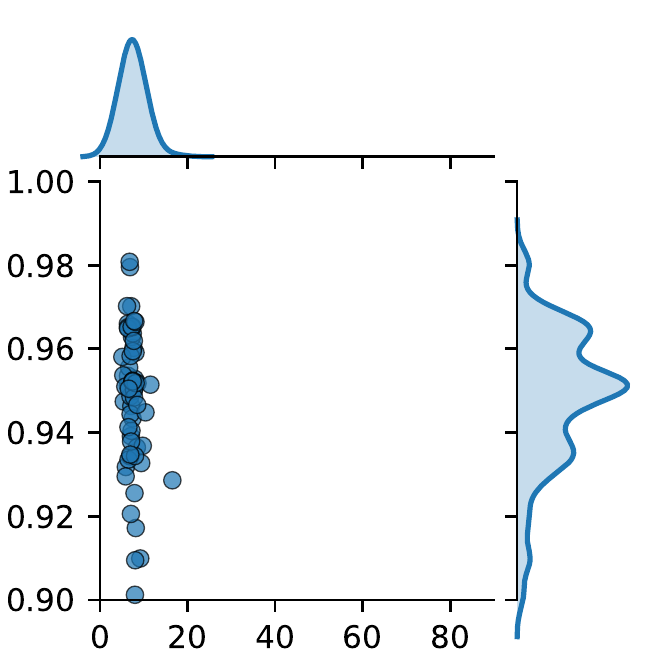}
					\put(0, 0){\textbf{\textcolor{black}{\large (11 - 288 filters)}}}
				\end{overpic}
  \end{minipage}}\end{minipage}
  \hfill
  \begin{minipage}{0.32\textwidth}\framebox{\begin{minipage}{\textwidth}
				\begin{overpic}[width=\textwidth, tics = 10, trim = 0 0 0 0 , clip]
					{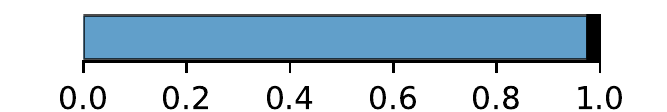}
				\end{overpic}
				\vspace*{2mm}
				\begin{overpic}[width=\textwidth, tics = 10, trim = 0 0 0 0 , clip]
					{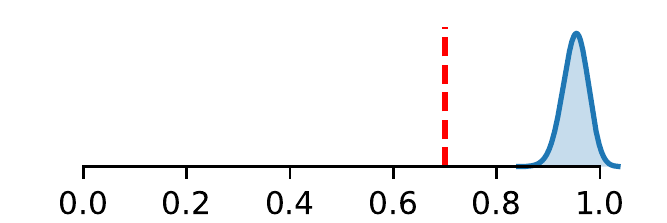}
				\end{overpic}
				\begin{overpic}[width=\textwidth, tics = 10, trim = 0 -10 0 0 , clip]
					{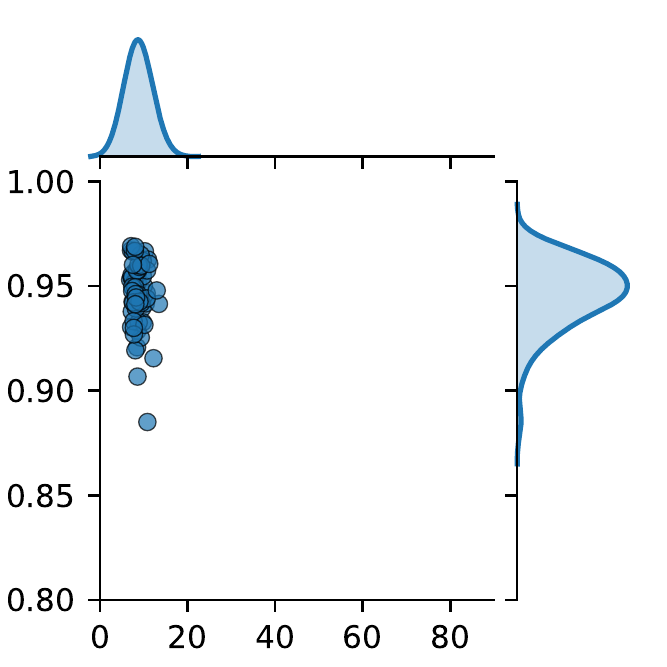}
					\put(0, 0){\textbf{\textcolor{black}{\large (11 - 320 filters)}}}
				\end{overpic}
  \end{minipage}}\end{minipage}

  \begin{minipage}{0.32\textwidth}\framebox{\begin{minipage}{\textwidth}
				\begin{overpic}[width=\textwidth, tics = 10, trim = 0 0 0 0 , clip]
					{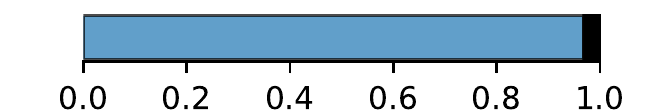}
				\end{overpic}
				\vspace*{2mm}
				\begin{overpic}[width=\textwidth, tics = 10, trim = 0 0 0 0 , clip]
					{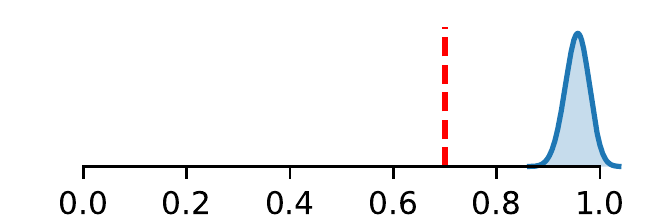}
				\end{overpic}
				\begin{overpic}[width=\textwidth, tics = 10, trim = 0 -10 0 0 , clip]
					{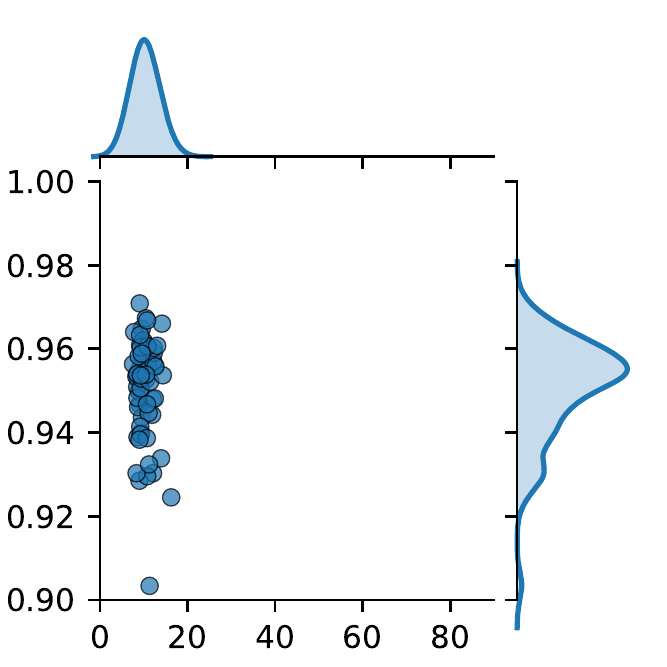}
					\put(0, 0){\textbf{\textcolor{black}{\large (11 - 342 filters)}}}
				\end{overpic}
  \end{minipage}}\end{minipage}
  \hfill
  \begin{minipage}{0.32\textwidth}\framebox{\begin{minipage}{\textwidth}
				\begin{overpic}[width=\textwidth, tics = 10, trim = 0 0 0 0 , clip]
					{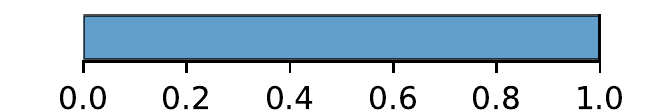}
				\end{overpic}
				\vspace*{2mm}
				\begin{overpic}[width=\textwidth, tics = 10, trim = 0 0 0 0 , clip]
					{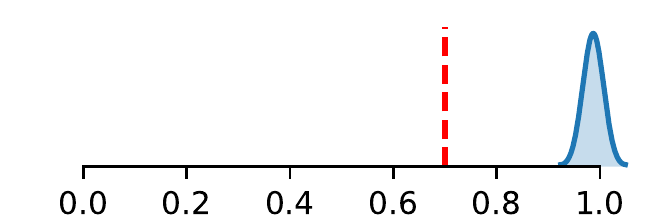}
				\end{overpic}
				\begin{overpic}[width=\textwidth, tics = 10, trim = 0 -10 0 0 , clip]
					{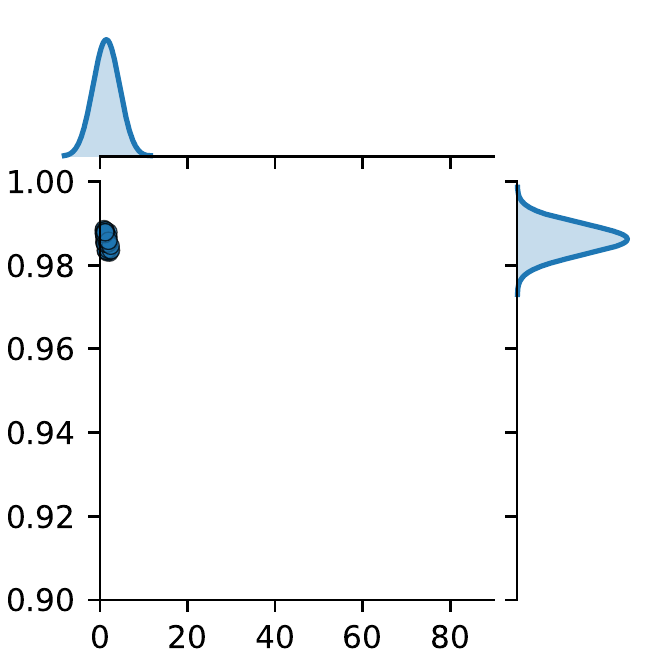}
					\put(0, 0){\textbf{\textcolor{black}{\large (12 - 32 filters)}}}
				\end{overpic}
  \end{minipage}}\end{minipage}
  \hfill
  \begin{minipage}{0.32\textwidth}\framebox{\begin{minipage}{\textwidth}
				\begin{overpic}[width=\textwidth, tics = 10, trim = 0 0 0 0 , clip]
					{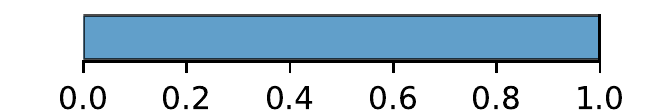}
				\end{overpic}
				\vspace*{2mm}
				\begin{overpic}[width=\textwidth, tics = 10, trim = 0 0 0 0 , clip]
					{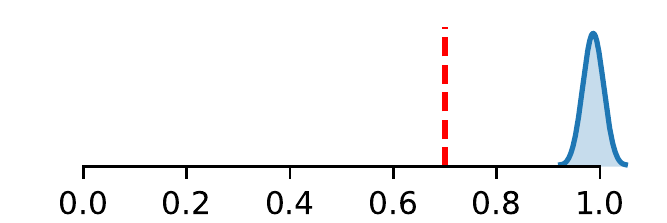}
				\end{overpic}
				\begin{overpic}[width=\textwidth, tics = 10, trim = 0 -10 0 0 , clip]
					{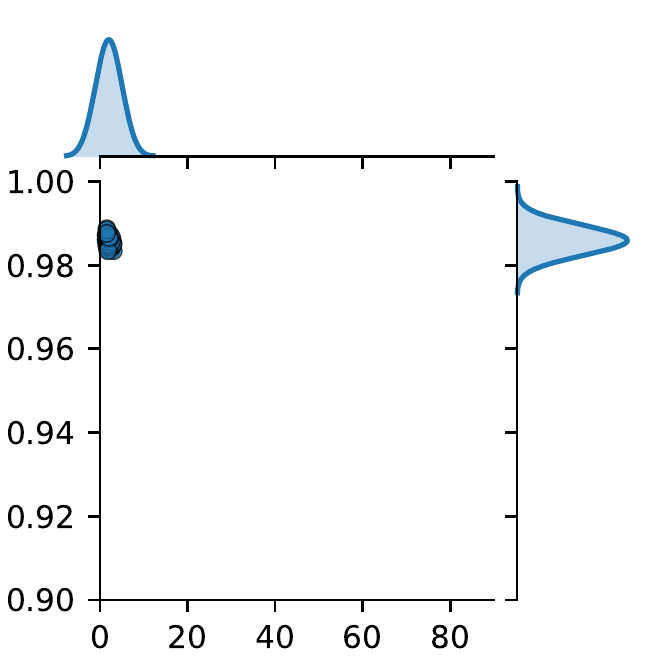}
					\put(0, 0){\textbf{\textcolor{black}{\large (12 - 64 filters)}}}
				\end{overpic}
  \end{minipage}}\end{minipage}

  \begin{minipage}{0.32\textwidth}\framebox{\begin{minipage}{\textwidth}
				\begin{overpic}[width=\textwidth, tics = 10, trim = 0 0 0 0 , clip]
					{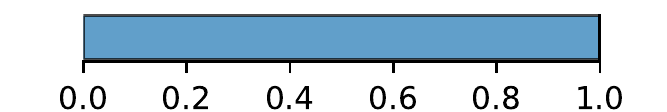}
				\end{overpic}
				\vspace*{2mm}
				\begin{overpic}[width=\textwidth, tics = 10, trim = 0 0 0 0 , clip]
					{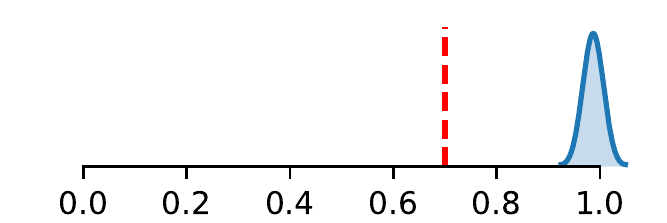}
				\end{overpic}
				\begin{overpic}[width=\textwidth, tics = 10, trim = 0 -10 0 0 , clip]
					{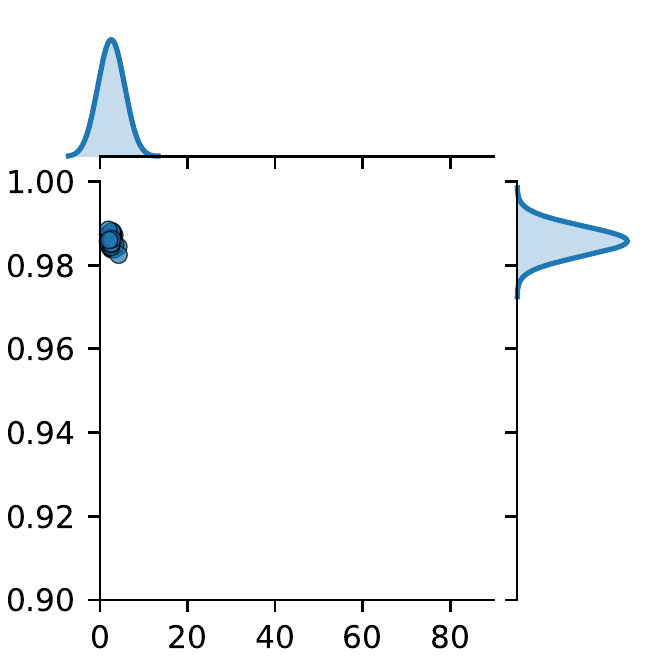}
					\put(0, 0){\textbf{\textcolor{black}{\large (12 - 96 filters)}}}
				\end{overpic}
  \end{minipage}}\end{minipage}
  \hfill
  \begin{minipage}{0.32\textwidth}\framebox{\begin{minipage}{\textwidth}
				\begin{overpic}[width=\textwidth, tics = 10, trim = 0 0 0 0 , clip]
					{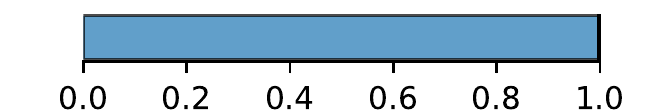}
				\end{overpic}
				\vspace*{2mm}
				\begin{overpic}[width=\textwidth, tics = 10, trim = 0 0 0 0 , clip]
					{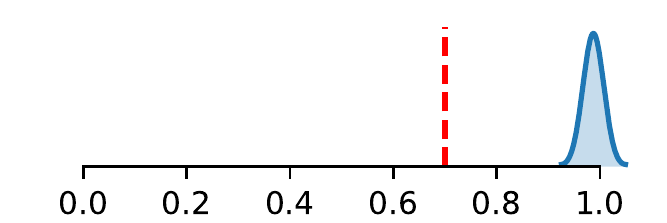}
				\end{overpic}
				\begin{overpic}[width=\textwidth, tics = 10, trim = 0 -10 0 0 , clip]
					{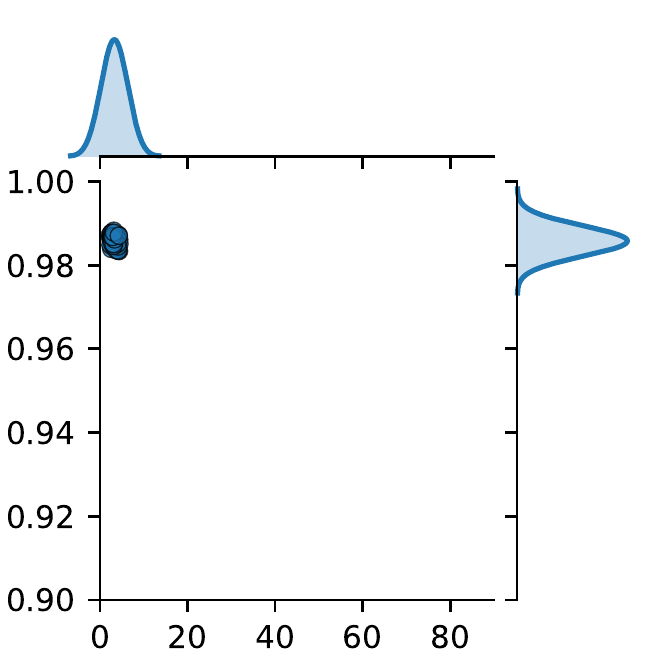}
					\put(0, 0){\textbf{\textcolor{black}{\large (12 - 128 filters)}}}
				\end{overpic}
  \end{minipage}}\end{minipage}
  \hfill
  \begin{minipage}{0.32\textwidth}\framebox{\begin{minipage}{\textwidth}
				\begin{overpic}[width=\textwidth, tics = 10, trim = 0 0 0 0 , clip]
					{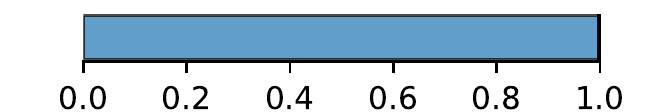}
				\end{overpic}
				\vspace*{2mm}
				\begin{overpic}[width=\textwidth, tics = 10, trim = 0 0 0 0 , clip]
					{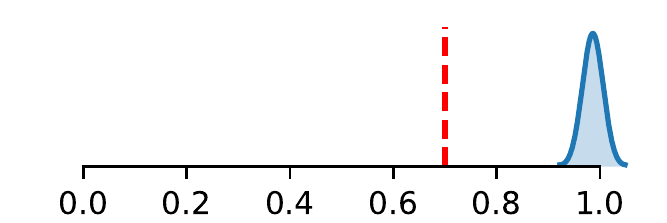}
				\end{overpic}
				\begin{overpic}[width=\textwidth, tics = 10, trim = 0 -10 0 0 , clip]
					{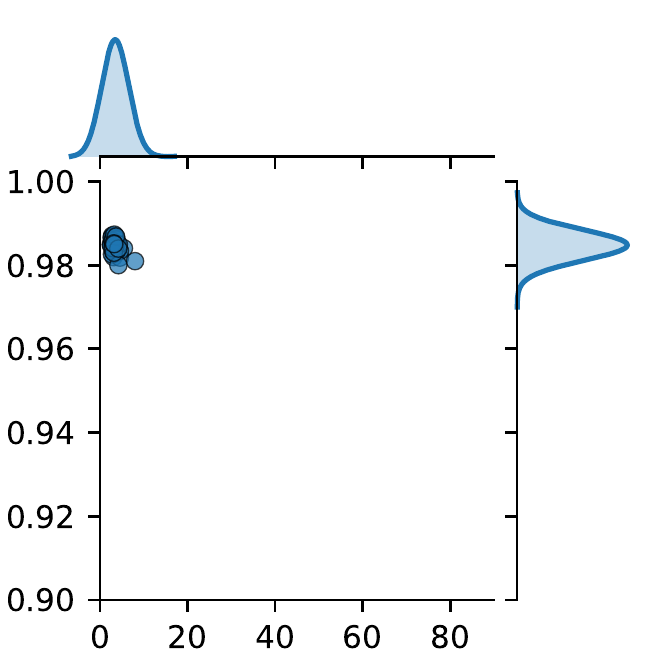}
					\put(0, 0){\textbf{\textcolor{black}{\large (12 - 160 filters)}}}
				\end{overpic}
  \end{minipage}}\end{minipage}

  \begin{minipage}{0.32\textwidth}\framebox{\begin{minipage}{\textwidth}
				\begin{overpic}[width=\textwidth, tics = 10, trim = 0 0 0 0 , clip]
					{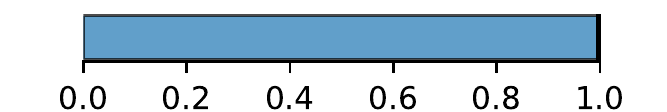}
				\end{overpic}
				\vspace*{2mm}
				\begin{overpic}[width=\textwidth, tics = 10, trim = 0 0 0 0 , clip]
					{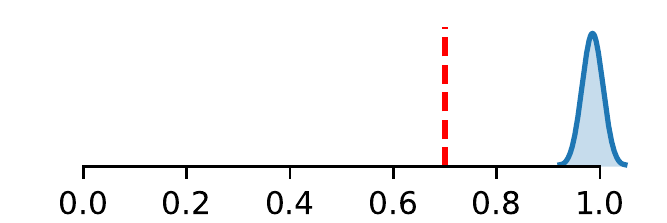}
				\end{overpic}
				\begin{overpic}[width=\textwidth, tics = 10, trim = 0 -10 0 0 , clip]
					{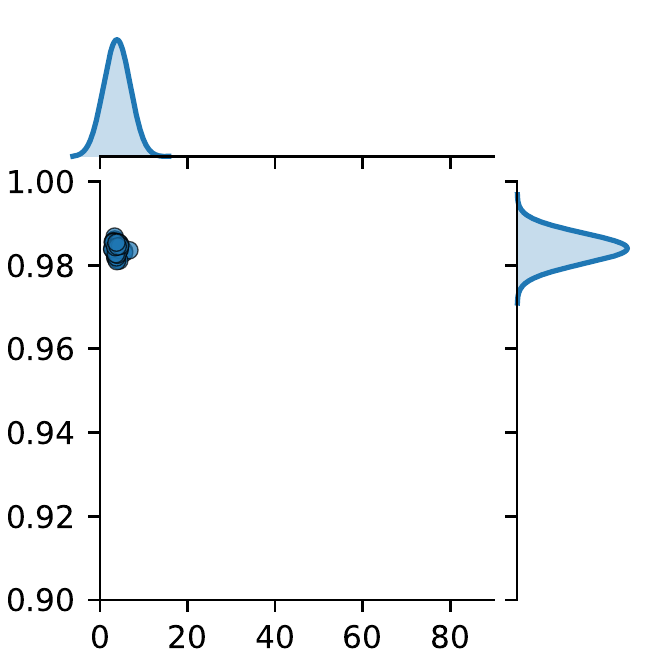}
					\put(0, 0){\textbf{\textcolor{black}{\large (12 - 192 filters)}}}
				\end{overpic}
  \end{minipage}}\end{minipage}
  \hfill
  \begin{minipage}{0.32\textwidth}\framebox{\begin{minipage}{\textwidth}
				\begin{overpic}[width=\textwidth, tics = 10, trim = 0 0 0 0 , clip]
					{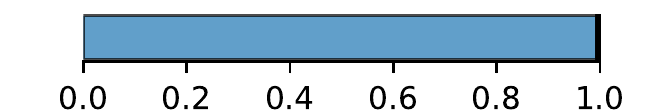}
				\end{overpic}
				\vspace*{2mm}
				\begin{overpic}[width=\textwidth, tics = 10, trim = 0 0 0 0 , clip]
					{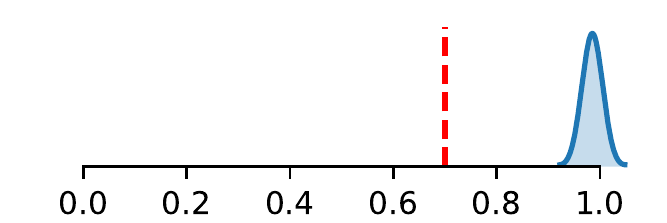}
				\end{overpic}
				\begin{overpic}[width=\textwidth, tics = 10, trim = 0 -10 0 0 , clip]
					{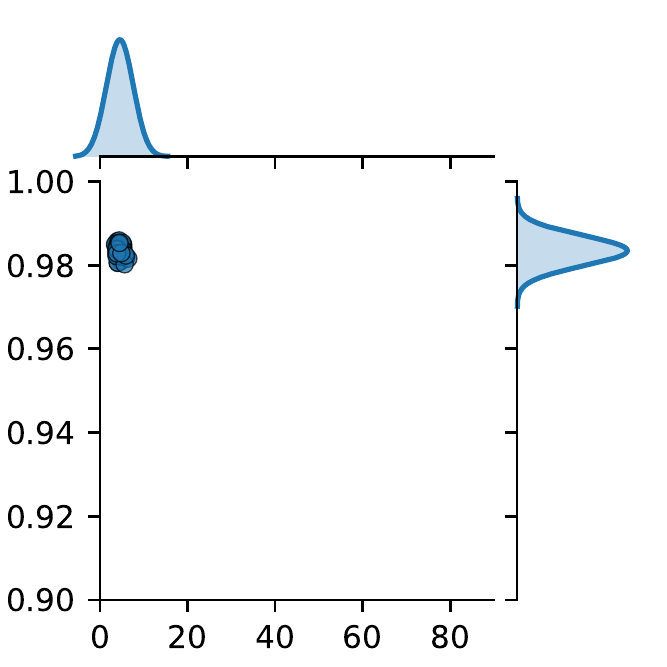}
					\put(0, 0){\textbf{\textcolor{black}{\large (12 - 224 filters)}}}
				\end{overpic}
  \end{minipage}}\end{minipage}
  \hfill
  \begin{minipage}{0.32\textwidth}\framebox{\begin{minipage}{\textwidth}
				\begin{overpic}[width=\textwidth, tics = 10, trim = 0 0 0 0 , clip]
					{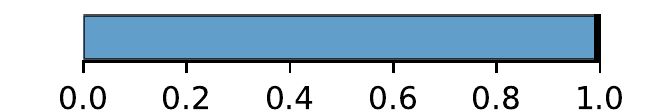}
				\end{overpic}
				\vspace*{2mm}
				\begin{overpic}[width=\textwidth, tics = 10, trim = 0 0 0 0 , clip]
					{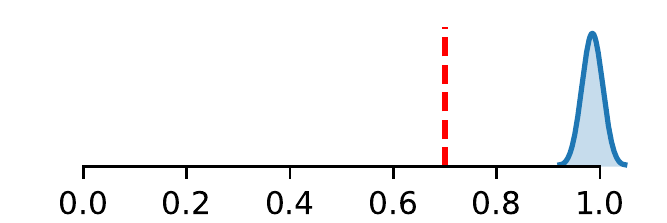}
				\end{overpic}
				\begin{overpic}[width=\textwidth, tics = 10, trim = 0 -10 0 0 , clip]
					{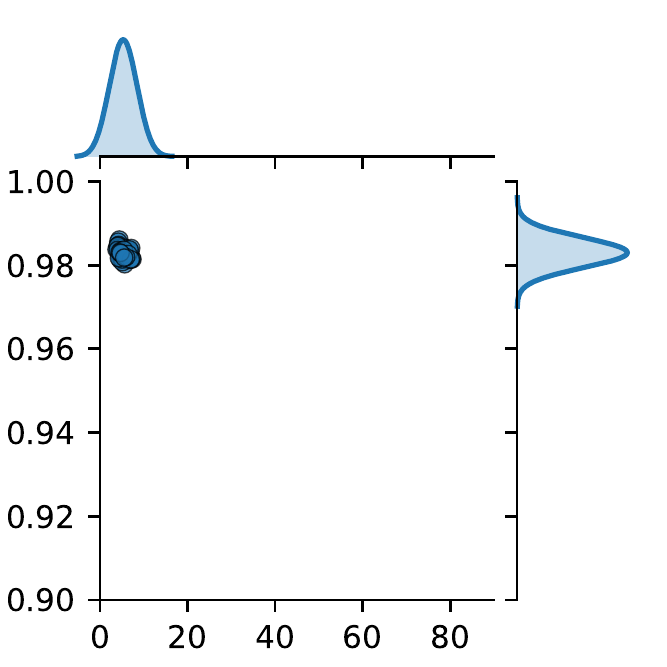}
					\put(0, 0){\textbf{\textcolor{black}{\large (12 - 256 filters)}}}
				\end{overpic}
  \end{minipage}}\end{minipage}

  \begin{minipage}{0.32\textwidth}\framebox{\begin{minipage}{\textwidth}
				\begin{overpic}[width=\textwidth, tics = 10, trim = 0 0 0 0 , clip]
					{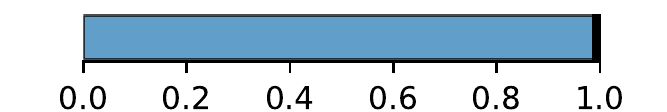}
				\end{overpic}
				\vspace*{2mm}
				\begin{overpic}[width=\textwidth, tics = 10, trim = 0 0 0 0 , clip]
					{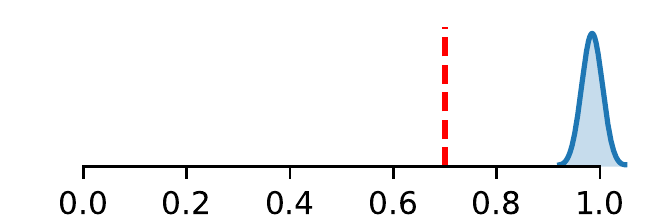}
				\end{overpic}
				\begin{overpic}[width=\textwidth, tics = 10, trim = 0 -10 0 0 , clip]
					{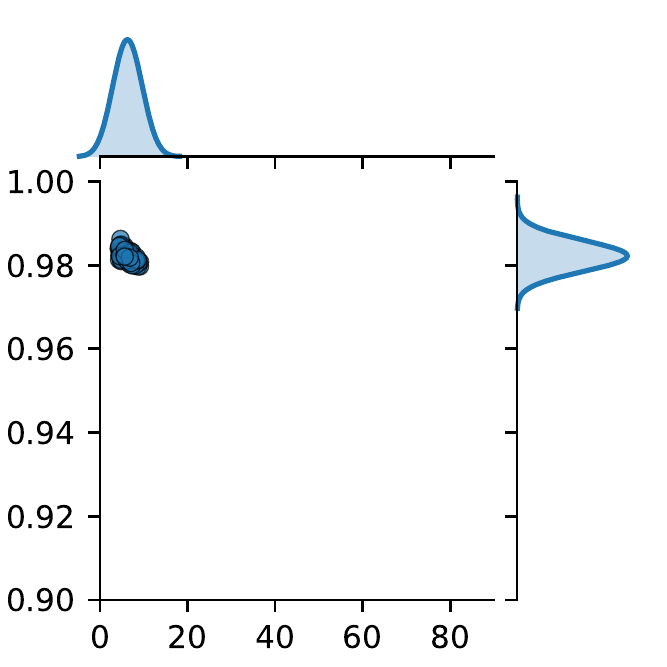}
					\put(0, 0){\textbf{\textcolor{black}{\large (12 - 288 filters)}}}
				\end{overpic}
  \end{minipage}}\end{minipage}
  \hfill
  \begin{minipage}{0.32\textwidth}\framebox{\begin{minipage}{\textwidth}
				\begin{overpic}[width=\textwidth, tics = 10, trim = 0 0 0 0 , clip]
					{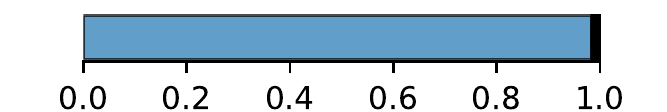}
				\end{overpic}
				\vspace*{2mm}
				\begin{overpic}[width=\textwidth, tics = 10, trim = 0 0 0 0 , clip]
					{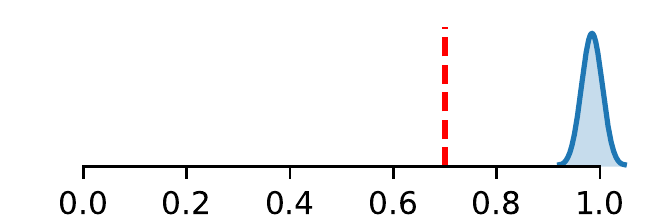}
				\end{overpic}
				\begin{overpic}[width=\textwidth, tics = 10, trim = 0 -10 0 0 , clip]
					{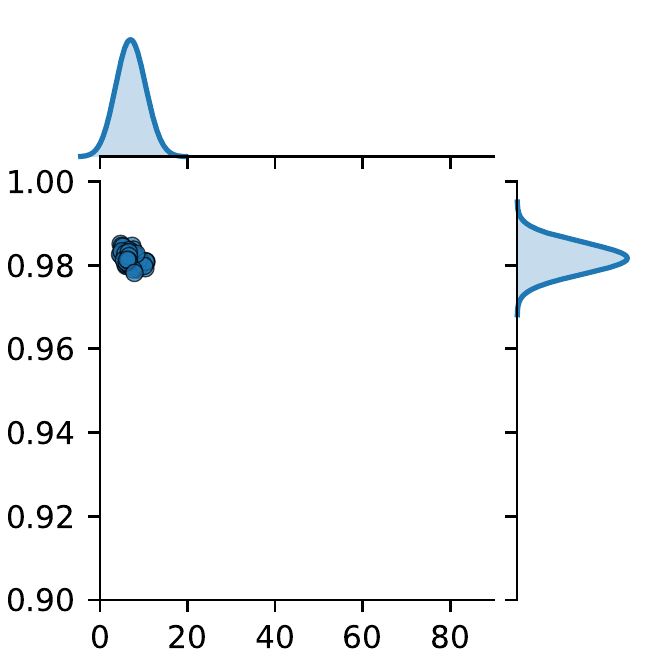}
					\put(0, 0){\textbf{\textcolor{black}{\large (12 - 320 filters)}}}
				\end{overpic}
  \end{minipage}}\end{minipage}
  \hfill
  \begin{minipage}{0.32\textwidth}\framebox{\begin{minipage}{\textwidth}
				\begin{overpic}[width=\textwidth, tics = 10, trim = 0 0 0 0 , clip]
					{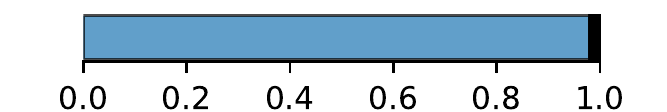}
				\end{overpic}
				\vspace*{2mm}
				\begin{overpic}[width=\textwidth, tics = 10, trim = 0 0 0 0 , clip]
					{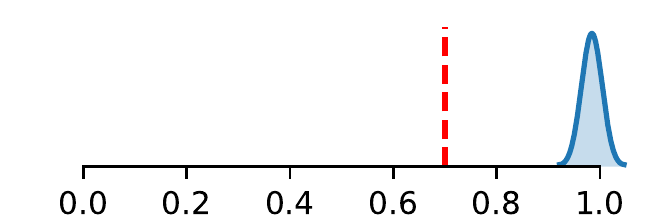}
				\end{overpic}
				\begin{overpic}[width=\textwidth, tics = 10, trim = 0 -10 0 0 , clip]
					{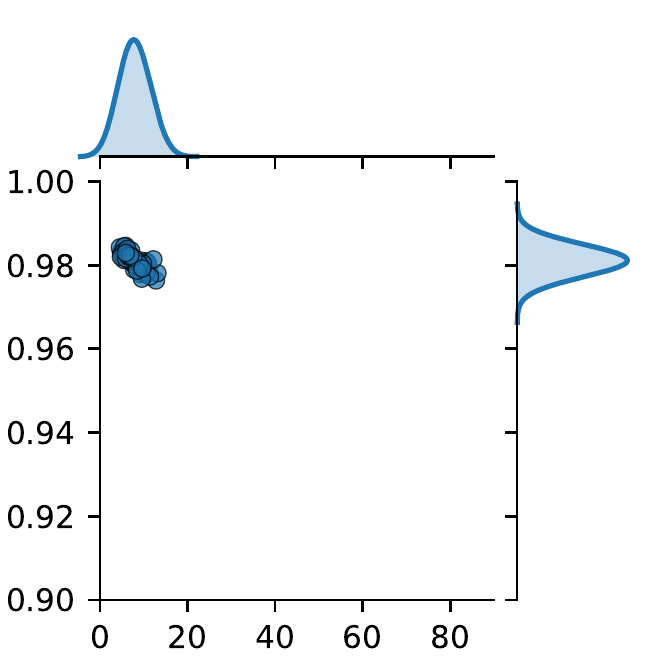}
					\put(0, 0){\textbf{\textcolor{black}{\large (12 - 352 filters)}}}
				\end{overpic}
  \end{minipage}}\end{minipage}

  \begin{minipage}{0.32\textwidth}\framebox{\begin{minipage}{\textwidth}
				\begin{overpic}[width=\textwidth, tics = 10, trim = 0 0 0 0 , clip]
					{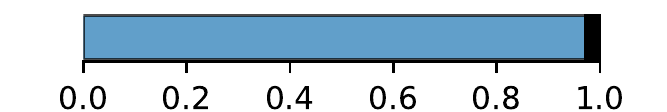}
				\end{overpic}
				\vspace*{2mm}
				\begin{overpic}[width=\textwidth, tics = 10, trim = 0 0 0 0 , clip]
					{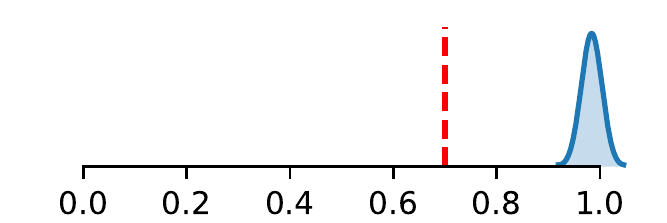}
				\end{overpic}
				\begin{overpic}[width=\textwidth, tics = 10, trim = 0 -10 0 0 , clip]
					{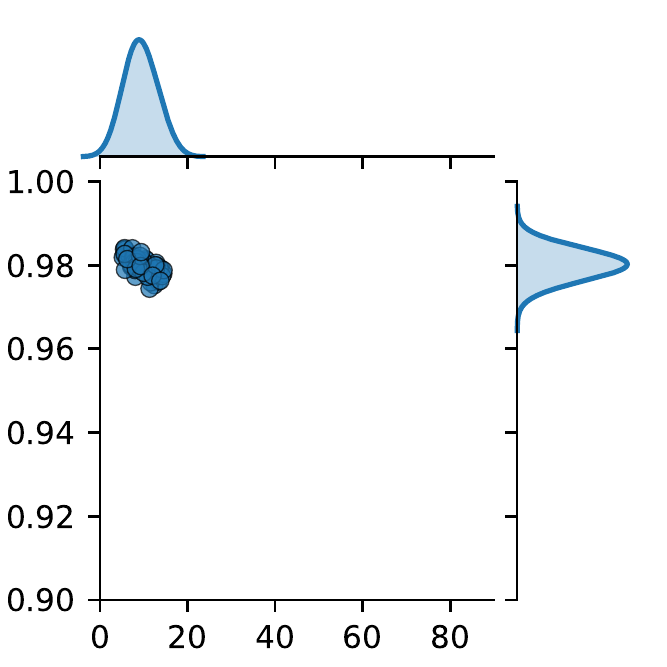}
					\put(0, 0){\textbf{\textcolor{black}{\large (12 - 384 filters)}}}
				\end{overpic}
  \end{minipage}}\end{minipage}
  \hfill
  \begin{minipage}{0.32\textwidth}\framebox{\begin{minipage}{\textwidth}
				\begin{overpic}[width=\textwidth, tics = 10, trim = 0 0 0 0 , clip]
					{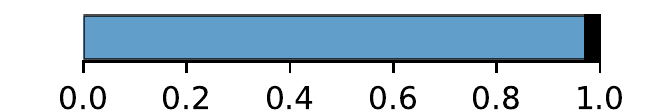}
				\end{overpic}
				\vspace*{2mm}
				\begin{overpic}[width=\textwidth, tics = 10, trim = 0 0 0 0 , clip]
					{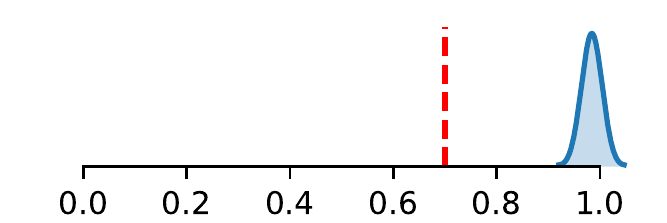}
				\end{overpic}
				\begin{overpic}[width=\textwidth, tics = 10, trim = 0 -10 0 0 , clip]
					{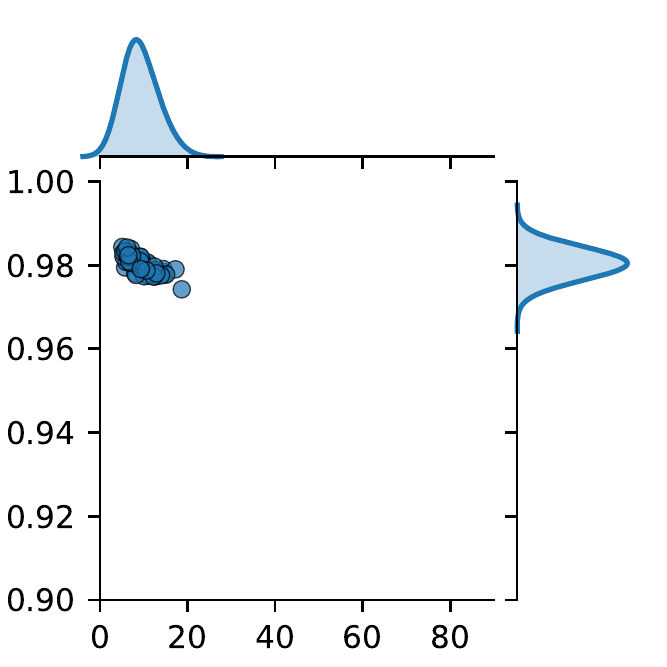}
					\put(0, 0){\textbf{\textcolor{black}{\large (12 - 416 filters)}}}
				\end{overpic}
  \end{minipage}}\end{minipage}
  \hfill
  \begin{minipage}{0.32\textwidth}\framebox{\begin{minipage}{\textwidth}
				\begin{overpic}[width=\textwidth, tics = 10, trim = 0 0 0 0 , clip]
					{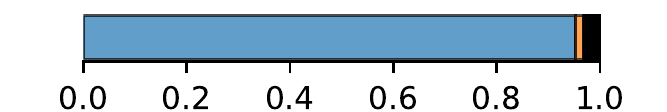}
				\end{overpic}
				\vspace*{2mm}
				\begin{overpic}[width=\textwidth, tics = 10, trim = 0 0 0 0 , clip]
					{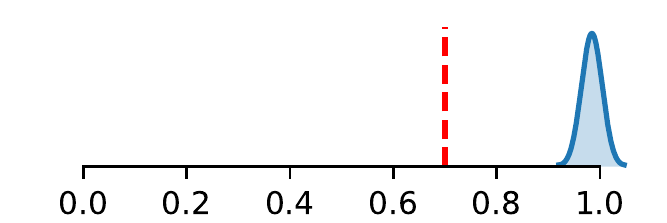}
				\end{overpic}
				\begin{overpic}[width=\textwidth, tics = 10, trim = 0 -10 0 0 , clip]
					{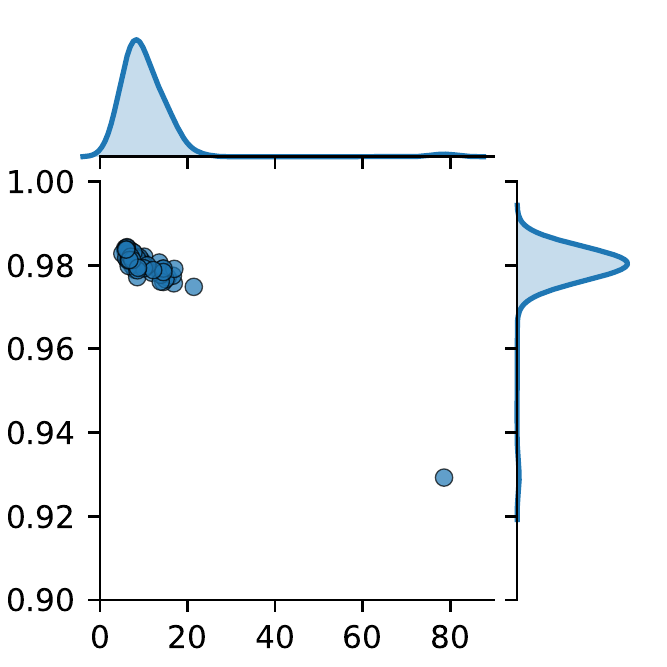}
					\put(0, 0){\textbf{\textcolor{black}{\large (12 - 448 filters)}}}
				\end{overpic}
  \end{minipage}}\end{minipage}

  \begin{minipage}{0.32\textwidth}\framebox{\begin{minipage}{\textwidth}
				\begin{overpic}[width=\textwidth, tics = 10, trim = 0 0 0 0 , clip]
					{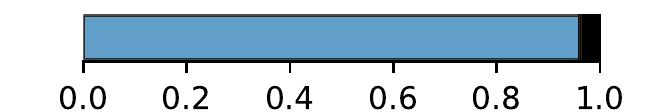}
				\end{overpic}
				\vspace*{2mm}
				\begin{overpic}[width=\textwidth, tics = 10, trim = 0 0 0 0 , clip]
					{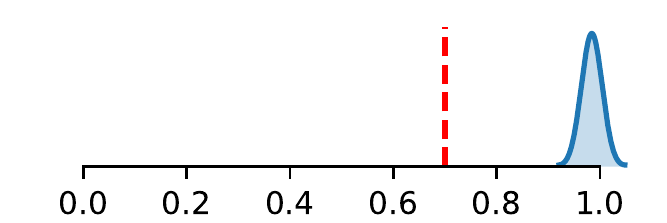}
				\end{overpic}
				\begin{overpic}[width=\textwidth, tics = 10, trim = 0 -10 0 0 , clip]
					{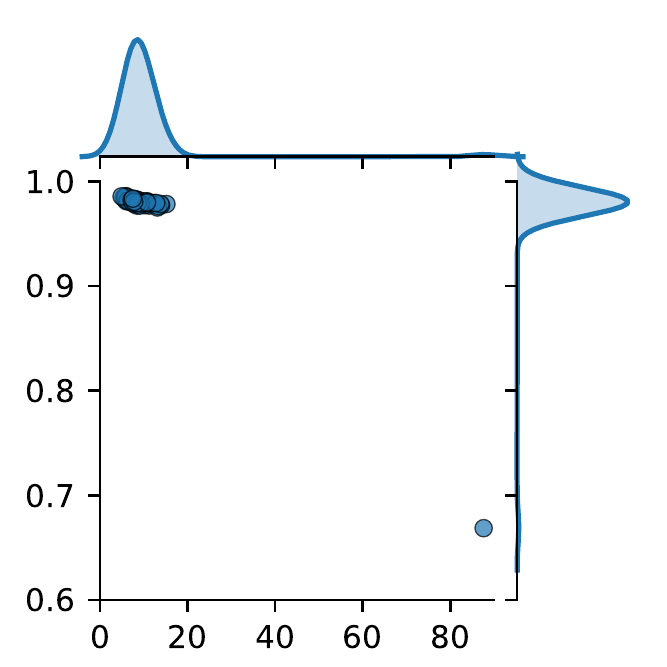}
					\put(0, 0){\textbf{\textcolor{black}{\large (12 - 480 filters)}}}
				\end{overpic}
  \end{minipage}}\end{minipage}
  \hfill
  \begin{minipage}{0.32\textwidth}\framebox{\begin{minipage}{\textwidth}
				\begin{overpic}[width=\textwidth, tics = 10, trim = 0 0 0 0 , clip]
					{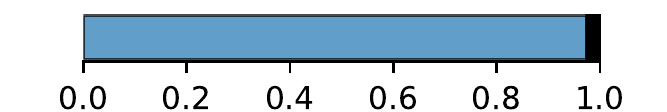}
				\end{overpic}
				\vspace*{2mm}
				\begin{overpic}[width=\textwidth, tics = 10, trim = 0 0 0 0 , clip]
					{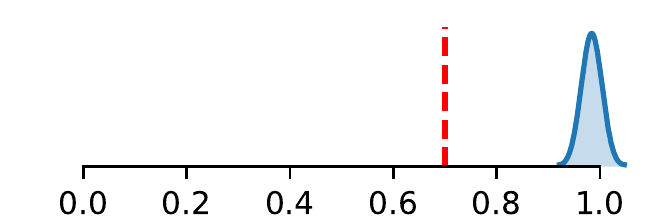}
				\end{overpic}
				\begin{overpic}[width=\textwidth, tics = 10, trim = 0 -10 0 0 , clip]
					{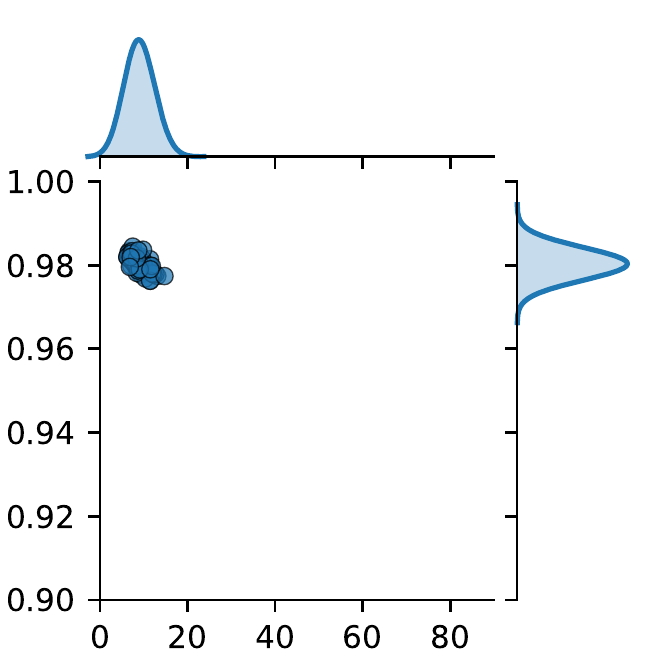}
					\put(0, 0){\textbf{\textcolor{black}{\large (12 - 512 filters)}}}
				\end{overpic}
  \end{minipage}}\end{minipage}
  \hfill
  \begin{minipage}{0.32\textwidth}\framebox{\begin{minipage}{\textwidth}
				\begin{overpic}[width=\textwidth, tics = 10, trim = 0 0 0 0 , clip]
					{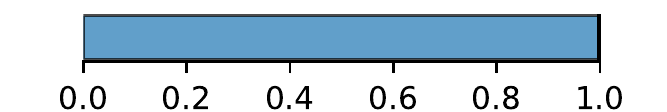}
				\end{overpic}
				\vspace*{2mm}
				\begin{overpic}[width=\textwidth, tics = 10, trim = 0 0 0 0 , clip]
					{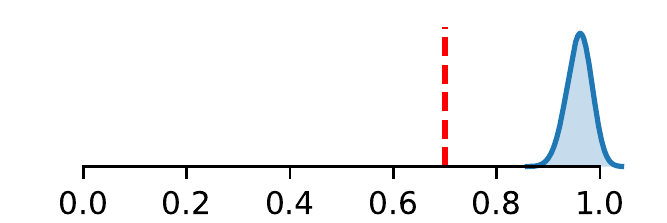}
				\end{overpic}
				\begin{overpic}[width=\textwidth, tics = 10, trim = 0 -10 0 0 , clip]
					{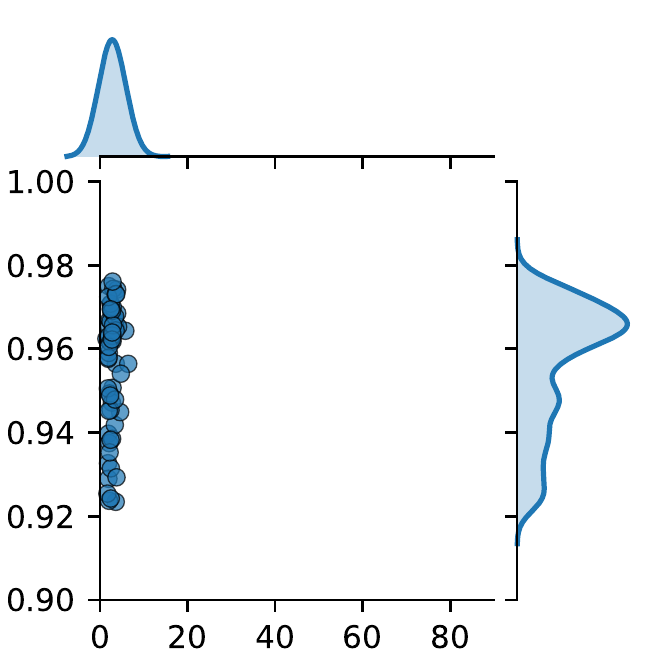}
					\put(0, 0){\textbf{\textcolor{black}{\large (13)}}}
				\end{overpic}
  \end{minipage}}\end{minipage}

  \begin{minipage}{0.32\textwidth}\framebox{\begin{minipage}{\textwidth}
				\begin{overpic}[width=\textwidth, tics = 10, trim = 0 0 0 0 , clip]
					{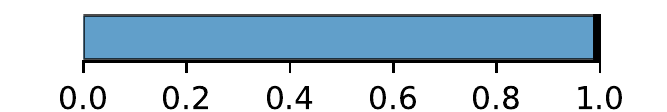}
				\end{overpic}
				\vspace*{2mm}
				\begin{overpic}[width=\textwidth, tics = 10, trim = 0 0 0 0 , clip]
					{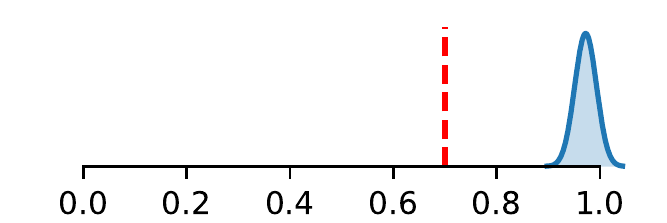}
				\end{overpic}
				\begin{overpic}[width=\textwidth, tics = 10, trim = 0 -10 0 0 , clip]
					{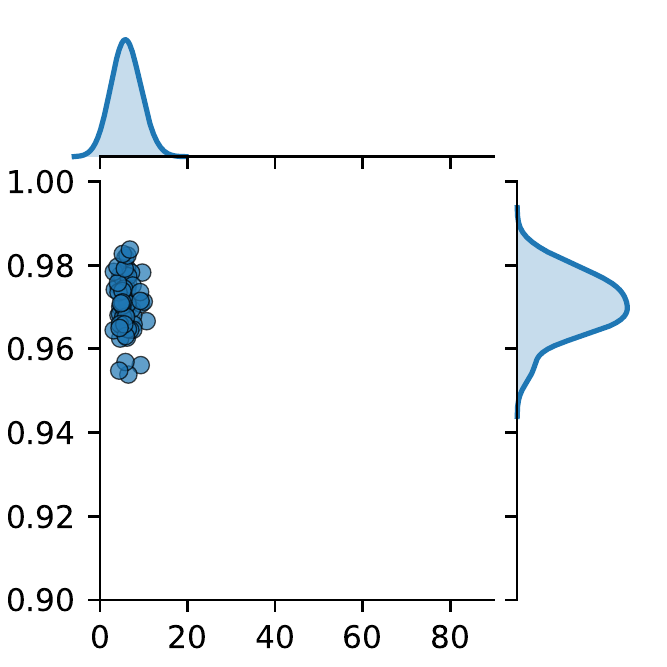}
					\put(0, 0){\textbf{\textcolor{black}{\large (14)}}}
				\end{overpic}
  \end{minipage}}\end{minipage}
  \hfill
  \begin{minipage}{0.32\textwidth}\framebox{\begin{minipage}{\textwidth}
				\begin{overpic}[width=\textwidth, tics = 10, trim = 0 0 0 0 , clip]
					{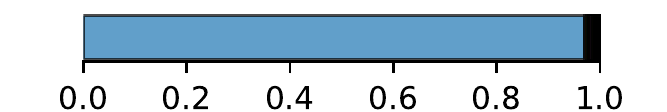}
				\end{overpic}
				\vspace*{2mm}
				\begin{overpic}[width=\textwidth, tics = 10, trim = 0 0 0 0 , clip]
					{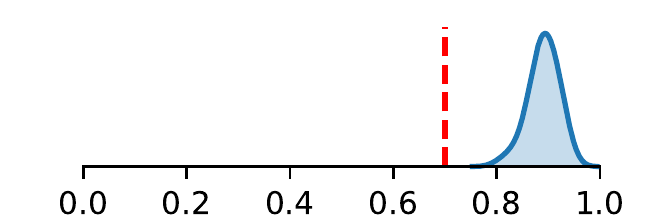}
				\end{overpic}
				\begin{overpic}[width=\textwidth, tics = 10, trim = 0 -10 0 0 , clip]
					{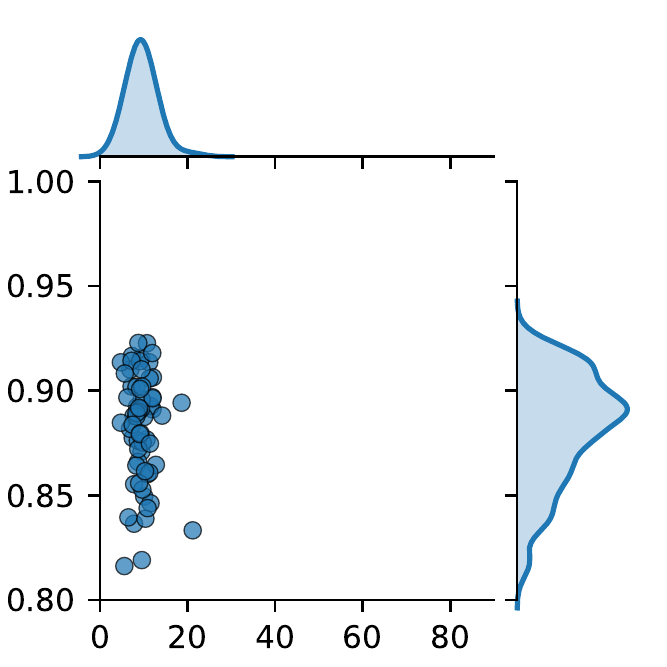}
					\put(0, 0){\textbf{\textcolor{black}{\large (15)}}}
				\end{overpic}
  \end{minipage}}\end{minipage}
  \hfill
  \begin{minipage}{0.32\textwidth}\framebox{\begin{minipage}{\textwidth}
				\begin{overpic}[width=\textwidth, tics = 10, trim = 0 0 0 0 , clip]
					{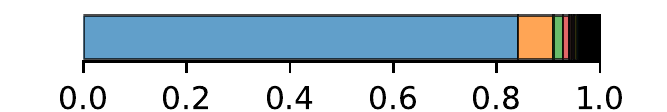}
				\end{overpic}
				\vspace*{2mm}
				\begin{overpic}[width=\textwidth, tics = 10, trim = 0 0 0 0 , clip]
					{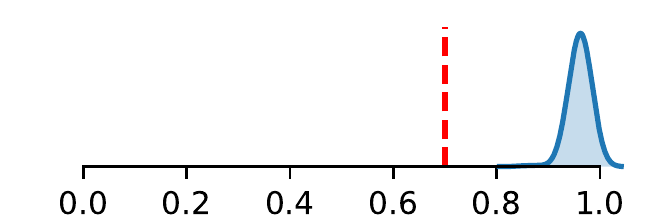}
				\end{overpic}
				\begin{overpic}[width=\textwidth, tics = 10, trim = 0 -10 0 0 , clip]
					{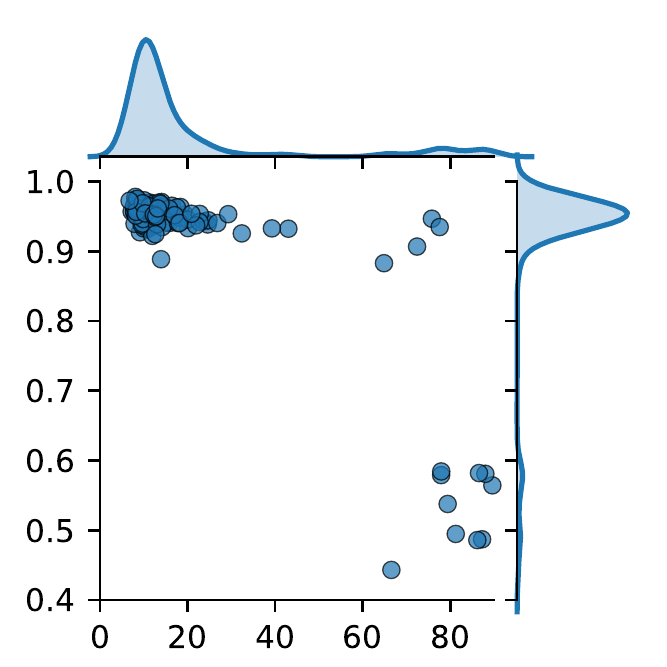}
					\put(0, 0){\textbf{\textcolor{black}{\large (16)}}}
				\end{overpic}
  \end{minipage}}\end{minipage}

  \begin{minipage}{0.32\textwidth}\framebox{\begin{minipage}{\textwidth}
				\begin{overpic}[width=\textwidth, tics = 10, trim = 0 0 0 0 , clip]
					{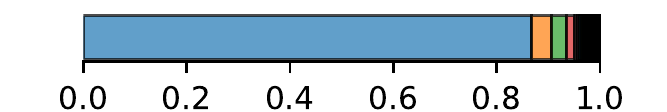}
				\end{overpic}
				\vspace*{2mm}
				\begin{overpic}[width=\textwidth, tics = 10, trim = 0 0 0 0 , clip]
					{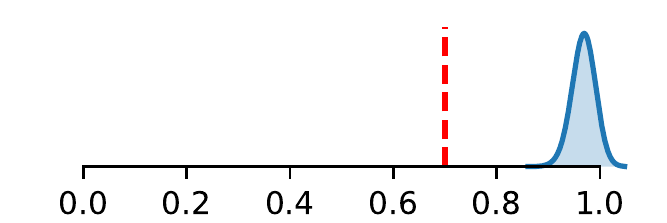}
				\end{overpic}
				\begin{overpic}[width=\textwidth, tics = 10, trim = 0 -10 0 0 , clip]
					{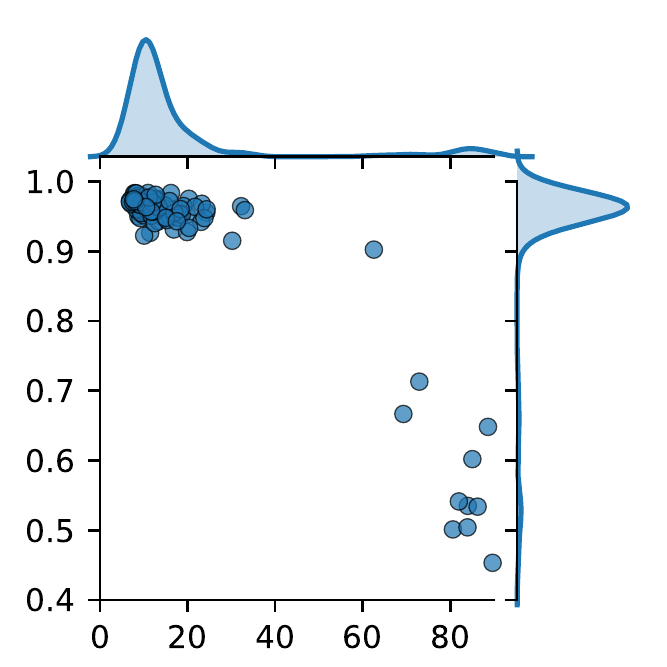}
					\put(0, 0){\textbf{\textcolor{black}{\large (17)}}}
				\end{overpic}
  \end{minipage}}\end{minipage}
  \hfill
  \begin{minipage}{0.32\textwidth}\framebox{\begin{minipage}{\textwidth}
				\begin{overpic}[width=\textwidth, tics = 10, trim = 0 0 0 0 , clip]
					{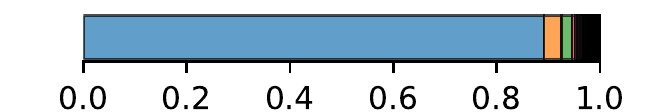}
				\end{overpic}
				\vspace*{2mm}
				\begin{overpic}[width=\textwidth, tics = 10, trim = 0 0 0 0 , clip]
					{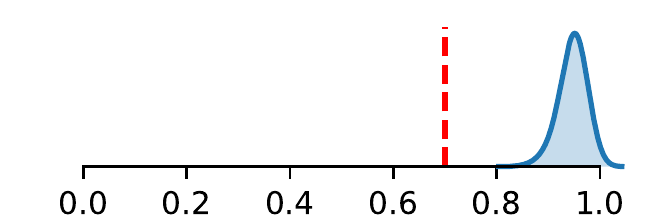}
				\end{overpic}
				\begin{overpic}[width=\textwidth, tics = 10, trim = 0 -10 0 0 , clip]
					{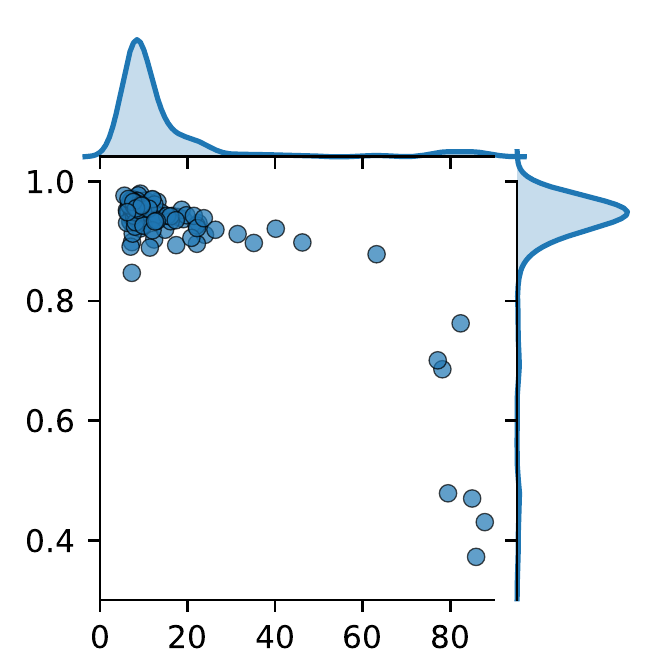}
					\put(0, 0){\textbf{\textcolor{black}{\large (18)}}}
				\end{overpic}
  \end{minipage}}\end{minipage}
  \hfill
  \begin{minipage}{0.32\textwidth}\framebox{\begin{minipage}{\textwidth}
				\begin{overpic}[width=\textwidth, tics = 10, trim = 0 0 0 0 , clip]
					{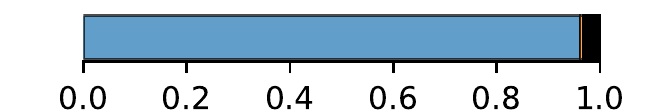}
				\end{overpic}
				\vspace*{2mm}
				\begin{overpic}[width=\textwidth, tics = 10, trim = 0 0 0 0 , clip]
					{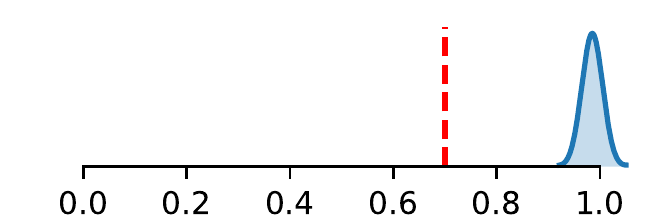}
				\end{overpic}
				\begin{overpic}[width=\textwidth, tics = 10, trim = 0 -10 0 0 , clip]
					{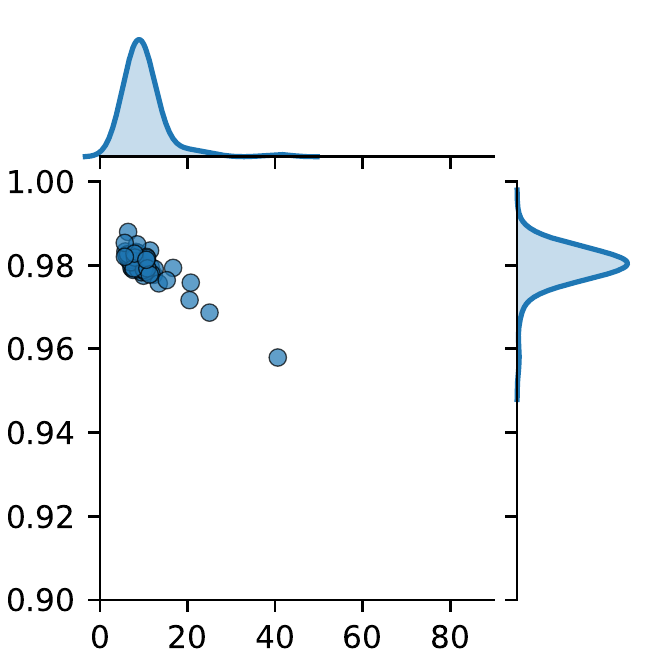}
					\put(0, 0){\textbf{\textcolor{black}{\large (19)}}}
				\end{overpic}
  \end{minipage}}\end{minipage}

  \begin{minipage}{0.32\textwidth}\framebox{\begin{minipage}{\textwidth}
				\begin{overpic}[width=\textwidth, tics = 10, trim = 0 0 0 0 , clip]
					{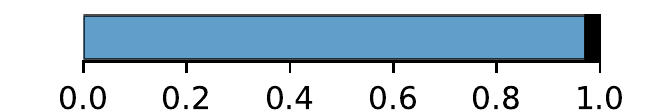}
				\end{overpic}
				\vspace*{2mm}
				\begin{overpic}[width=\textwidth, tics = 10, trim = 0 0 0 0 , clip]
					{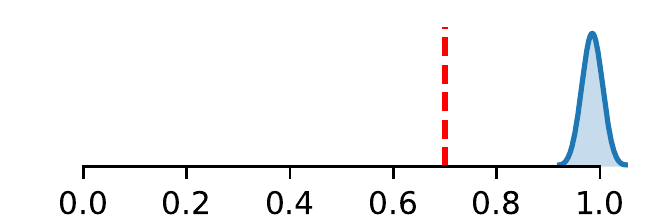}
				\end{overpic}
				\begin{overpic}[width=\textwidth, tics = 10, trim = 0 -10 0 0 , clip]
					{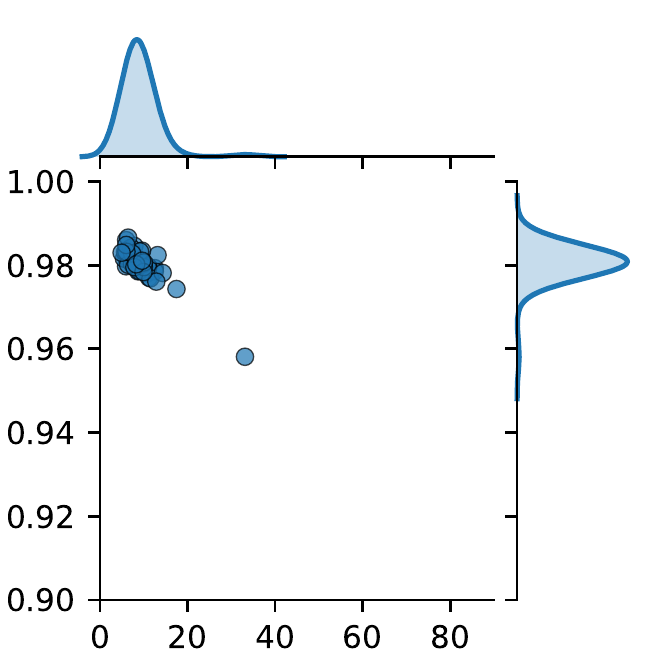}
					\put(0, 0){\textbf{\textcolor{black}{\large (20)}}}
				\end{overpic}
  \end{minipage}}\end{minipage}
  \hfill
  \begin{minipage}{0.32\textwidth}\framebox{\begin{minipage}{\textwidth}
				\begin{overpic}[width=\textwidth, tics = 10, trim = 0 0 0 0 , clip]
					{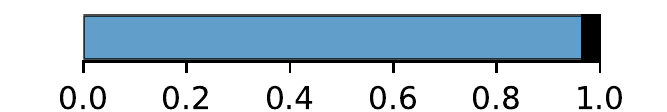}
				\end{overpic}
				\vspace*{2mm}
				\begin{overpic}[width=\textwidth, tics = 10, trim = 0 0 0 0 , clip]
					{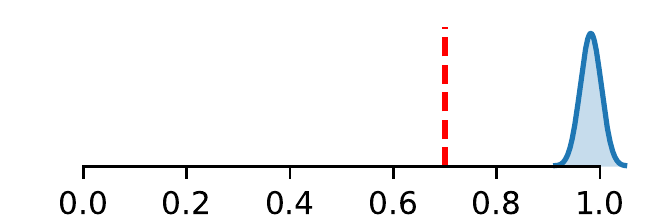}
				\end{overpic}
				\begin{overpic}[width=\textwidth, tics = 10, trim = 0 -10 0 0 , clip]
					{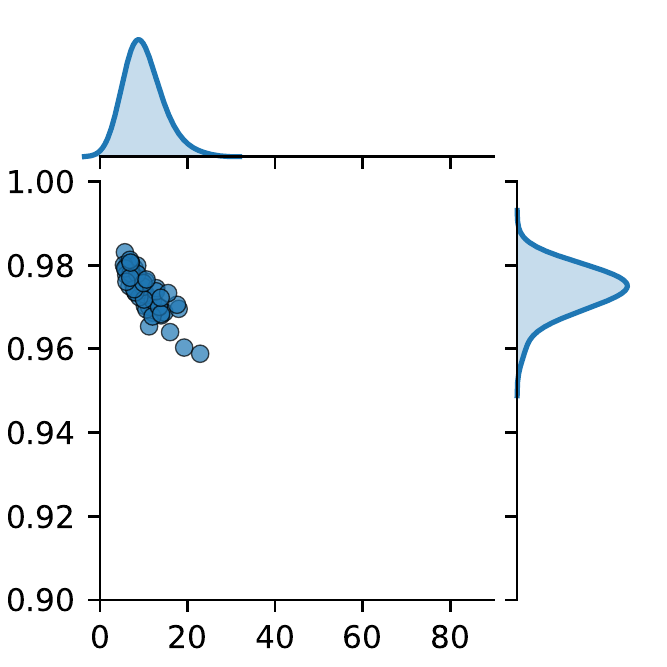}
					\put(0, 0){\textbf{\textcolor{black}{\large (21)}}}
				\end{overpic}
  \end{minipage}}\end{minipage}
  \hfill
  \begin{minipage}{0.32\textwidth}\framebox{\begin{minipage}{\textwidth}
				\begin{overpic}[width=\textwidth, tics = 10, trim = 0 0 0 0 , clip]
					{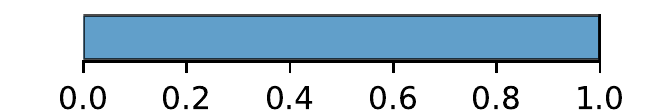}
				\end{overpic}
				\vspace*{2mm}
				\begin{overpic}[width=\textwidth, tics = 10, trim = 0 0 0 0 , clip]
					{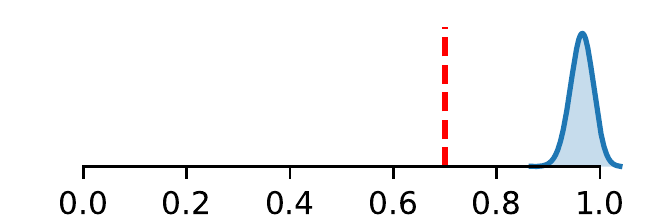}
				\end{overpic}
				\begin{overpic}[width=\textwidth, tics = 10, trim = 0 -10 0 0 , clip]
					{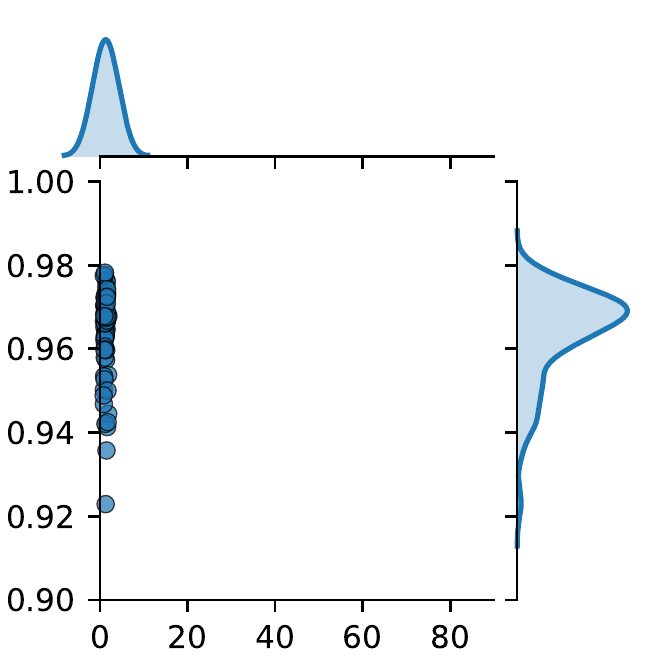}
					\put(0, 0){\textbf{\textcolor{black}{\large (22)}}}
				\end{overpic}
  \end{minipage}}\end{minipage}

  \begin{minipage}{0.32\textwidth}\framebox{\begin{minipage}{\textwidth}
				\begin{overpic}[width=\textwidth, tics = 10, trim = 0 0 0 0 , clip]
					{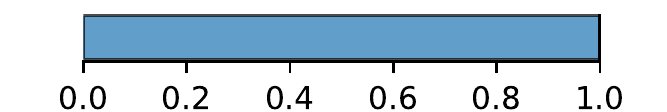}
				\end{overpic}
				\vspace*{2mm}
				\begin{overpic}[width=\textwidth, tics = 10, trim = 0 0 0 0 , clip]
					{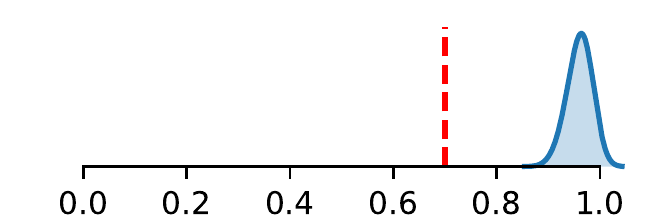}
				\end{overpic}
				\begin{overpic}[width=\textwidth, tics = 10, trim = 0 -10 0 0 , clip]
					{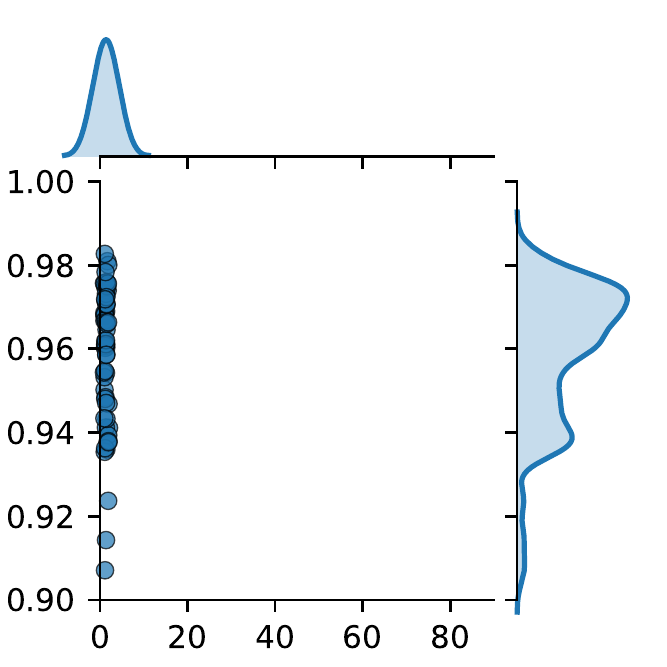}
					\put(0, 0){\textbf{\textcolor{black}{\large (23)}}}
				\end{overpic}
  \end{minipage}}\end{minipage}
  \hfill
  \begin{minipage}{0.32\textwidth}\framebox{\begin{minipage}{\textwidth}
				\begin{overpic}[width=\textwidth, tics = 10, trim = 0 0 0 0 , clip]
					{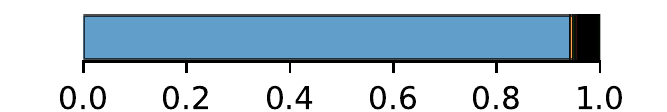}
				\end{overpic}
				\vspace*{2mm}
				\begin{overpic}[width=\textwidth, tics = 10, trim = 0 0 0 0 , clip]
					{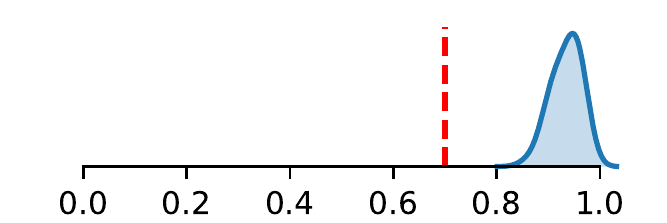}
				\end{overpic}
				\begin{overpic}[width=\textwidth, tics = 10, trim = 0 -10 0 0 , clip]
					{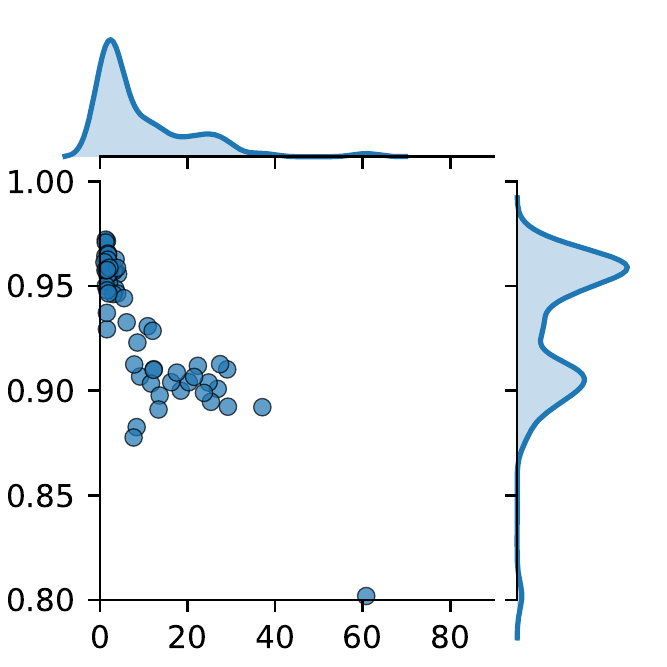}
					\put(0, 0){\textbf{\textcolor{black}{\large (24)}}}
				\end{overpic}
  \end{minipage}}\end{minipage}
  \hfill
  \begin{minipage}{0.32\textwidth}\framebox{\begin{minipage}{\textwidth}
				\begin{overpic}[width=\textwidth, tics = 10, trim = 0 0 0 0 , clip]
					{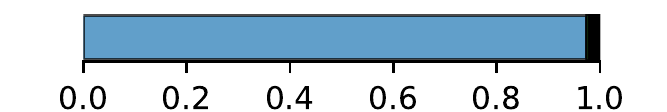}
				\end{overpic}
				\vspace*{2mm}
				\begin{overpic}[width=\textwidth, tics = 10, trim = 0 0 0 0 , clip]
					{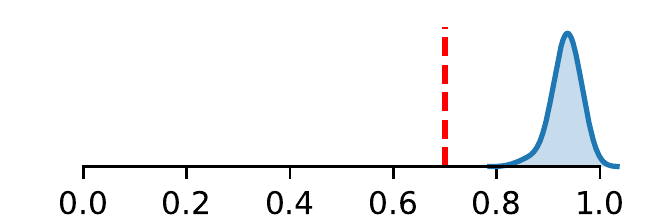}
				\end{overpic}
				\begin{overpic}[width=\textwidth, tics = 10, trim = 0 -10 0 0 , clip]
					{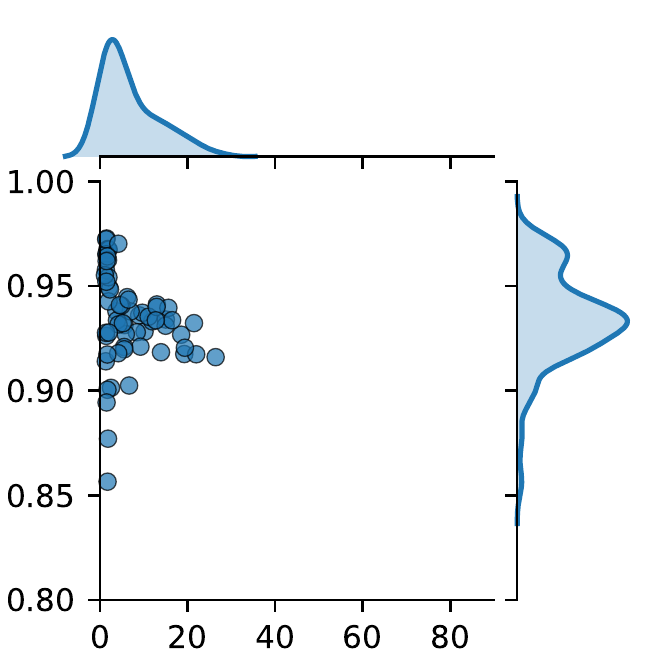}
					\put(0, 0){\textbf{\textcolor{black}{\large (25)}}}
				\end{overpic}
  \end{minipage}}\end{minipage}

\end{center}